%% file: main.tex
\documentclass[journal,twoside,svgnames]{IEEEtran}
\usepackage{amsmath,amsfonts}
\usepackage{algorithmic}
\usepackage{algorithm}
\usepackage{array}
\usepackage[caption=false,font=normalsize,labelfont=sf,textfont=sf]{subfig}
\usepackage{textcomp}
\usepackage{stfloats}
\usepackage{url}
\usepackage{verbatim}
\usepackage{graphicx}
\usepackage{cite}
\usepackage{amsmath}
\usepackage{adjustbox}
\usepackage{booktabs}
\usepackage{tabularx}
\usepackage{subfig}
\usepackage[table]{xcolor}
\usepackage{siunitx}
\usepackage{tikz}
\usetikzlibrary{arrows.meta,
                calc,
                decorations.pathreplacing,
                decorations.text,
                bending,
                fit,
                intersections,
                matrix,
                positioning,
                shapes,
                backgrounds,
                matrix}
\usepackage[utf8]{inputenc}
\usepackage{array}
\usepackage{makecell}
\usepackage{multirow} 
\usepackage{hyperref}
\newcolumntype{?}[1]{!{\vrule width #1}}
\usepackage[subtle]{savetrees}
\usepackage{balance}

\begin{document}

\title{Set2Seq Transformer: Temporal and Position-Aware Set Representations for Sequential Multiple-Instance Learning}

\author{%
  Athanasios Efthymiou,
  Stevan Rudinac,~\IEEEmembership{Member,~IEEE},
  Monika Kackovic,
  Nachoem Wijnberg,\\
  and Marcel Worring,~\IEEEmembership{Senior~Member,~IEEE}
\thanks{The authors are with the University of Amsterdam, Amsterdam, The Netherlands (e-mail: a.efthymiou@uva.nl, s.rudinac@uva.nl, m.kackovic@uva.nl, n.m.wijnberg@uva.nl, m.worring@uva.nl).}
}

\markboth{}%
{Efthymiou \MakeLowercase{et al.}: Set2Seq Transformer: Temporal and Position-Aware Set Representations for Sequential Multiple-Instance Learning}

\maketitle

\begin{abstract}
In many real-world applications, modeling both the internal structure of sets and their temporal relationships is essential for capturing complex underlying patterns. Sequential multiple-instance learning aims to address this challenge by learning permutation-invariant representations of sets distributed across discrete timesteps. However, existing methods either focus on learning set representations at a static level, ignoring temporal dynamics, or treat sequences as ordered lists of individual elements, lacking explicit mechanisms for representing sets. Crucially, effective modeling of such sequences of sets often requires encoding both the positional ordering across timesteps and their absolute temporal values to jointly capture relative progression and temporal context. In this work, we propose Set2Seq Transformer, a novel architecture that jointly models permutation-invariant set structure and temporal dependencies by learning temporal and position-aware representations of sets within a sequence in an end-to-end multimodal manner. We evaluate our Set2Seq Transformer on two tasks that require modeling set structure alongside temporal and positional patterns, but differ significantly in domain, modality, and objective. First, we consider a fine art analysis task, modeling artists' oeuvres for predicting artistic success using a novel dataset, WikiArt-Seq2Rank. Second, we utilize our Set2Seq Transformer for short-term wildfire danger forecasting. Through extensive experimentation, we show that our Set2Seq Transformer consistently improves over traditional static multiple-instance learning methods by effectively learning permutation-invariant set, temporal, and position-aware representations across diverse domains, modalities, and tasks. We release all code and datasets on this \href{https://github.com/thefth/set2seq-transformer}{GitHub repository}.
\end{abstract}


\input{Sections/introduction}

\input{Sections/related_work}

\input{Sections/approach}

\input{Sections/experimental_setup}

\input{Sections/results}

\input{Sections/discussion}

\balance

\bibliographystyle{IEEEtran}
\bibliography{bibliography}

\clearpage

\input{Sections/appendix}

\end{document}

%% file: Sections/introduction.tex
\section{Introduction}\label{sec:introduction}

Learning representations of sequences of sets is essential in various domains where observations at each timestep are naturally organized as unordered set-structured inputs. For example, in observational cosmology, detecting transients requires processing sets of astronomical observations at different sky locations, with the ordered sequence of these sets encoding a transient's appearance over time~\cite{transient_detection, machine_learning_observational_cosmology}, while in biomedical and public health tasks, predicting microsatellite status involves analyzing sets of patient-level measurements collected over clinical visits~\cite{mil_somatic_mutations, deep_learning_microsatellite_instability}. Similarly, in cultural industries, analyzing visual artists’ oeuvres entails modeling sets of artworks created at discrete timesteps\mbox{\cite{understanding_onset_hot_streaks, hot_streaks_artistic_cultural_scientific_careers}}, while in environmental forecasting, wildfire prediction relies on sequences of sets of environmental observations captured by spatially distributed sensor readings~\cite{mesogeos, wildfire_db}. Across these domains, the fundamental challenge is dual in nature: instances within each timestep lack inherent ordering and require permutation-invariant representations, whereas the timesteps themselves form structured temporal sequences that require explicit sequential modeling.

\begin{figure}[t]
\centering
\input Figures/seq2rank_task.tex
\caption{Illustration of sequential multiple-instance learning for predicting artistic success. Given an artist's body of work as a sequence (depicted in \color{red(process)}{red}\color{black}) of sets of artworks (depicted in \color{cyan(process)}{cyan}\color{black}) created at discrete timesteps, the task is to predict the artist's success according to a specific criterion. The horizontal axis represents the artist’s career stage, while the vertical axis the temporal aspect. We utilize positional encodings to represent the relative order of discrete timesteps within an artist's career, with learnable embeddings for the absolute time values (e.g., year) associated with each timestep.}
\label{fig:task}
\end{figure}

Advances in deep learning have inspired work in various real-world tasks in multimedia and related disciplines that require learning representations of set-structured inputs, ranging from 3D point cloud classification~\cite{pointnet, multiple_instance_learning_for_medical_image_and_video_analysis, order_matters, mpct, 3d_shapenets} and anomaly detection~\cite{set_transformer, deep_sets, set_norm} to few-shot learning~\cite{bruno, gradient_based_meta_learning, prototypical_networks_few_shot_learning, matching_networks} and beyond~\cite{sam_mil, attention_based_deep_mil, gmvit, wstan}, predominantly in static settings. However, modeling sequences of sets poses challenges not addressed by the existing approaches. Unlike standard supervised learning scenarios, where inputs are either individual elements or standalone unordered sets, sequential multiple-instance learning considers inputs in which each timestep comprises an unordered set of instances. Set-based representation methods~\cite{deep_sets, set_transformer, set_norm, attention_based_deep_mil, rep_sets} effectively model permutation-invariant relationships within individual sets but assume static inputs and do not capture interactions across timesteps. Conversely, sequential modeling methods~\cite{gpt3, bert, transformer} capture long-range temporal dependencies in ordered data, yet process inputs as sequences of individual elements and lack explicit mechanisms to represent sets. Naively combining these approaches in a two-stage pipeline---where set representations are learned independently and then processed sequentially---prevents joint optimization and leads to suboptimal representations.

Beyond integrating set and temporal structure, capturing only a single temporal aspect is often insufficient. In many real-world tasks, effective modeling of sequences of sets requires two complementary forms of temporal information: the sequential ordering of timesteps and their absolute temporal values. Positional information encodes the relative ordering within a sequence (e.g., timestep index 1, 2, \ldots, $N$), capturing progression patterns associated with different positions in the sequence regardless of absolute time. Temporal information, in contrast, encodes absolute calendar-time context (e.g., specific years or dates when events occur), capturing period-specific effects, long-range interactions, and broader contextual patterns tied to specific historical moments. For instance, as shown in Figure~\ref{fig:task}, in cultural industries, analyzing visual artists’ oeuvres requires reasoning over sets of artworks created at discrete timesteps, where relative career stage captures broad progression patterns (e.g., early experimentation, mature mastery), while absolute year (e.g., 1907 vs. 1945) reflects historical context such as artistic movements or cultural events. Similarly, in environmental forecasting, wildfire danger forecasting relies on relative temporal progression to capture short-term environmental trends (e.g., recent temperature increases), while absolute time reflects seasonal and interannual climate variability (e.g., summer drought vs. winter rain patterns). To capture both forms of temporal information jointly, we propose the Set2Seq Transformer, an end-to-end framework for sequential multiple-instance learning that integrates permutation-invariant set encoders with positional and temporal representations, enabling joint learning of set-level structure and sequence-level dependencies.

We evaluate the performance of our Set2Seq Transformer, focusing on two distinct tasks, namely predicting artistic success and short-term wildfire danger forecasting. For the task of predicting artistic success, we introduce a novel dataset, WikiArt-Seq2Rank~\cite{wikiart_seq2rank}, that consists of 58,458 artworks created by 849 renowned artists, where each artist is associated with scalar scores derived from multiple external success indicators. The task is formulated as a point-wise visual learning-to-rank problem, extending recent work~\cite{brmw} that examines the relationship between visual originality and career success. For the wildfire forecasting task, we use the Mesogeos dataset~\cite{mesogeos}, which involves predicting short-term wildfire danger from spatially distributed environmental observations evolving over time. These two diverse tasks enable a systematic evaluation of the performance of our Set2Seq Transformer in modeling permutation-invariant set structure alongside temporal and positional information across distinct domains. In summary, our main contributions are as follows:

\begin{itemize}

\item We introduce the Set2Seq Transformer, a novel architecture that jointly learns set, temporal, and position-aware representations for end-to-end sequential multiple-instance learning.

\item We introduce the WikiArt-Seq2Rank dataset and its associated visual learning-to-rank tasks for predicting artistic performance across multiple representative success indicators, formulated as a sequential multiple-instance learning problem.

\item We introduce a sequential multiple-instance formulation of short-term wildfire danger forecasting on the Mesogeos dataset and extend it with a proactive early-forecasting setup that evaluates performance under gradually increasing temporal context.

\item We comprehensively benchmark our Set2Seq Transformer, demonstrating its effectiveness and generalizability across structurally and semantically distinct tasks.

\end{itemize}

The rest of this paper is structured as follows. In Section~\ref{sec:related_work}, we review relevant literature on multiple-instance learning, learning temporal and position-aware representations, visual learning-to-rank and automatic fine art analysis. In Section~\ref{sec:approach}, we detail our Set2Seq Transformer architecture, and in Section~\ref{sec:experimental_setup} we present the experimental setup. Finally, in Section~\ref{sec:results} we report the experimental results, and in Section~\ref{sec:conclusion} we conclude with a brief summary of our findings.

%% file: Figures/seq2rank_task.tex
\newcommand\HUGE{\fontsize{70}{60}\selectfont}
\newcommand\HUGECDOTS{\fontsize{80}{60}\selectfont}

\definecolor{red(process)}{HTML}{b81609}
\definecolor{cyan(process)}{HTML}{09abb8}

\begin{adjustbox}{max totalsize={0.99\columnwidth}{\textheight},center,margin=.01cm 0 0 0}
\begin{tikzpicture}

\tikzstyle{group_rectangle}= [rectangle, draw, dashed, dash pattern=on 4mm off 1mm, rounded corners=1000pt, inner sep=10pt]

\tikzstyle{group_draw}= [rectangle, draw, dashed, line width=5pt, dash pattern=on 25mm off 10mm, inner sep=15pt, inner sep=1pt, rounded corners=1000pt]

\definecolor{red(process)}{HTML}{b81609}

\definecolor{cyan(process)}{HTML}{09abb8}

\usetikzlibrary{calc,3d}

\begin{scope}[]
\node[canvas is xy plane at z=0] at (0,0) (image_2_0) {\includegraphics[width=8cm]{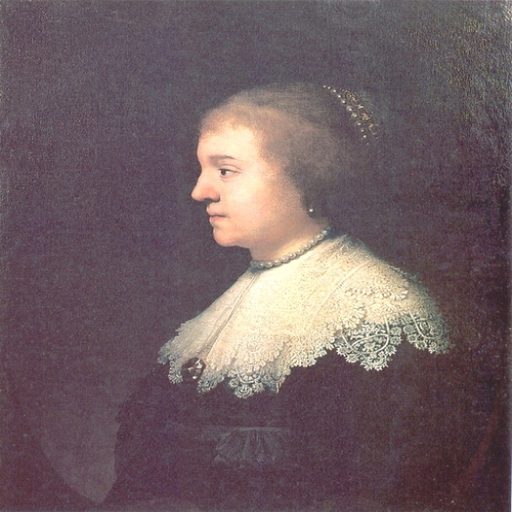}};
\node[canvas is xy plane at z=.5] at (0,0) {\includegraphics[width=8cm]{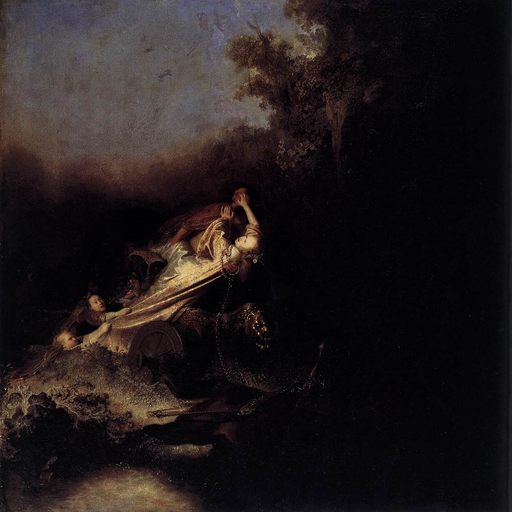}};
\node[canvas is xy plane at z=1] at (0,0) {\includegraphics[width=8cm]{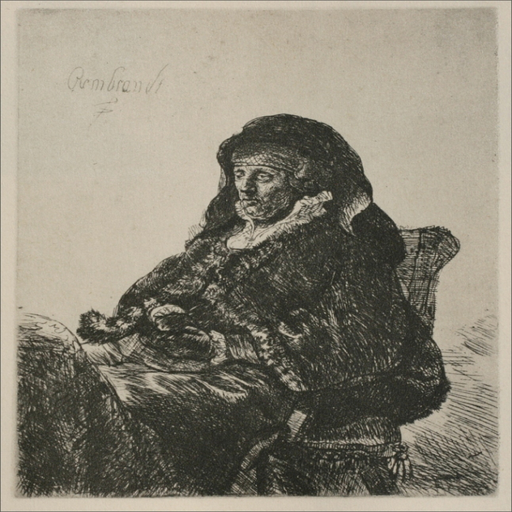}};
\node[canvas is xy plane at z=1.5] at (0,0) (image_2_1) {\includegraphics[width=8cm]{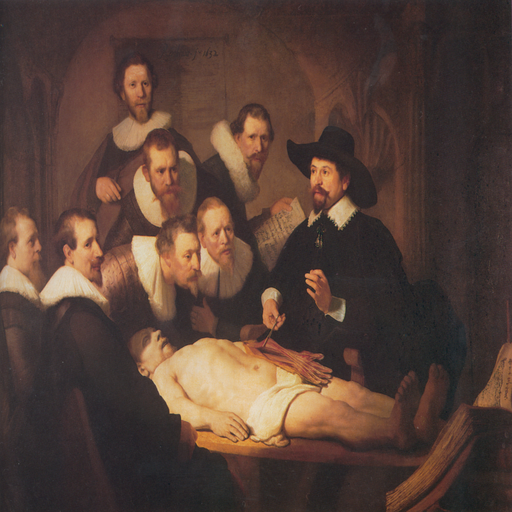}};
\node[draw, group_rectangle, cyan(process), draw, group_rectangle, rounded corners, line width=5pt, dash pattern=on 25mm off 10mm, inner sep=10pt, fit={(image_2_0) (image_2_1)}] (group_0) {};
\end{scope}

\begin{scope}[xshift=12cm, yshift=.5cm]
\node[canvas is xy plane at z=0] at (0,0) (image_2_0_) {\includegraphics[width=8cm]{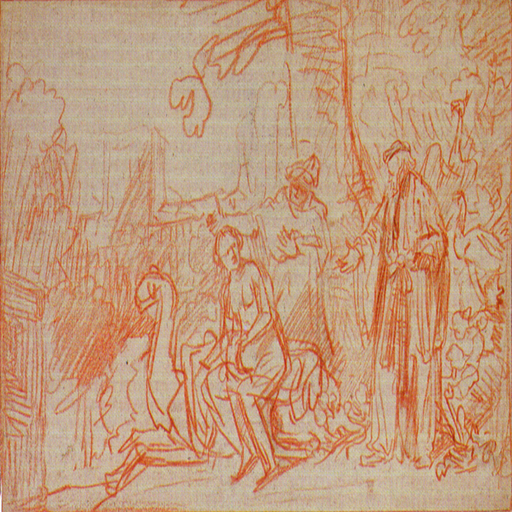}};
\node[canvas is xy plane at z=.5] at (0,0) {\includegraphics[width=8cm]{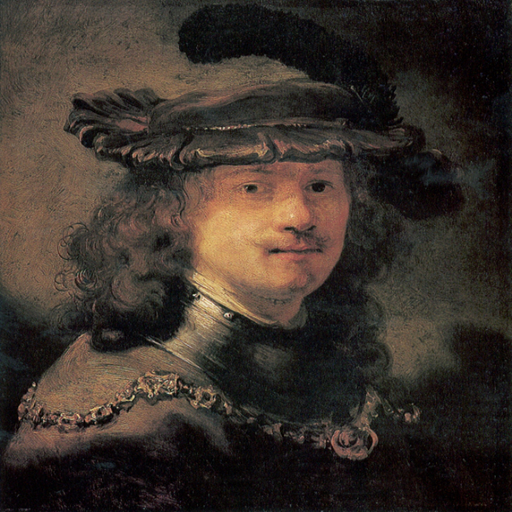}};
\node[canvas is xy plane at z=1] at (0,0) {\includegraphics[width=8cm]{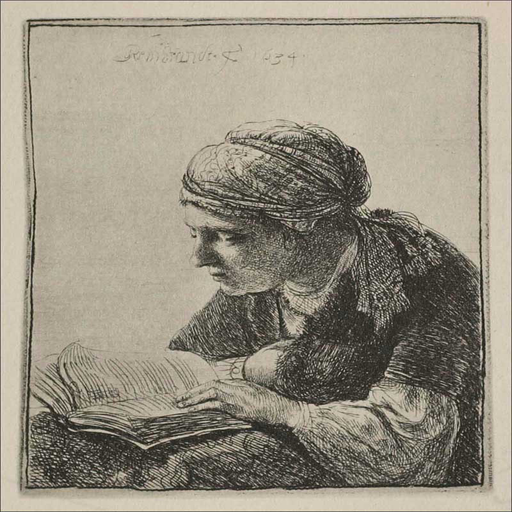}};
\node[canvas is xy plane at z=1.5] at (0,0) (image_2_1_) {\includegraphics[width=8cm]{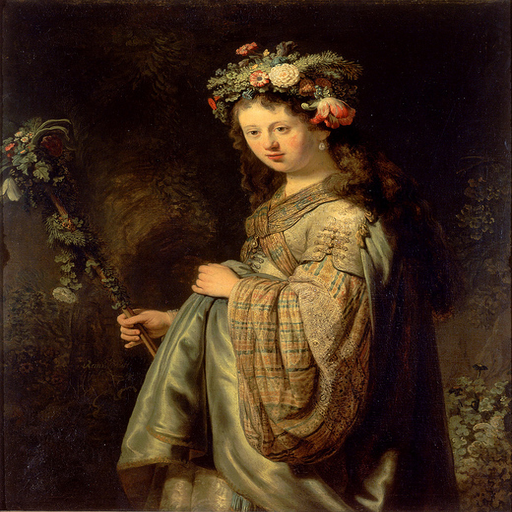}};
\node[draw, group_rectangle, cyan(process), draw, group_rectangle, rounded corners, line width=5pt, dash pattern=on 25mm off 10mm, inner sep=10pt, fit={(image_2_0_) (image_2_1_)}] (group_0) {};
\end{scope}

\begin{scope}[xshift=24cm, yshift=1cm]
\node[canvas is xy plane at z=0] at (0,0) (image_2_2__) {\includegraphics[width=8cm]{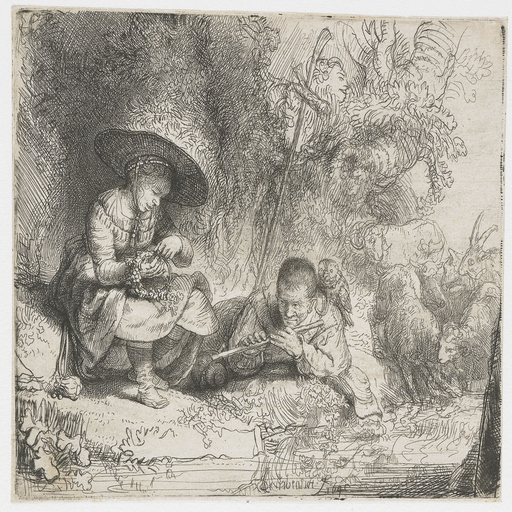}};
\node[canvas is xy plane at z=.5] at (0,0) {\includegraphics[width=8cm]{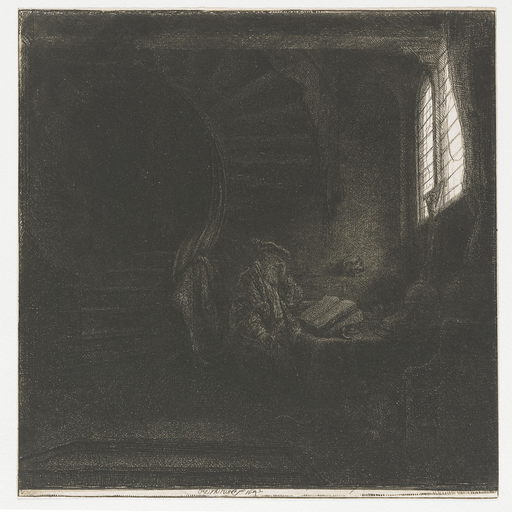}};
\node[canvas is xy plane at z=1] at (0,0) {\includegraphics[width=8cm]{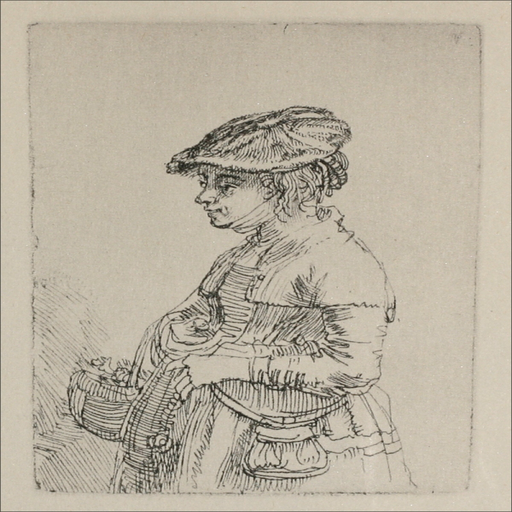}};
\node[canvas is xy plane at z=1.5] at (0,0) (image_2_3__)  {\includegraphics[width=8cm]{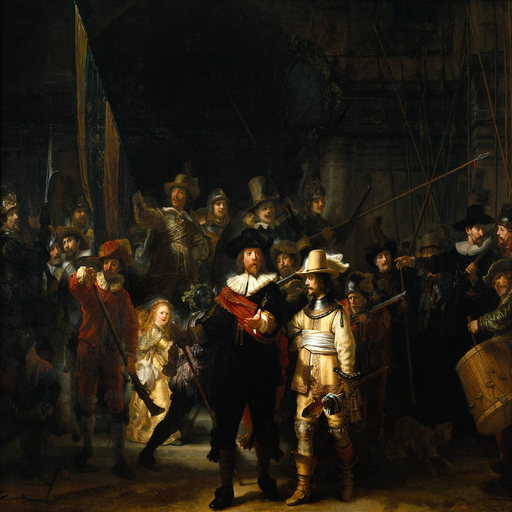}};
\node[draw, group_rectangle, cyan(process), draw, group_rectangle, rounded corners, line width=5pt, dash pattern=on 25mm off 10mm, inner sep=10pt, fit={(image_2_2__) (image_2_3__)}] (group_0) {};
\end{scope}

\begin{scope}[xshift=36cm, yshift=1.5cm]
\node[canvas is xy plane at z=0] at (0,0) (image_2_2_) {\includegraphics[width=8cm]{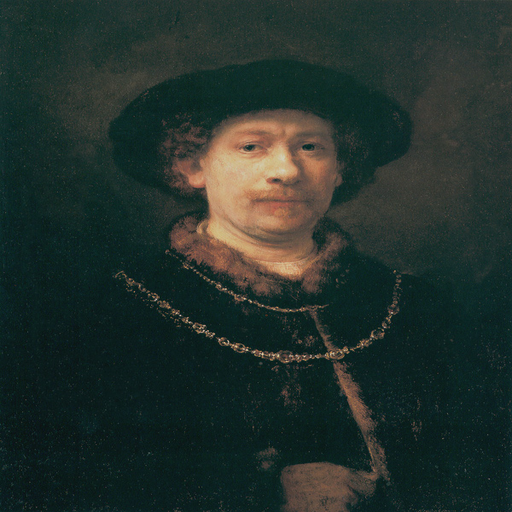}};
\node[canvas is xy plane at z=.5] at (0,0) {\includegraphics[width=8cm]{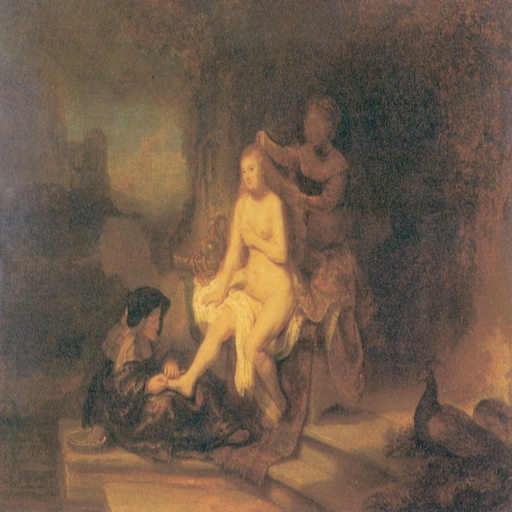}};
\node[canvas is xy plane at z=1] at (0,0) {\includegraphics[width=8cm]{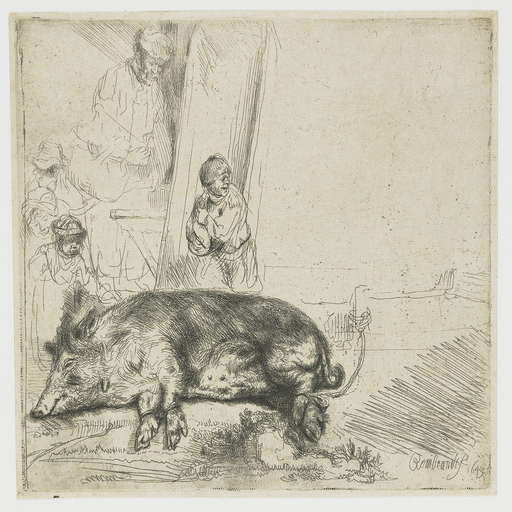}};
\node[canvas is xy plane at z=1.5] at (0,0) (image_2_3_) {\includegraphics[width=8cm]{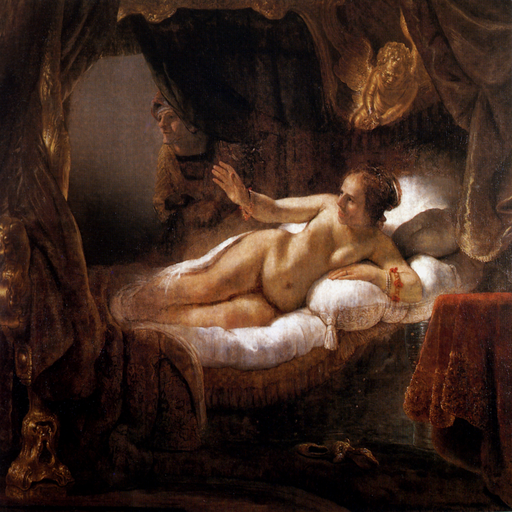}};
\node[draw, group_rectangle, cyan(process), draw, group_rectangle, rounded corners, line width=5pt, dash pattern=on 25mm off 10mm, inner sep=10pt, fit={(image_2_2_) (image_2_3_)}] (group_0) {};
\end{scope}

\begin{scope}[xshift=48cm, yshift=2cm]
\node[canvas is xy plane at z=0] at (0,0) (image_2_2) {\includegraphics[width=8cm]{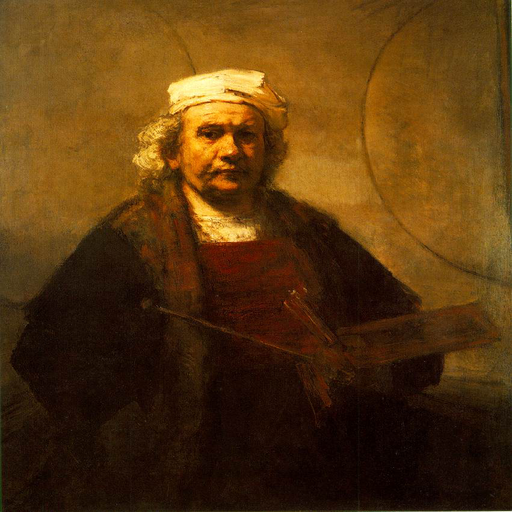}};
\node[canvas is xy plane at z=.5] at (0,0) {\includegraphics[width=8cm]{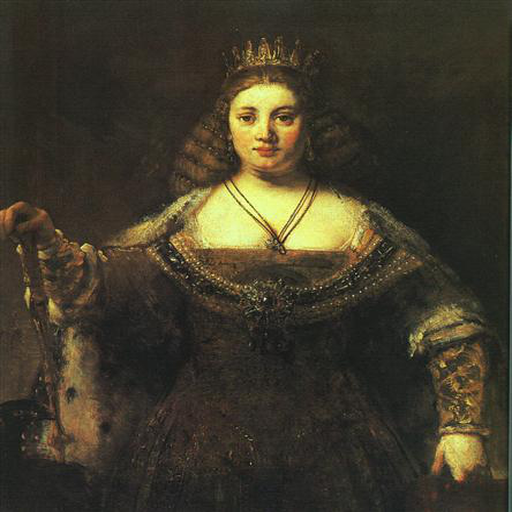}};
\node[canvas is xy plane at z=1] at (0,0) {\includegraphics[width=8cm]{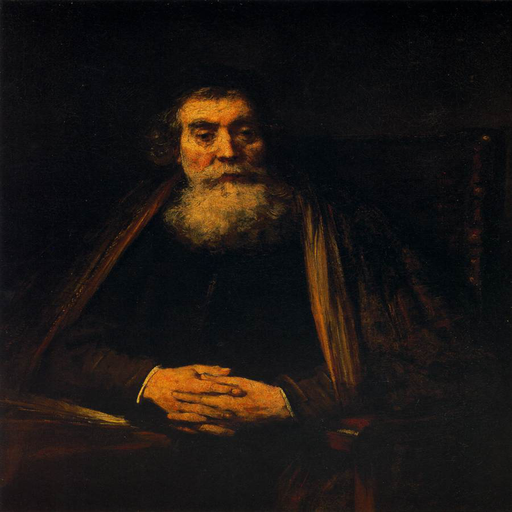}};
\node[canvas is xy plane at z=1.5] at (0,0) (image_2_3) {\includegraphics[width=8cm]{Images/images_sequence_2/image_16.png}};
\node[draw, group_rectangle, cyan(process), draw, group_rectangle, rounded corners, line width=5pt, dash pattern=on 25mm off 10mm, inner sep=10pt, fit={(image_2_2) (image_2_3)}] (group_0) {};
\end{scope}

\node[draw, group_rectangle, red(process), rounded corners, line width=5pt, dash pattern=on 25mm off 10mm, inner sep=40pt, fit={(image_2_0) (image_2_1) (image_2_2) (image_2_3)}] (group_2) {};

\begin{scope}
\node[canvas is xy plane at z=0] at (0,17) (image_1_0) {\includegraphics[width=8cm]{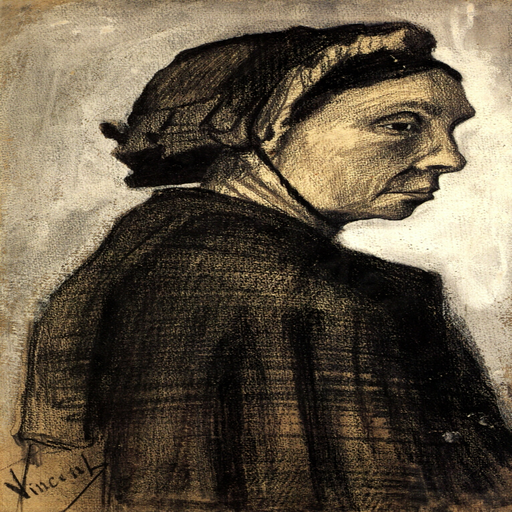}};
\node[canvas is xy plane at z=.5] at (0,17) {\includegraphics[width=8cm]{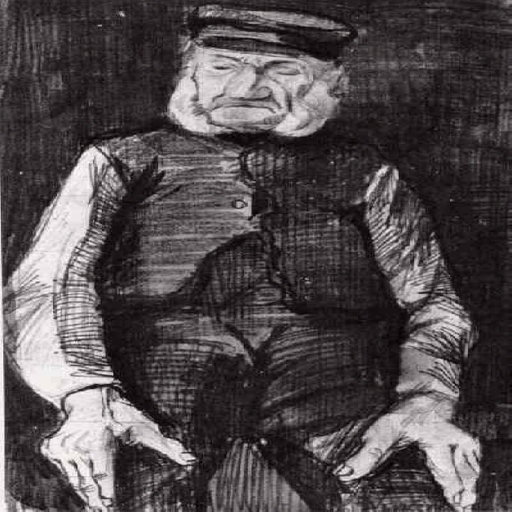}};
\node[canvas is xy plane at z=1] at (0,17) {\includegraphics[width=8cm]{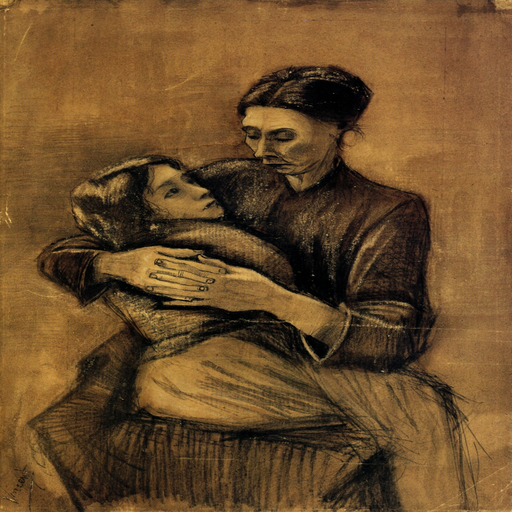}};
\node[canvas is xy plane at z=1.5] at (0,17) (image_1_1) {\includegraphics[width=8cm]{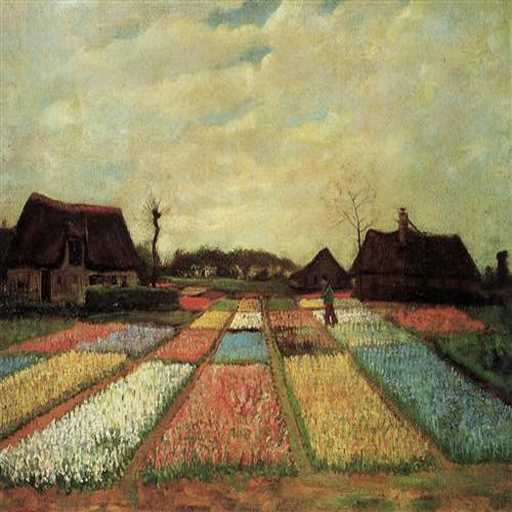}};
\node[draw, group_rectangle, cyan(process), draw, group_rectangle, rounded corners, line width=5pt, dash pattern=on 25mm off 10mm, inner sep=10pt, fit={(image_1_0) (image_1_1)}] (group_0) {};
\end{scope}

\begin{scope}[xshift=12cm, yshift=.5cm]
\node[canvas is xy plane at z=0] at (0,17) (image_1_2__) {\includegraphics[width=8cm]{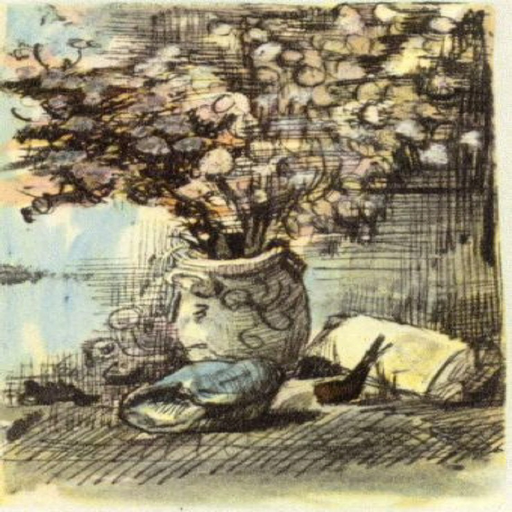}};
\node[canvas is xy plane at z=.5] at (0,17) {\includegraphics[width=8cm]{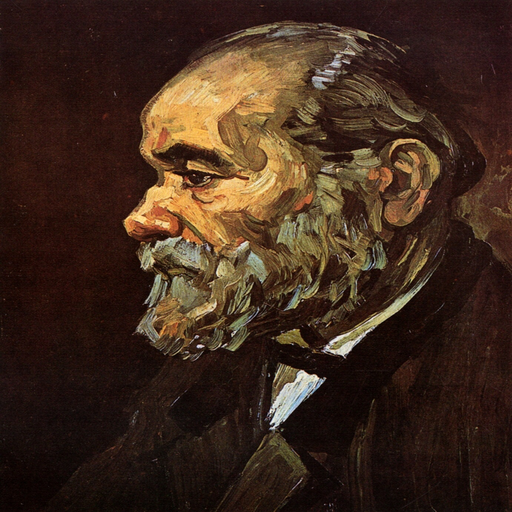}};
\node[canvas is xy plane at z=1] at (0,17) {\includegraphics[width=8cm]{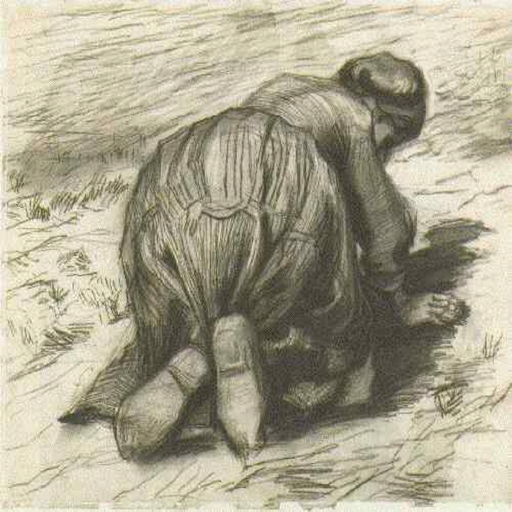}};
\node[canvas is xy plane at z=1.5] at (0,17) (image_1_3__) {\includegraphics[width=8cm]{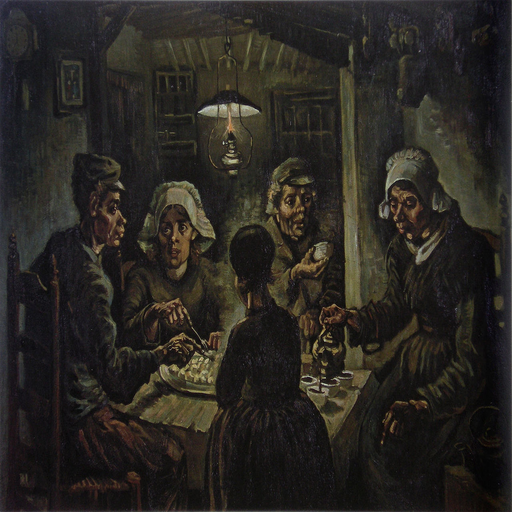}};
\node[draw, group_rectangle, cyan(process), draw, group_rectangle, rounded corners, line width=5pt, dash pattern=on 25mm off 10mm, inner sep=10pt, fit={(image_1_2__) (image_1_3__)}] (group_0) {};
\end{scope}

\begin{scope}[xshift=24cm, yshift=1cm]
\node[canvas is xy plane at z=0] at (0,17) (image_1_2__) {\includegraphics[width=8cm]{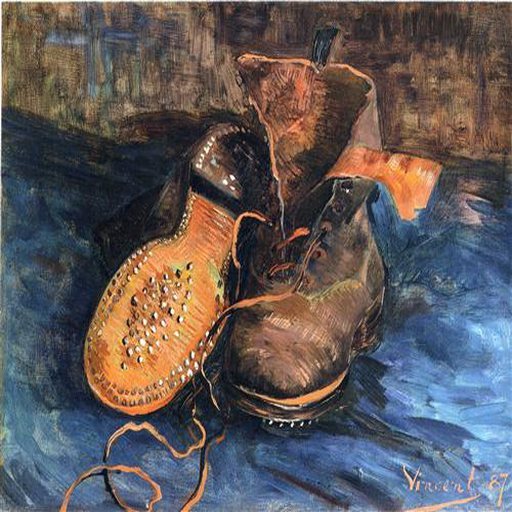}};
\node[canvas is xy plane at z=.5] at (0,17) {\includegraphics[width=8cm]{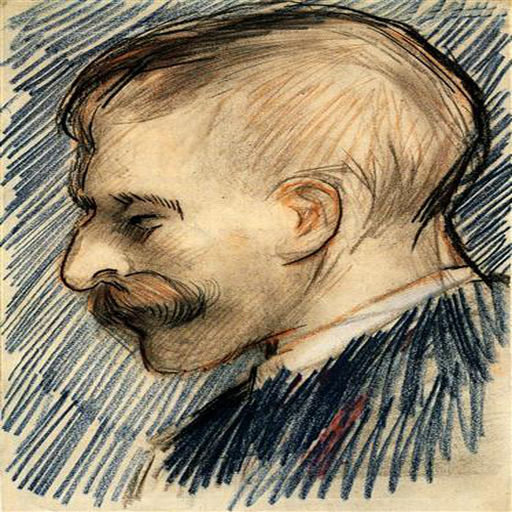}};
\node[canvas is xy plane at z=1] at (0,17) {\includegraphics[width=8cm]{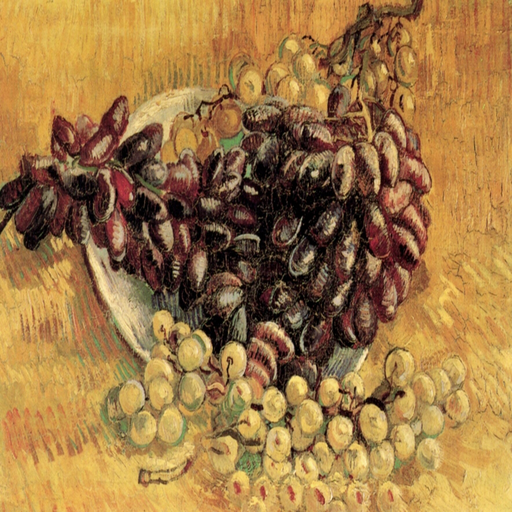}};
\node[canvas is xy plane at z=1.5] at (0,17) (image_1_3__) {\includegraphics[width=8cm]{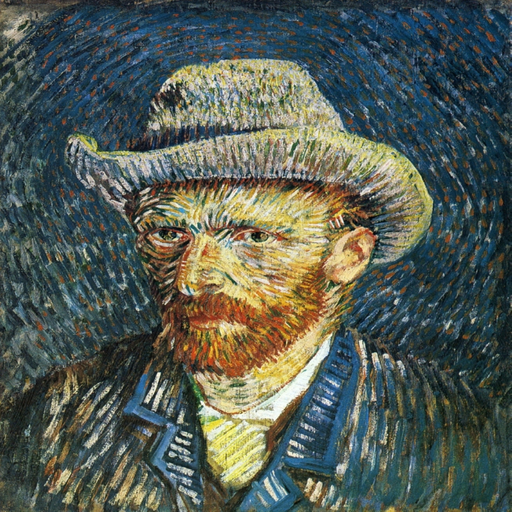}};
\node[draw, group_rectangle, cyan(process), draw, group_rectangle, rounded corners, line width=5pt, dash pattern=on 25mm off 10mm, inner sep=10pt, fit={(image_1_2__) (image_1_3__)}] (group_0) {};
\end{scope}

\begin{scope}[xshift=36cm, yshift=1.5cm]
\node[canvas is xy plane at z=0] at (0,17) (image_1_2_) {\includegraphics[width=8cm]{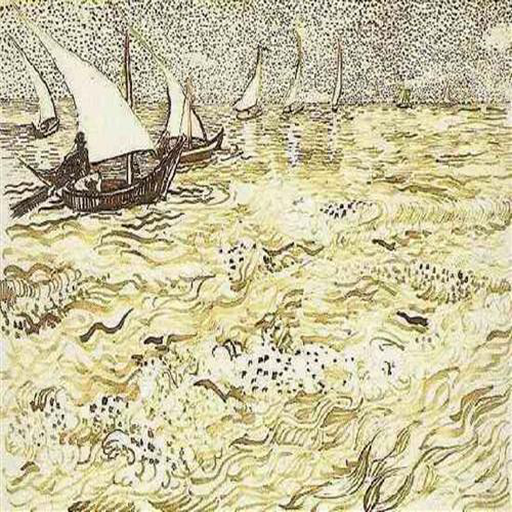}};
\node[canvas is xy plane at z=.5] at (0,17) {\includegraphics[width=8cm]{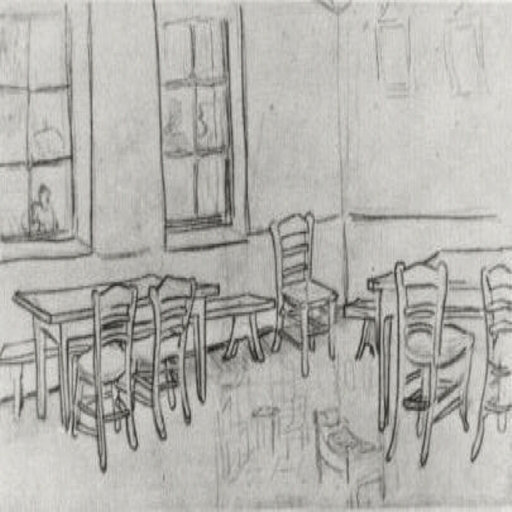}};
\node[canvas is xy plane at z=1] at (0,17) {\includegraphics[width=8cm]{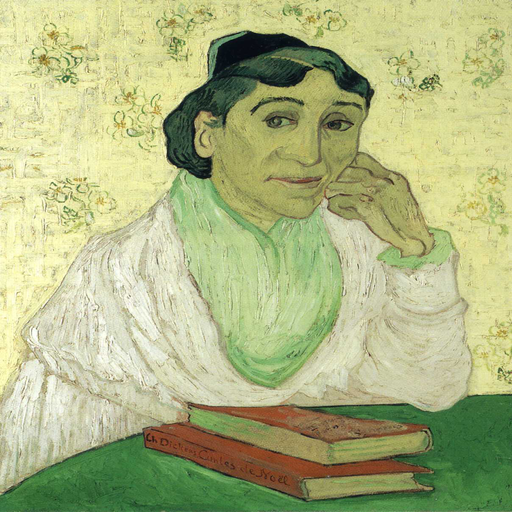}};
\node[canvas is xy plane at z=1.5] at (0,17) (image_1_3_) {\includegraphics[width=8cm]{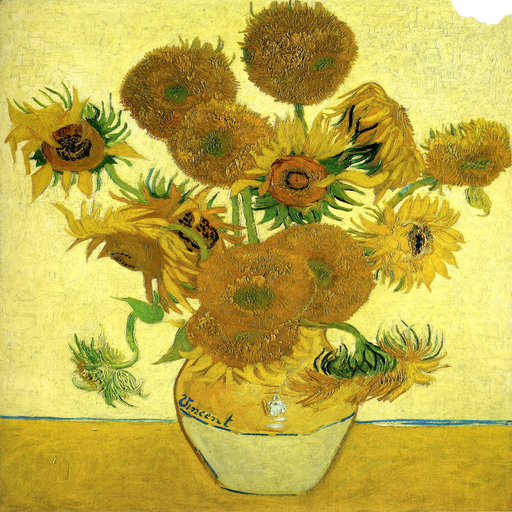}};
\node[draw, group_rectangle, cyan(process), draw, group_rectangle, rounded corners, line width=5pt, dash pattern=on 25mm off 10mm, inner sep=10pt, fit={(image_1_2_) (image_1_3_)}] (group_0) {};
\end{scope}

\begin{scope}[xshift=48cm, yshift=2cm]
\node[canvas is xy plane at z=0] at (0,17) (image_1_2) {\includegraphics[width=8cm]{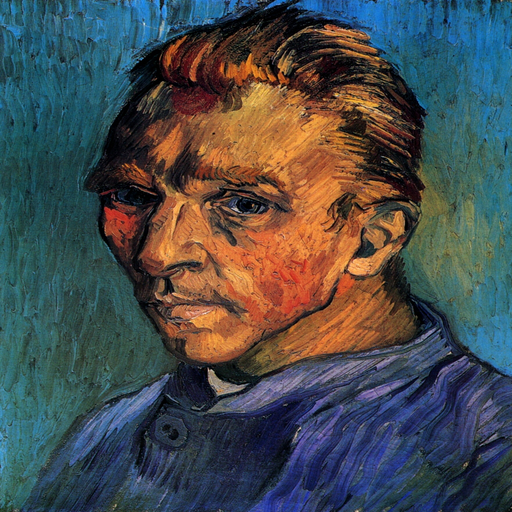}};
\node[canvas is xy plane at z=.5] at (0,17) {\includegraphics[width=8cm]{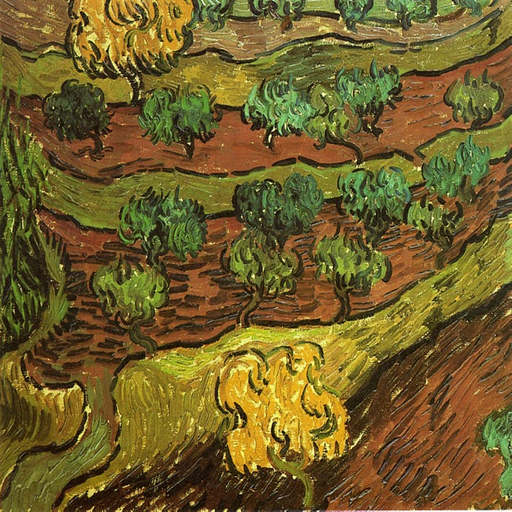}};
\node[canvas is xy plane at z=1] at (0,17) {\includegraphics[width=8cm]{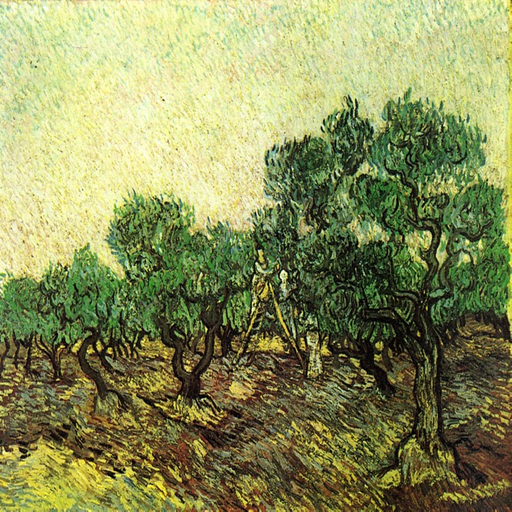}};
\node[canvas is xy plane at z=1.5] at (0,17) (image_1_3) {\includegraphics[width=8cm]{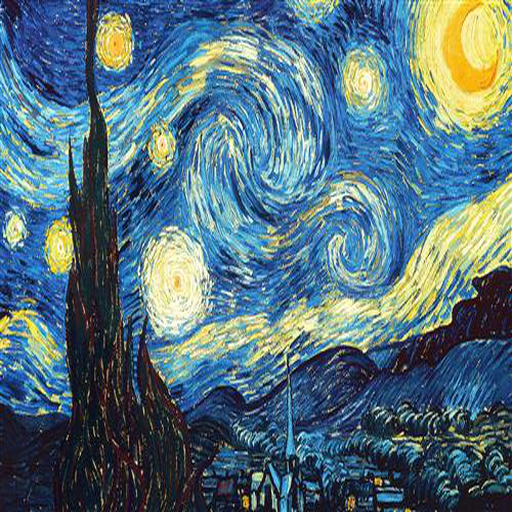}};
\node[draw, group_rectangle, cyan(process), draw, group_rectangle, rounded corners, line width=5pt, dash pattern=on 25mm off 10mm, inner sep=10pt, fit={(image_1_2) (image_1_3)}] (group_0) {};
\end{scope}

\node[draw, group_rectangle, red(process), draw, group_rectangle, red(process), rounded corners, line width=5pt, dash pattern=on 25mm off 10mm, inner sep=40pt, fit={(image_1_0) (image_1_1) (image_1_2) (image_1_3)}] (group_1) {};

% \node[rotate=90, yshift=-.5cm] (group_3) at ($(group_2)!0.5!(group_1)$) {\HUGECDOTS $\cdots$};

\begin{scope}
\node[canvas is xy plane at z=0] at (0,33) (image_0_0) {\includegraphics[width=8cm]{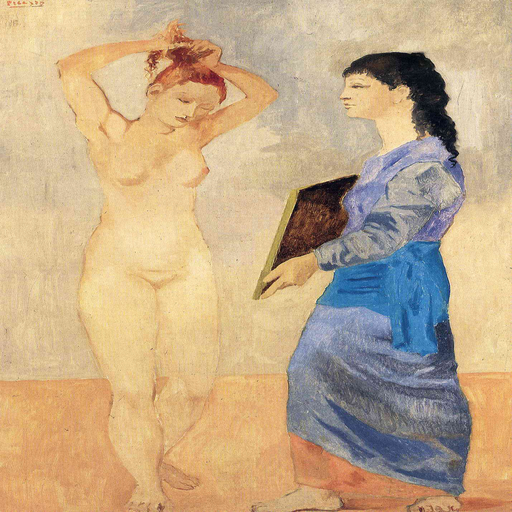}};
\node[canvas is xy plane at z=.5] at (0,33) {\includegraphics[width=8cm]{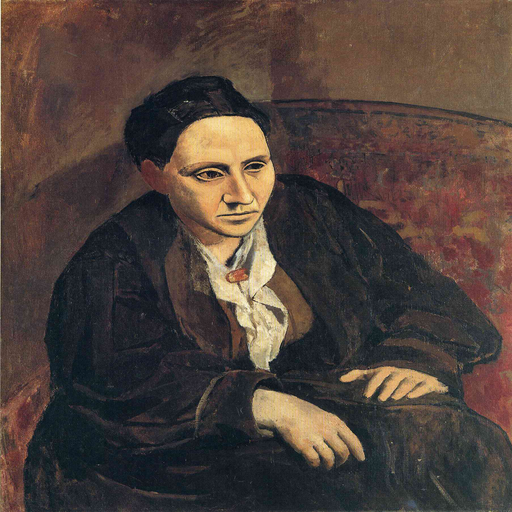}};
\node[canvas is xy plane at z=1] at (0,33) {\includegraphics[width=8cm]{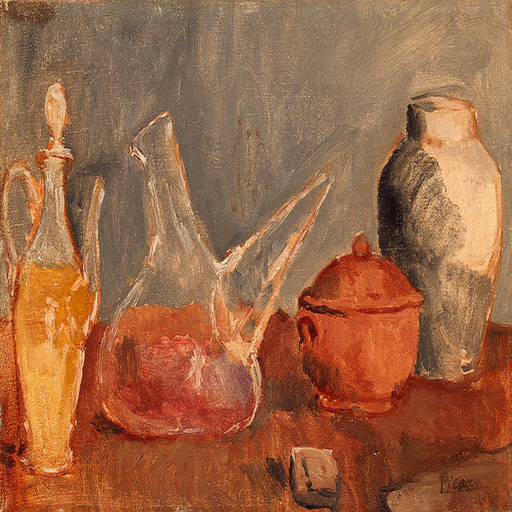}};
\node[canvas is xy plane at z=1.5] at (0,33) (image_0_1) {\includegraphics[width=8cm]{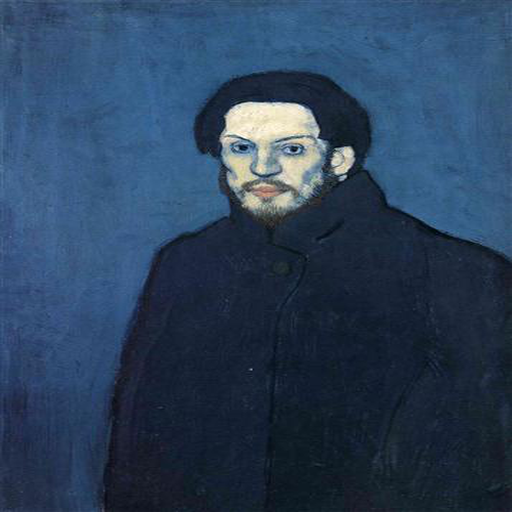}};
\node[draw, group_rectangle, cyan(process), draw, group_rectangle, rounded corners, line width=5pt, dash pattern=on 25mm off 10mm, inner sep=10pt, fit={(image_0_0) (image_0_1)}] (group_0) {};
\end{scope}

\begin{scope}[xshift=12cm, yshift=.5cm]
\node[canvas is xy plane at z=0] at (0,33) (image_0_0_) {\includegraphics[width=8cm]{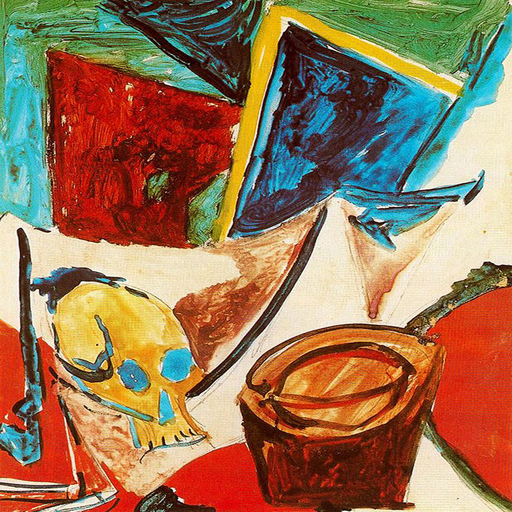}};
\node[canvas is xy plane at z=.5] at (0,33) {\includegraphics[width=8cm]{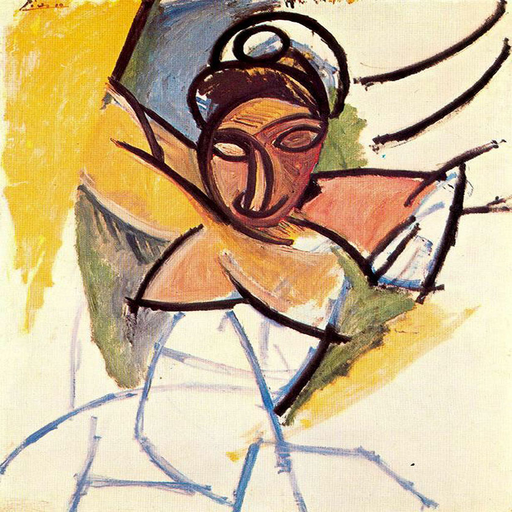}};
\node[canvas is xy plane at z=1] at (0,33) {\includegraphics[width=8cm]{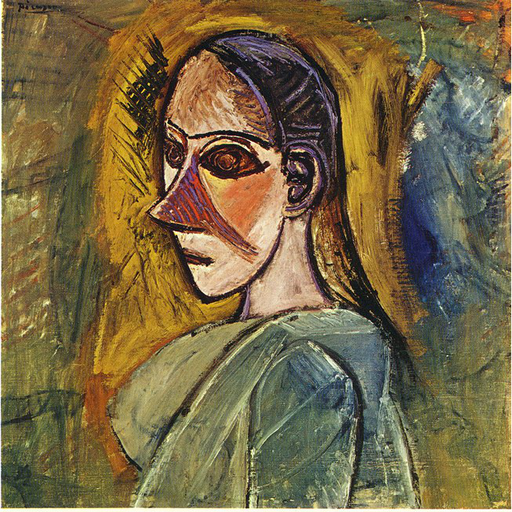}};
\node[canvas is xy plane at z=1.5] at (0,33) (image_0_1_) {\includegraphics[width=8cm]{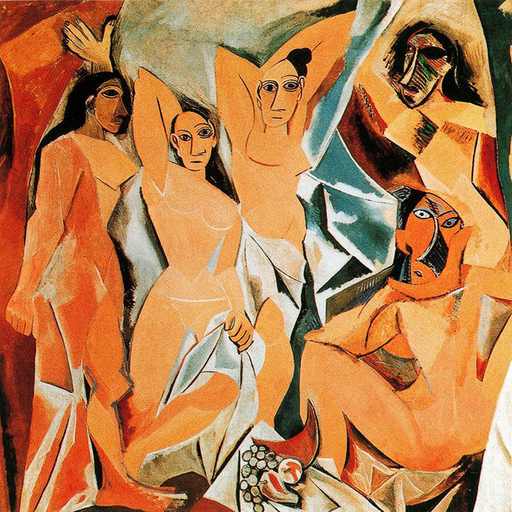}};
\node[draw, group_rectangle, cyan(process), draw, group_rectangle, rounded corners, line width=5pt, dash pattern=on 25mm off 10mm, inner sep=10pt, fit={(image_0_0_) (image_0_1_)}] (group_0) {};
\end{scope}

\begin{scope}[xshift=24cm, yshift=1cm]
\node[canvas is xy plane at z=0] at (0,33) (image_0_0__) {\includegraphics[width=8cm]{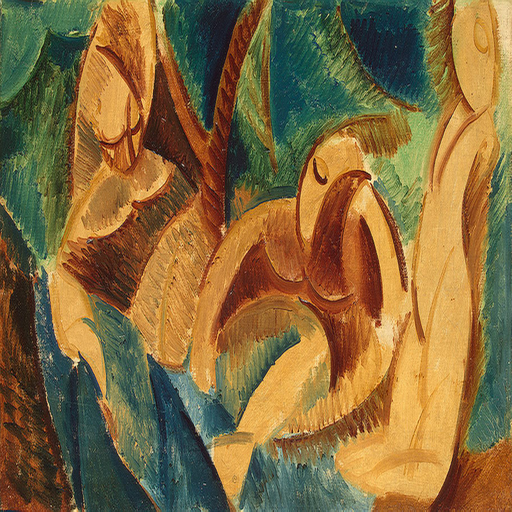}};
\node[canvas is xy plane at z=.5] at (0,33) {\includegraphics[width=8cm]{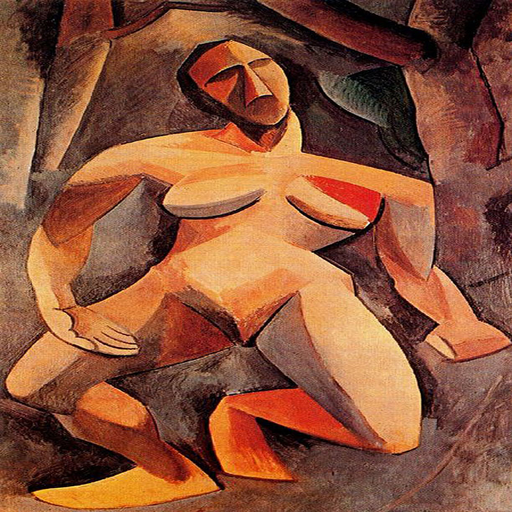}};
\node[canvas is xy plane at z=1] at (0,33) {\includegraphics[width=8cm]{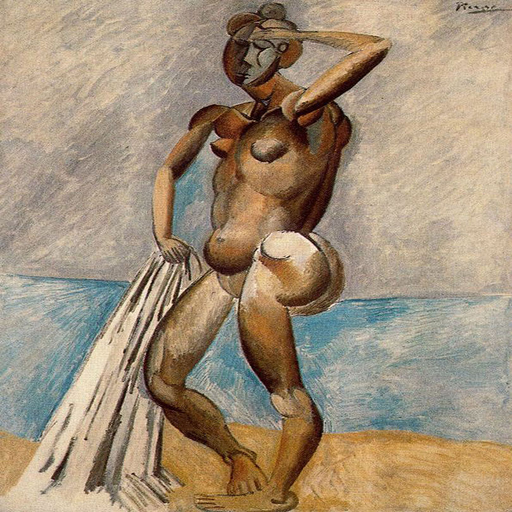}};
\node[canvas is xy plane at z=1.5] at (0,33) (image_0_1__) {\includegraphics[width=8cm]{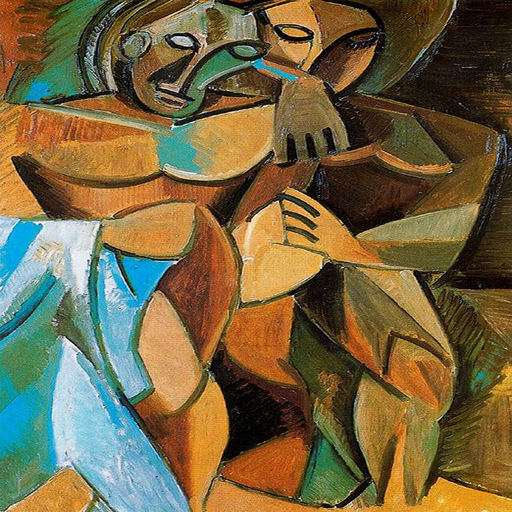}};
\node[draw, group_rectangle, cyan(process), draw, group_rectangle, rounded corners, line width=5pt, dash pattern=on 25mm off 10mm, inner sep=10pt, fit={(image_0_0__) (image_0_1__)}] (group_0) {};
\end{scope}

\begin{scope}[xshift=36cm, yshift=1.5cm]
\node[canvas is xy plane at z=0] at (0,33) (image_0_2_) {\includegraphics[width=8cm]{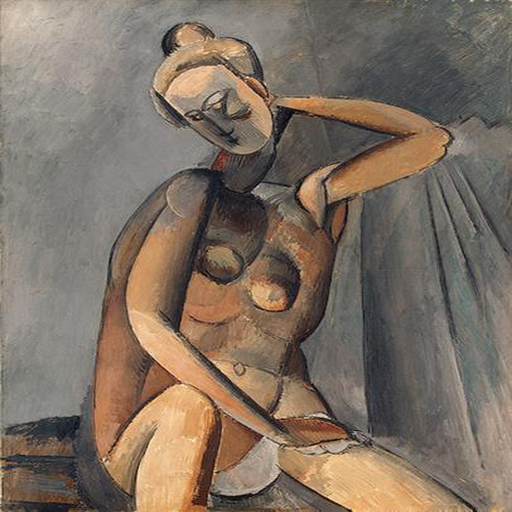}};
\node[canvas is xy plane at z=.5] at (0,33) {\includegraphics[width=8cm]{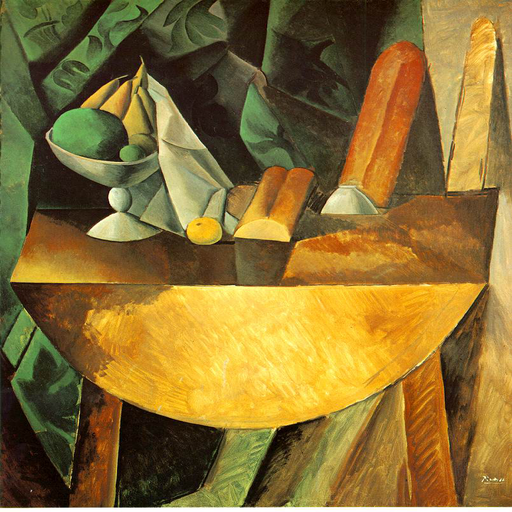}};
\node[canvas is xy plane at z=1] at (0,33) {\includegraphics[width=8cm]{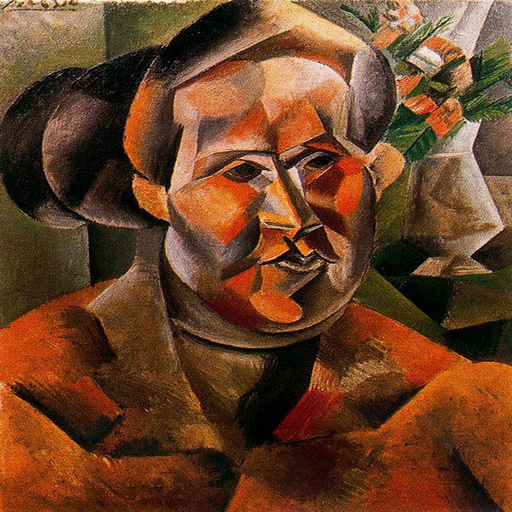}};
\node[canvas is xy plane at z=1.5] at (0,33) (image_0_3_) {\includegraphics[width=8cm]{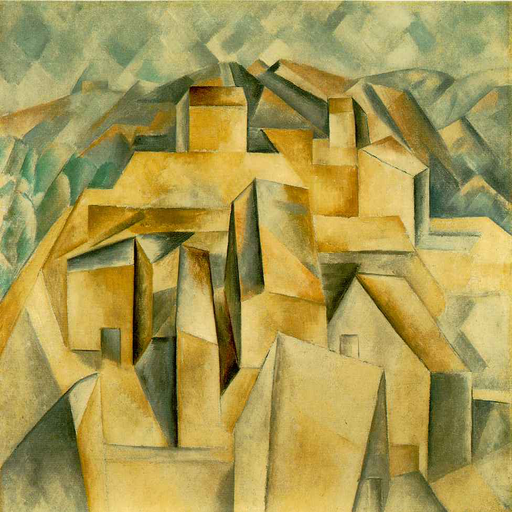}};
\node[draw, group_rectangle, cyan(process), draw, group_rectangle, rounded corners, line width=5pt, dash pattern=on 25mm off 10mm, inner sep=10pt, fit={(image_0_2_) (image_0_3_)}] (group_0) {};
\end{scope}

\begin{scope}[xshift=48cm, yshift=2cm]
\node[canvas is xy plane at z=0] at (0,33) (image_0_2) {\includegraphics[width=8cm]{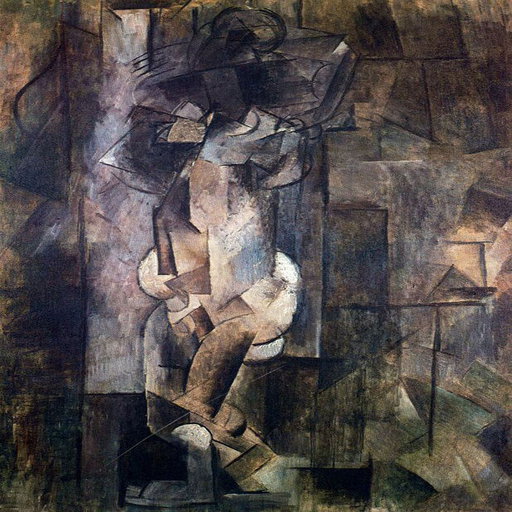}};
\node[canvas is xy plane at z=.5] at (0,33) {\includegraphics[width=8cm]{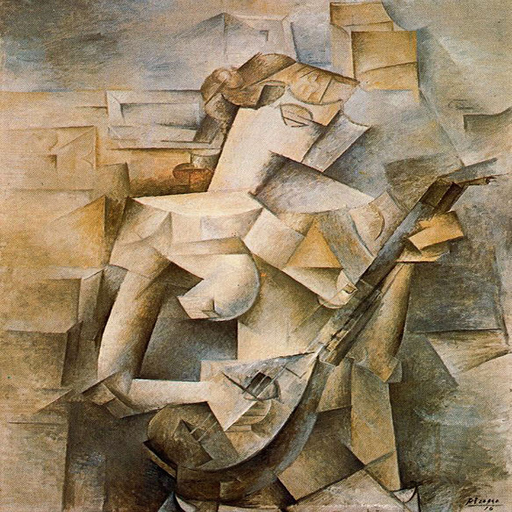}};
\node[canvas is xy plane at z=1] at (0,33) {\includegraphics[width=8cm]{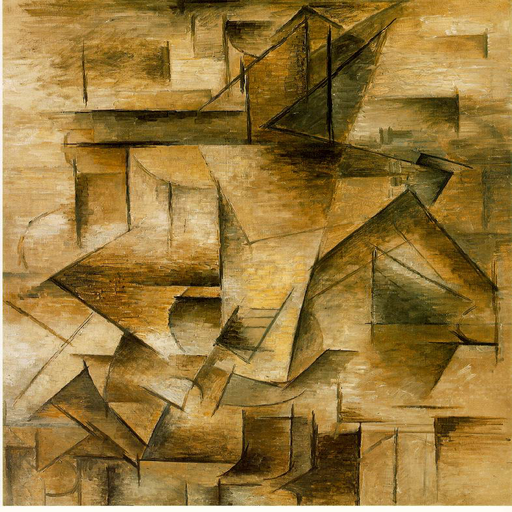}};
\node[canvas is xy plane at z=1.5] at (0,33) (image_0_3) {\includegraphics[width=8cm]{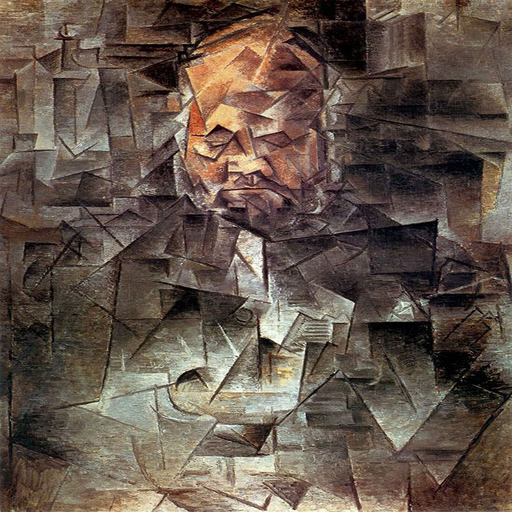}};
\node[draw, group_rectangle, cyan(process), draw, group_rectangle, rounded corners, line width=5pt, dash pattern=on 25mm off 10mm, inner sep=10pt, fit={(image_0_2) (image_0_3)}] (group_0) {};
\end{scope}

\node[draw, group_rectangle, red(process), draw, group_rectangle, red(process), rounded corners, line width=5pt, dash pattern=on 25mm off 10mm, inner sep=40pt, fit={(image_0_0) (image_0_1) (image_0_2) (image_0_3)}] (group_0) {};

 \node[below left=4cm and .5cm of image_2_0, xshift=-.8cm] (time_0) {};

\node[right=60cm of time_0, xshift=.8cm] (time_1) {};

\path[draw, thick, postaction={decorate,decoration={text along path, text align=center,text={|\HUGE\bf|Career Progress},raise=-70pt}}, -{Latex[length=8mm,width=8mm]}] (time_0.east) to (time_1.west);

 \node[below left=2cm and 3cm of image_2_0, xshift=-.8cm] (time_0_) {};

\node[above=48cm of time_0_] (time_1_) {};

\path[draw, thick, postaction={decorate,decoration={text along path, text align=center,text={|\HUGE\bf|Time},raise=40pt}}, -{Latex[length=8mm,width=8mm]}] (time_0_.north) to (time_1_.south);

 \node[right=4cm of group_0] (score_0) {\HUGE 0.8};
    \path[draw, thick, -{Latex[length=4mm,width=4mm]}] (group_0.east) to (score_0.west);

    \node[right=4cm of group_1] (score_1) {\HUGE 0.9};
    \path[draw, thick, -{Latex[length=4mm,width=4mm]}] (group_1.east) to (score_1.west);

    \node[right=4cm of group_2] (score_2) {\HUGE 0.5};
    \path[draw, thick, -{Latex[length=4mm,width=4mm]}] (group_2.east) to (score_2.west);

    \node[above right=-1.5cm and 8cm of score_0, minimum height=6cm, minimum width=8cm, inner sep=1cm] (ranking_0) {\HUGE 0.9};

    \draw[rounded corners, line width=5pt, dash pattern=on 25mm off 10mm, inner sep=15pt](ranking_0.south west)--(ranking_0.south east);
   
    \node[below=0cm of ranking_0, minimum height=6cm, minimum width=8cm, inner sep=1cm] (ranking_1) {\HUGE 0.8};

    \draw[rounded corners, line width=5pt, dash pattern=on 25mm off 10mm, inner sep=15pt](ranking_1.south west)--(ranking_1.south east);

    \node[below=6cm of ranking_1, minimum height=6cm, minimum width=8cm, inner sep=1cm] (ranking_2) {\HUGE 0.5};

    \draw[rounded corners, line width=5pt, dash pattern=on 25mm off 10mm, inner sep=15pt](ranking_2.north west)--(ranking_2.north east);
    \draw[rounded corners, line width=5pt, dash pattern=on 25mm off 10mm, inner sep=15pt](ranking_2.south west)--(ranking_2.south east);
    
    \node[rotate=90, yshift=-.1cm] (ranking_3) at ($(ranking_1)!0.5!(ranking_2)$) {\HUGECDOTS \dots};
    
    \node[below=6cm of ranking_2, minimum height=6cm, minimum width=8cm, inner sep=1cm] (ranking_4) {\phantom{\HUGE 0.0}};

    \node[rotate=90, yshift=-.1cm] (ranking_5) at ($(ranking_2)!0.5!(ranking_4)$) {\HUGECDOTS \dots};

    \node[rotate=90, yshift=-.1cm] (ranking_6) at ($(ranking_4)!0.!(ranking_5)$) {\HUGECDOTS \dots};

     \draw[rounded corners, line width=5pt, dash pattern=on 25mm off 10mm, inner sep=15pt](ranking_4.north west)--(ranking_4.north east);

    \node[draw, group_rectangle, rounded corners, line width=5pt, dash pattern=on 25mm off 10mm, inner sep=15pt, rounded corners, inner sep=.1cm, fit={(ranking_0) (ranking_4)}] (ranking) {};

   \draw[-{Latex[length=10mm,width=10mm]}]    (score_0.north east) to[out=90,in=180] ([yshift=.2cm]ranking_1.west);
    
    \draw[-{Latex[length=10mm,width=10mm]}]    (score_1.north east) to[out=90,in=180] ([yshift=.2cm]ranking_0.west);
    
    \draw[-{Latex[length=10mm,width=10mm]}]    (score_2.north east) to[out=90,in=180] ([yshift=.2cm]ranking_2.west);

    \node[below right=-15cm and 1cm of ranking, rotate=90] (ranking_text) {\HUGE{\textbf{Relevance}}};

\end{tikzpicture}
\end{adjustbox}

%% file: Sections/related_work.tex
\section{Related Work}\label{sec:related_work}

\begin{figure*}[t]
\centering
\input Figures/set_transformer2transformer.tex
\caption{This figure illustrates the Set2Seq Transformer. For an ordered sequence $\mathcal{T}$ of unordered sets $\mathcal{S}$, we learn a representation $\mathbf{s}_i$ for each set at timestep $t_i$. Positional encoding $\mathbf{u}_i$ and temporal embedding $\mathbf{v}_i$ are used for each timestep $t_i$, and summed with $\mathbf{s}_i$ to obtain the final representation $\boldsymbol{\tau}_i$ for the timestep $t_i$. These representations are then passed through a Transformer encoder, followed by a fully-connected layer, to generate the final prediction $\hat{y}$, derived from the last timestep $t_N$.}
\label{fig:set_transformer2transformer}
\end{figure*}

Deep learning has enabled a broad range of methods for multiple-instance learning~\cite{deep0tag, set_norm, set_transformer, deep_sets} and sequence modeling~\cite{gpt3, bert, transformer, order_matters}. Furthermore, advancements in computer vision and multimedia have spurred research in diverse applications, such as visual learning-to-rank~\cite{ranking_cnn_for_age_estimation, ordinal_regression_for_age_estimation, vitor} and automated fine art analysis~\cite{exploring_representativity_art_paintings, elgammal2018shape, strezoski_omniart}. Below, we briefly survey representative methods from these domains.

\subsection{Multiple-Instance Learning}\label{subsec:multiple_instance_learning}

Multiple-instance learning has been a subject of long-standing research~\cite{mil_axis_parallel_rectangles, a_framework_for_mil}. Early methods such as PointNet~\cite{pointnet}, DeepSets~\cite{deep_sets}, and RepSet~\cite{rep_sets} focus on permutation-invariant and equivariant set representations, leveraging aggregation operators and matching-based mechanisms. Ilse et al.~\cite{attention_based_deep_mil} introduce attention-based pooling to provide insight into the contribution of individual instances within sets. Lee et al.~\cite{set_transformer} propose Set Transformer, an architecture that employs self-attention for learning permutation-invariant representations. More recently, Zhang et al.~\cite{set_norm} address limitations of DeepSets~\cite{deep_sets} and Set Transformer~\cite{set_transformer}, achieving improved robustness and performance across benchmarks.

Advances in deep learning have also inspired work on sequential multiple-instance learning, where inputs consist of temporally ordered sets~\cite{sequences_of_sets, set2seq_methods, set_interdependence_transformer, permutation_invariant_feature_restructuring}. The seminal work of Vinyals et al.~\cite{order_matters} utilizes a seq2seq framework with RNNs and LSTMs~\cite{lstm} coupled with attention to learn representations of ordered inputs. Subsequent work extends these ideas to tasks such as sentence ordering~\cite{sentence_ordering_rnn, learning_to_order_sentences, reorder_bart}, text summarization~\cite{extracting_diverse_keyphrases}, and multi-agent systems~\cite{value_decomposition_networks}. Related work has also considered temporal sequences of sets, particularly in autoregressive settings for next-set prediction~\cite{next_basket_prediction, repeat_bias_aware_temporal_set_prediction, continuous_time_user_preference_temporal_sets_prediction, temporal_sets_prediction_siplified, predicting_temporal_sets}, where the objective is to forecast future set composition. Unlike next-set prediction methods, our work focuses on sequence-level prediction over entire trajectories, requiring permutation-invariant representations within timesteps while jointly modeling temporal dependencies across timesteps to capture trajectory-level properties.

\subsection{Learning Temporal and Position-Aware Representations}\label{subsec:learning_position_aware_represntations}

Sequence modeling has been extensively studied, with standard approaches such as RNNs~\cite{650093} and their variants~\cite{lstm, gru}, achieving strong performance across diverse tasks~\cite{seq2seq, order_matters, video_captioning_with_attention_based_lstm}. The success of Transformers~\cite{transformer} has driven extensive work on position-aware representations, including sinusoidal positional encodings and relative position representations~\cite{transformer_xl, self_attention_relative_position_representations}, which encode the ordering of elements within sequences. Extensions have explored learnable positional embeddings~\cite{bert, convolutional_sequence_to_sequence_learning, t5} and alternative encoding schemes~\cite{alibi, roformer}. However, these methods model positional structure within sequences---the relative ordering of elements---rather than the absolute temporal context of when observations occur.

Prior work has explored temporal-aware representations that explicitly incorporate time. Methods such as Time2Vec~\cite{time2vec}, Transformer Hawkes Process~\cite{thp}, Temporal Fusion Transformer~\cite{tft}, and TimeMIL~\cite{time_mil} integrate temporal information for time series and event sequences. However, these approaches operate on sequences where each timestep corresponds to a single instance and do not model the distinction between relative positional structure and absolute calendar-time context. In contrast, our work models sequences of unordered sets, requiring permutation-invariant representations within timesteps alongside temporal modeling across timesteps.

\subsection{Visual Learning-to-Rank}\label{subsec:visual_learning_to_rank}

Learning-to-rank is one of the most prominent tasks in information retrieval and it has been successfully applied in a number of works~\cite{sentence_ordering_rnn, reorder_bart, ranking_cnn_for_age_estimation}. Recent advancements in computer vision and the large amount of available datasets has increased research also in visual learning-to-rank. For instance, van den Akker et al.~\cite{vitor} proposes ViTOR, a method that learns to rank webpages solely based on their visual appearance. In addition, many visual learning-to-rank approaches have been adapted for the specific application of age estimation ~\cite{ranking_cnn_for_age_estimation, ordinal_regression_for_age_estimation, order_learning_age_estimation}. Contrary to this line of research, which typically focus on ranking individual images or instances based on estimated age, our approach is a point-wise visual learning-to-rank method that ranks entire sequences of sets, capturing both temporal dependencies and relative positioning across multiple sets within a sequence.

\subsection{Automatic Fine Art Analysis}\label{subsec:learning_visual_representations_for_fine_art_analysis}

The large amount of publicly available art-related datasets~\cite{strezoski_omniart, tan_artgan, the_met_dataset} has inspired research in learning visual representations for various fine art analysis tasks, ranging from fine art categorization~\cite{artgraph, elgammal2018shape, artsagenet} and studying semantic principles of the fine art history~\cite{cetinic_art_history_cnn, principals_semantics} to style transfer~\cite{aesthetic_aware_image_style_transfer, learning_dynamic_style_kernels_for_artistic_style_transfer} among many others~\cite{explain_me_the_painting, taming_clip_art, gallery_gpt, cetinic_understanding, exploring_representativity_art_paintings, artemis_v2}. 
Recent work has also examined the relationship between artistic features and career success~\cite{brmw}, establishing frameworks for ranking artists based on various success indicators. Building upon this foundation, we extend the analysis from static feature-based approaches to end-to-end sequential modeling. In contrast to prior work that analyzes artworks at a static single-instance level, we learn representations of sequences of sets of artworks created by artists over time, capturing temporal career progression to predict artistic performance across multiple representative success indicators.

%% file: Figures/set_transformer2transformer.tex
\begin{adjustbox}{max totalsize={\textwidth}{.85\textheight},center}
\begin{tikzpicture}

\tikzstyle{group_circle}= [circle, draw, dashed]
  \tikzstyle{group_embeddings}= [rectangle, draw, thick, rounded corners=5pt, inner sep=3pt]
  
    \node[] (image_0_0) {\includegraphics[width=.05\textwidth]{Images/images_sequence_0/image_0.png}};
\node[circle,draw=black,fill=gray!50,inner sep=0pt,minimum size=4pt, rotate=90,
above left=.3cm and -.3 of image_0_0] (univariate-0) {};
\node[circle,draw=black,fill=gray!50,inner sep=0pt,minimum size=4pt, rotate=-90, right=.2cm of univariate-0] (univariate-1) {};
\node[circle,draw=black,fill=gray!50,inner sep=0pt,minimum size=4pt, rotate=90,  right=.2cm of univariate-1] (univariate-2) {};
\node[circle,draw=black,fill=gray!50,inner sep=0pt,minimum size=4pt, rotate=-90, right=.2cm of univariate-2] (univariate-3) {};

\node[group_embeddings, fit={(univariate-0) (univariate-3)}] (embedding_1) {};
    \node[below right = .2cmof image_0_0] (image_0_1) {\includegraphics[width=.05\textwidth]{Images/images_sequence_0/image_1.png}};
    \node[circle,draw=black,fill=gray!50,inner sep=0pt,minimum size=4pt, rotate=90,
above left=.3cm and -.3 of image_0_1] (univariate-0) {};
\node[circle,draw=black,fill=gray!50,inner sep=0pt,minimum size=4pt, rotate=-90, right=.2cm of univariate-0] (univariate-1) {};
\node[circle,draw=black,fill=gray!50,inner sep=0pt,minimum size=4pt, rotate=90,  right=.2cm of univariate-1] (univariate-2) {};
\node[circle,draw=black,fill=gray!50,inner sep=0pt,minimum size=4pt, rotate=-90, right=.2cm of univariate-2] (univariate-3) {};

\node[group_embeddings, fit={(univariate-0) (univariate-3)}] (embedding_2) {};

    \node[above right = .2cmof image_0_1] (image_0_2) {\LARGE{$\cdots$}};
    
    \node[right = .2cmof image_0_2] (image_0_3) {\includegraphics[width=.05\textwidth]{Images/images_sequence_0/image_2.png}};

\node[circle,draw=black,fill=gray!50,inner sep=0pt,minimum size=4pt, rotate=90,
above left=.3cm and -.3 of image_0_3] (univariate-0) {};
\node[circle,draw=black,fill=gray!50,inner sep=0pt,minimum size=4pt, rotate=-90, right=.2cm of univariate-0] (univariate-1) {};
\node[circle,draw=black,fill=gray!50,inner sep=0pt,minimum size=4pt, rotate=90,  right=.2cm of univariate-1] (univariate-2) {};
\node[circle,draw=black,fill=gray!50,inner sep=0pt,minimum size=4pt, rotate=-90, right=.2cm of univariate-2] (univariate-3) {};

\node[group_embeddings, fit={(univariate-0) (univariate-3)}] (embedding_3) {};

\node[group_embeddings, fit={(univariate-0) (univariate-3)}] (embedding_1) {};
    \node[above right = .2cm and .6cm of image_0_0] (image_0_4) {\includegraphics[width=.05\textwidth]{Images/images_sequence_0/image_10.png}};
    \node[circle,draw=black,fill=gray!50,inner sep=0pt,minimum size=4pt, rotate=90,
above left=.3cm and -.3 of image_0_4] (univariate-0) {};
\node[circle,draw=black,fill=gray!50,inner sep=0pt,minimum size=4pt, rotate=-90, right=.2cm of univariate-0] (univariate-1) {};
\node[circle,draw=black,fill=gray!50,inner sep=0pt,minimum size=4pt, rotate=90,  right=.2cm of univariate-1] (univariate-2) {};
\node[circle,draw=black,fill=gray!50,inner sep=0pt,minimum size=4pt, rotate=-90, right=.2cm of univariate-2] (univariate-3) {};

\node[group_embeddings, fit={(univariate-0) (univariate-3)}] (embedding_2) {};

\node[circle, draw, dashed, minimum width=5cm, above=of image_0_1, yshift=-2.5cm, xshift=.4cm, label=below:\Large{$t_{1}$}] (set_0) {};

    \node[draw, rectangle, fill=AntiqueWhite!75, minimum height=.75cm,  minimum width=5.8cm, above=3.5cm of $(image_0_0)!0.5!(image_0_3)$] (set_transformer_0) {\Large{\large{\textbf{Set Representation}}}};

    \draw[-latex] (set_0.north) -- (set_0.north|-set_transformer_0.south);

    \node[above=.0cm of set_transformer_0] (transformer_0_0) {\Huge{+}};

    \node[draw, rectangle, fill=LightSteelBlue!50, minimum height=.75cm,  minimum width=5.8cm, above=.1cm of transformer_0_0] (transformer_0_1) {\large{\textbf{Positional Encoding}}};

    \node[above=.0cm of transformer_0_1] (transformer_0_2) {\Huge{+}};

    \node[draw, rectangle, fill=DarkSeaGreen!50, minimum height=.75cm,  minimum width=5.8cm, above=.1cm of transformer_0_2] (transformer_0_3) {\large{\textbf{Temporal Embedding}}};
    
    \node[right = 8cmof image_0_0] (image_0_4) {\includegraphics[width=.05\textwidth]{Images/images_sequence_0/image_4.png}};
    \node[circle,draw=black,fill=gray!50,inner sep=0pt,minimum size=4pt, rotate=90,
above left=.3cm and -.3 of image_0_4] (univariate-0) {};
\node[circle,draw=black,fill=gray!50,inner sep=0pt,minimum size=4pt, rotate=-90, right=.2cm of univariate-0] (univariate-1) {};
\node[circle,draw=black,fill=gray!50,inner sep=0pt,minimum size=4pt, rotate=90,  right=.2cm of univariate-1] (univariate-2) {};
\node[circle,draw=black,fill=gray!50,inner sep=0pt,minimum size=4pt, rotate=-90, right=.2cm of univariate-2] (univariate-3) {};

\node[group_embeddings, fit={(univariate-0) (univariate-3)}] (embedding_1) {};

    \node[below right = .2cmof image_0_4] (image_0_5) {\includegraphics[width=.05\textwidth]{Images/images_sequence_0/image_5.png}};
    \node[circle,draw=black,fill=gray!50,inner sep=0pt,minimum size=4pt, rotate=90,
above left=.3cm and -.3 of image_0_5] (univariate-0) {};
\node[circle,draw=black,fill=gray!50,inner sep=0pt,minimum size=4pt, rotate=-90, right=.2cm of univariate-0] (univariate-1) {};
\node[circle,draw=black,fill=gray!50,inner sep=0pt,minimum size=4pt, rotate=90,  right=.2cm of univariate-1] (univariate-2) {};
\node[circle,draw=black,fill=gray!50,inner sep=0pt,minimum size=4pt, rotate=-90, right=.2cm of univariate-2] (univariate-3) {};

\node[group_embeddings, fit={(univariate-0) (univariate-3)}] (embedding_1) {};

    \node[above right = .2cmof image_0_5] (image_0_6) {\LARGE{$\cdots$}};
    
    \node[right = .2cmof image_0_6] (image_0_7) {\includegraphics[width=.05\textwidth]{Images/images_sequence_0/image_6.png}};
    \node[circle,draw=black,fill=gray!50,inner sep=0pt,minimum size=4pt, rotate=90,
above left=.3cm and -.3 of image_0_7] (univariate-0) {};
\node[circle,draw=black,fill=gray!50,inner sep=0pt,minimum size=4pt, rotate=-90, right=.2cm of univariate-0] (univariate-1) {};
\node[circle,draw=black,fill=gray!50,inner sep=0pt,minimum size=4pt, rotate=90,  right=.2cm of univariate-1] (univariate-2) {};
\node[circle,draw=black,fill=gray!50,inner sep=0pt,minimum size=4pt, rotate=-90, right=.2cm of univariate-2] (univariate-3) {};

\node[group_embeddings, fit={(univariate-0) (univariate-3)}] (embedding_1) {};

\node[group_embeddings, fit={(univariate-0) (univariate-3)}] (embedding_1) {};
    \node[above right = .2cm and .6cm of image_0_4] (image_0_11) {\includegraphics[width=.05\textwidth]{Images/images_sequence_0/image_11.png}};
    \node[circle,draw=black,fill=gray!50,inner sep=0pt,minimum size=4pt, rotate=90,
above left=.3cm and -.3 of image_0_11] (univariate-0) {};
\node[circle,draw=black,fill=gray!50,inner sep=0pt,minimum size=4pt, rotate=-90, right=.2cm of univariate-0] (univariate-1) {};
\node[circle,draw=black,fill=gray!50,inner sep=0pt,minimum size=4pt, rotate=90,  right=.2cm of univariate-1] (univariate-2) {};
\node[circle,draw=black,fill=gray!50,inner sep=0pt,minimum size=4pt, rotate=-90, right=.2cm of univariate-2] (univariate-3) {};

\node[group_embeddings, fit={(univariate-0) (univariate-3)}] (embedding_1) {};

\node[circle, draw, dashed, minimum width=5cm, above=of image_0_5, yshift=-2.5cm, xshift=.4cm, label=below:\Large{$t_{2}$}] (set_1) {};

    \node[draw, rectangle, fill=AntiqueWhite!75, minimum height=.75cm,  minimum width=5.8cm, above=3.5cm of $(image_0_4)!0.5!(image_0_7)$] (set_transformer_1) {\large{\textbf{Set Representation}}};

     \draw[-latex] (set_1.north) -- (set_1.north|-set_transformer_1.south);
     
    \node[above=.0cm of set_transformer_1] (transformer_1_0) {\Huge{+}};

    \node[draw, rectangle, fill=LightSteelBlue!50, minimum height=.75cm,  minimum width=5.8cm, above=.1cm of transformer_1_0] (transformer_1_1) {\large{\textbf{Positional Encoding}}};

    \node[above=.0cm of transformer_1_1] (transformer_1_2) {\Huge{+}};

    \node[draw, rectangle, fill=DarkSeaGreen!50, minimum height=.75cm,  minimum width=5.8cm, above=.1cm of transformer_1_2] (transformer_1_3) {\large{\textbf{Temporal Embedding}}};
    
    \node[right = 11.17cmof image_0_4] (image_0_8) {\includegraphics[width=.05\textwidth]{Images/images_sequence_0/image_7.png}};
     \node[circle,draw=black,fill=gray!50,inner sep=0pt,minimum size=4pt, rotate=90,
above left=.3cm and -.3 of image_0_8] (univariate-0) {};
\node[circle,draw=black,fill=gray!50,inner sep=0pt,minimum size=4pt, rotate=-90, right=.2cm of univariate-0] (univariate-1) {};
\node[circle,draw=black,fill=gray!50,inner sep=0pt,minimum size=4pt, rotate=90,  right=.2cm of univariate-1] (univariate-2) {};
\node[circle,draw=black,fill=gray!50,inner sep=0pt,minimum size=4pt, rotate=-90, right=.2cm of univariate-2] (univariate-3) {};

\node[group_embeddings, fit={(univariate-0) (univariate-3)}] (embedding_1) {};

    \node[below right = .2cmof image_0_8] (image_0_9) {\includegraphics[width=.05\textwidth]{Images/images_sequence_0/image_8.png}};
     \node[circle,draw=black,fill=gray!50,inner sep=0pt,minimum size=4pt, rotate=90,
above left=.3cm and -.3 of image_0_9] (univariate-0) {};
\node[circle,draw=black,fill=gray!50,inner sep=0pt,minimum size=4pt, rotate=-90, right=.2cm of univariate-0] (univariate-1) {};
\node[circle,draw=black,fill=gray!50,inner sep=0pt,minimum size=4pt, rotate=90,  right=.2cm of univariate-1] (univariate-2) {};
\node[circle,draw=black,fill=gray!50,inner sep=0pt,minimum size=4pt, rotate=-90, right=.2cm of univariate-2] (univariate-3) {};

\node[group_embeddings, fit={(univariate-0) (univariate-3)}] (embedding_1) {};

    \node[above right = .2cmof image_0_9] (image_0_10) {\LARGE{$\cdots$}};
    \node[right = .2cmof image_0_10] (image_0_11) {\includegraphics[width=.05\textwidth]{Images/images_sequence_0/image_9.png}};
     \node[circle,draw=black,fill=gray!50,inner sep=0pt,minimum size=4pt, rotate=90,
above left=.3cm and -.3 of image_0_11] (univariate-0) {};
\node[circle,draw=black,fill=gray!50,inner sep=0pt,minimum size=4pt, rotate=-90, right=.2cm of univariate-0] (univariate-1) {};
\node[circle,draw=black,fill=gray!50,inner sep=0pt,minimum size=4pt, rotate=90,  right=.2cm of univariate-1] (univariate-2) {};
\node[circle,draw=black,fill=gray!50,inner sep=0pt,minimum size=4pt, rotate=-90, right=.2cm of univariate-2] (univariate-3) {};

\node[group_embeddings, fit={(univariate-0) (univariate-3)}] (embedding_1) {};

    \node[above right = .2cm and .6cm of image_0_8] (image_0_12) {\includegraphics[width=.05\textwidth]{Images/images_sequence_0/image_12.png}};
    \node[circle,draw=black,fill=gray!50,inner sep=0pt,minimum size=4pt, rotate=90,
above left=.3cm and -.3 of image_0_12] (univariate-0) {};
\node[circle,draw=black,fill=gray!50,inner sep=0pt,minimum size=4pt, rotate=-90, right=.2cm of univariate-0] (univariate-1) {};
\node[circle,draw=black,fill=gray!50,inner sep=0pt,minimum size=4pt, rotate=90,  right=.2cm of univariate-1] (univariate-2) {};
\node[circle,draw=black,fill=gray!50,inner sep=0pt,minimum size=4pt, rotate=-90, right=.2cm of univariate-2] (univariate-3) {};

\node[group_embeddings, fit={(univariate-0) (univariate-3)}] (embedding_1) {};

\node[circle, draw, dashed, minimum width=5cm, above=of image_0_9, yshift=-2.5cm, xshift=.4cm, label=below:\Large{$t_{N}$}] (set_2) {};

     \node[draw, rectangle, fill=AntiqueWhite!75, minimum height=.75cm,  minimum width=5.8cm, above=3.5cm of $(image_0_8)!0.5!(image_0_11)$] (set_transformer_2) {\large{\textbf{Set Representation}}};

      \draw[-latex] (set_2.north) -- (set_2.north|-set_transformer_2.south);

    \node[above=.0cm of set_transformer_2] (transformer_2_0) {\Huge{+}};

    \node[draw, rectangle, fill=LightSteelBlue!50, minimum height=.75cm,  minimum width=5.8cm, above=.1cm of transformer_2_0] (transformer_2_1) {\large{\textbf{Positional Encoding}}};

    \node[above=.0cm of transformer_2_1] (transformer_2_2) {\Huge{+}};

    \node[draw, rectangle, fill=DarkSeaGreen!50, minimum height=.75cm,  minimum width=5.8cm, above=.1cm of transformer_2_2] (transformer_2_3) {\large{\textbf{Temporal Embedding}}};

    \node (ellipsis_0) at ($(set_1)!0.5!(set_2)$) {\Huge $\cdots$};

    \node (ellipsis_1) at ($(set_transformer_1)!0.5!(set_transformer_2)$) {\Huge $\cdots$};

    \node (ellipsis_2) at ($(transformer_1_1)!0.5!(transformer_2_1)$) {\Huge $\cdots$};

    \node (ellipsis_3) at ($(transformer_1_3)!0.5!(transformer_2_3)$) {\Huge $\cdots$};

    \node[draw, rectangle, fill=Pink!75, minimum height=.75cm,  minimum width=27.2cm, above=.9cm of $(transformer_0_3)!0.5!(transformer_2_3)$] (transformer) {\large{\textbf{Transformer Encoder}}};

    \draw[-latex] (transformer_0_3.north) -- (transformer_0_3.north|-transformer.south);

    \draw[-latex] (transformer_1_3.north) -- (transformer_1_3.north|-transformer.south);

    \draw[-latex] (transformer_2_3.north) -- (transformer_2_3.north|-transformer.south);

    % \node[above=1cm of transformer, label=north:\Large{\textbf{Output}}] (transformer_2_4) {\Large{\textbf{0.9}}};

    \node[above=1.8cm of transformer_2_3] (transformer_2_4) {\Large{\textbf{$\hat{y}$}}};

    \draw[latex-] (transformer_2_4.south) -- (transformer_2_4.south|-transformer.north);

\end{tikzpicture}
\end{adjustbox}

%% file: Sections/approach.tex
\section{Set2Seq Transformer}\label{sec:approach}

We now present the Set2Seq Transformer, a novel architecture for learning temporal and position-aware representations of sequences of sets. Formally, let $\mathcal{T}$ denote an ordered sequence of $N$ sets:
\begin{equation}
\label{eq:sequence_formulation}
\mathcal{T} = \langle\mathcal{S}_1, \mathcal{S}_2, \ldots, \mathcal{S}_N\rangle,
\end{equation}
where each set $\mathcal{S}_i$ contains multiple instances observed at timestep $t_i$. Each timestep $t_i$ is associated with a relative position within the sequence and an absolute temporal value. The objective is to predict a target value $y$ associated with the entire sequence $\mathcal{T}$. To this end, the Set2Seq Transformer first learns permutation-invariant representations of individual sets and then encodes the sequence using two complementary temporal mechanisms: positional encodings representing relative sequence position and temporal embeddings representing absolute calendar-time context. These representations are processed by a Transformer encoder, adaptable to diverse tasks, such as learning-to-rank or classification. A key advantage of Set2Seq Transformer is its ability to handle variable set cardinalities and sequence lengths, making it suitable for real-world scenarios with irregular or incomplete observations. Figure~\ref{fig:set_transformer2transformer} depicts the Set2Seq Transformer architecture.

\subsection{Set Representation Learning}\label{subsec:learning_permuation_invariant_representations}

Unlike standard sequential models that operate on ordered elements, in sequential multiple-instance learning instances within a set lack meaningful order. Permutation-invariant representations are therefore essential, ensuring that the learned representation remains unchanged under any permutation of instances within the set.

\subsubsection{Learning permutation-invariant set representations} Let $\mathcal{S}$ denote an unordered set of $K$ instances:
\begin{equation} \label{eq:set_representation}
    \mathcal{S} = \{x_1, x_2, \ldots, x_K\}.
\end{equation}
The objective is to learn a permutation-invariant representation $\mathbf{s}$ for $\mathcal{S}$ such that $\mathbf{s}$ remains unchanged under any permutation of its instances. We employ DeepSets~\cite{deep_sets} and Set Transformer~\cite{set_transformer} as alternative set encoders, each followed by a fully connected layer to obtain an $D$-dimensional representation $\mathbf{s} \in \mathbb{R}^D$.

\paragraph{DeepSets} For DeepSets, we use four fully-connected layers as the encoder followed by the aggregation operation and a decoder of three fully-connected layers. The dimensionality of all hidden layers is set to 256. We consider average (mean) and maximum (max) pooling, and use the Rectified Linear Unit (ReLU) activation between layers in both the encoder and decoder.

\paragraph{Set Transformer} For Set Transformer, we consider three variants. First, stacked Set Attention Blocks (SAB) followed by Pooling by Multi-head Attention (PMA), denoted as Set Transformer (SAB + PMA). Second, stacked Induced Set Attention Blocks (ISAB) followed by PMA, denoted as Set Transformer (ISAB + PMA). Third, stacked ISAB blocks followed by PMA and additional SAB layers, denoted as Set Transformer (ISAB + PMA + SAB). The dimensionality of all hidden layers is set to 256, and the number of attention heads to 4.

\subsubsection{Visual instance representation} To extract feature vectors for a visual instance $x_j$, we utilize a frozen ResNet-34~\cite{resnet} backbone pretrained on ImageNet~\cite{imagenet}. We obtain the visual representation $\mathbf{x}_j$ as:
\begin{equation}\label{eq:visual_embedding}
    \mathbf{x}_j = F_{GAP}(f_{\text{ResNet-34}}(x_j;\theta_{\text{ResNet-34}})) \in \mathbb{R}^D,
\end{equation}
where $F_{GAP}$ denotes the global average pooling operation, and $\theta_\text{ResNet-34}$ denotes the ResNet-34 parameters.

\subsubsection{Numerical instance representation} For numerical instances (e.g., multivariate observations), each instance $x_j$ consists of a vector of numerical features, which are normalized and encoded through a fully connected layer to obtain $\mathbf{x}_j \in \mathbb{R}^D$.

\subsection{Sequential Learning}\label{subsec:learning_permuation_variant_representations}
Inspired by its success in sequence modeling, we use the Transformer~\cite{transformer} to learn permutation-variant representations across timesteps. In our setting, sequential modeling must capture both relative sequence position and absolute temporal context. To this end, we construct timestep representations by combining permutation-invariant set representations with positional encodings representing relative position and temporal embeddings representing absolute calendar time. These representations are processed by a Transformer encoder to model temporal dependencies across timesteps.

\subsubsection{Learning representation of sequences of sets}
Given the sequence $\mathcal{T}$ defined in Equation~\ref{eq:sequence_formulation}, we now describe how permutation-invariant set representations $\mathbf{s}_i$ are combined with positional encodings $\mathbf{u}_i$ and temporal embeddings $\mathbf{v}_i$ to construct the input $\boldsymbol{\tau}_i$ to the Transformer encoder.

\subsubsection{Positional encoding} Following~\cite{transformer}, we use positional encodings $\mathbf{u}_i \in \mathbb{R}^D$ to represent the relative position of timestep $t_i$ within the sequence $\mathcal{T}$. These encodings capture progression within the sequence independently of the absolute temporal values associated with each timestep. We use sinusoidal functions to encode relative positions through smooth periodic patterns that generalize to unseen sequence lengths. The positional encoding $\mathbf{u}_i$ is obtained as:
\begin{align}\label{al:positional_encoding}
    \mathbf{u}_i(i, 2d) &= \sin(i/10000^{2d/D}),\\
    \mathbf{u}_i(i, 2d+1) &= \cos(i/10000^{2d/D}),
\end{align}
where $i$ denotes the position index within $\mathcal{T}$, and $d$ the embedding dimension index. While we adopt sinusoidal encodings, the formulation is not restricted to this scheme, and any encoding or learnable embedding representing relative position can be employed.

\subsubsection{Temporal embedding} In addition to positional encodings, we learn temporal embeddings $\mathbf{v}_i \in \mathbb{R}^D$ that represent the absolute temporal value associated with the timestep $t_i$. Unlike positional encodings, which represent relative ordering within the sequence, temporal embeddings encode calendar-time context, such as specific years, dates, or timestamps when observations occur. These embeddings enable the model to capture temporal effects tied to absolute time, such as long-term trends, seasonality, or historical context. Formally, the temporal embedding $\mathbf{v}_i$ is obtained as:
\begin{equation}\label{eq:temporal_embedding}
\mathbf{v}_i = f(a_i) \in \mathbb{R}^D,
\end{equation}
where $a_i$ denotes the absolute temporal value associated with timestep $t_i$, and $f$ denotes a learnable function over $a_i$. When $a_i$ includes multiple granularities (e.g., month and year), we decompose it as $a_i=(c_i, z_i)$, where $c_i$ denotes a cyclic component (e.g., month or season) and $z_i$ a linear component (e.g., year). The temporal embedding is then computed as:
\begin{equation}\label{eq:temporal_embedding_timestamp}
\mathbf{v}_i = \big[\mathbf{v}_i^{c}, \mathbf{v}_i^{z}\big],
\end{equation}
where $\mathbf{v}_i^{c}, \mathbf{v}_i^{z} \in \mathbb{R}^{D/2}$ represent the cyclic and linear embedding components, respectively (see Section~\ref{subsec:temporal_embedding} of the supplementary material for more details). The function $f$ can be implemented using any learnable embedding or encoding scheme; we use learnable embeddings to enable flexible modeling of calendar-time effects.

\subsubsection{Final input embedding} The final embedding $\boldsymbol{\tau}_i$ integrates three complementary components: the permutation-invariant set representation $\mathbf{s}_i$, the positional encoding $\mathbf{u}_i$ representing the relative sequence position, and the temporal embedding $\mathbf{v}_i$ representing the absolute calendar-time value associated with timestep $t_i$. These components are combined via element-wise summation:

\begin{equation}\label{eq:final_embedding}
    \boldsymbol{\tau}_i = \mathbf{s}_i + \mathbf{u}_i + \mathbf{v}_i.
\end{equation}
The obtained embedding $\boldsymbol{\tau}_i \in \mathbb{R}^D$ jointly encodes set structure, sequence position, and absolute temporal context, and is provided as input to the Transformer encoder.

\subsubsection{Transformer Encoder} For our Set2Seq Transformer, we utilize a Transformer encoder module to learn representations of the sequence $\mathcal{T} \in \mathbb{R}^{N \times D}$. Throughout this work, we set the embedding dimension to $D=512$, the number of Transformer encoder layers to $L=12$, the number of attention heads to $A=16$, and the feed-forward hidden dimension to $2D$. The obtained sequence representations are then passed through two fully connected layers with hidden dimensionality $D$ to generate the final prediction $\hat{y}$, typically using the representation of the last timestep. Our framework is applicable to multiple task settings. In this work, we consider two objectives for a given sequence $\mathcal{T}$: (i) learning-to-rank, where the model predicts a real-valued relevance score, and (ii) classification, where the model predicts a class label. We use task-specific losses---Mean Squared Error (MSE) for learning-to-rank and Cross-Entropy (CE) for classification. Formal definitions of both objectives are provided in Section~\ref{subsec:objectives} of the supplementary material.

%% file: Sections/experimental_setup.tex
\section{Experimental Setup}\label{sec:experimental_setup}

We evaluate the Set2Seq Transformer on two structurally and semantically distinct tasks---predicting artistic success and short-term wildfire danger forecasting---that differ in modality, sequence length, instance cardinality, and temporal coverage. In this section, we describe the task formulations, dataset characteristics, baseline methods, and evaluation protocols for each task. Table~\ref{tab:dataset_statistics} summarizes the statistics for each dataset considered.

\subsection{Predicting Artistic Success}\label{subsec:experimental_setup_predicting_artistic_success}

We start by providing details about the artistic success prediction task using our novel WikiArt-Seq2Rank dataset. We first formalize the task of predicting artistic success as a visual learning-to-rank problem, then describe the dataset construction, the ranking criteria, train/validation/test splits, baseline methods, and evaluation metrics.

\subsubsection{Task formulation} For predicting artistic success, each sequence $\mathcal{T}$ corresponds to an individual artist’s career, where each set $\mathcal{S}_i$ contains the artworks created within a discrete timestep $t_i$ (e.g., a specific year). Each artwork is encoded as a feature vector using a ResNet-34 backbone pretrained on ImageNet~\cite{imagenet}. The task is formulated as a pointwise learning-to-rank problem, where the objective is to predict a scalar score $y$ for each artist according to a specific success criterion (Figure~\ref{fig:task}).

\subsubsection{WikiArt-Seq2Rank} For the task of predicting artistic success, we introduce WikiArt-Seq2Rank, a novel dataset that extends the publicly available WikiArt collection~\cite{www.wikiart.org} by incorporating multiple rankings of 849 renowned visual artists based on various success criteria. This dataset builds upon our recent work~\cite{brmw} that utilized a subset of the WikiArt collection and deep-learning visual representations, combined with a manually defined measure of visual originality, to examine the relationship between artistic originality and career success. The original WikiArt dataset, extensively used in automatic fine art analysis~\cite{artemis_v2, artsagenet, elgammal2018shape, tan_artgan, vlkge}, consists of 75,921 paintings created by 1,111 artists from 1401 up to 2012, spanning major stylistic movements. We omit artworks without a known creation date and artists that have fewer than 10 artworks in the collection. This results in a dataset that consists of 59,458 artworks created by 849 artists. For each artist, we collected information from various publicly available sources to construct rankings based on artist appreciation, grounded in cultural industries theory~\cite{deep_learning_based_product_distinctiveness, relations_aesthetic_space, selection_systems}.

\paragraph{Rankings} We compile artist-success rankings from various sources, following and extending the approach introduced in~\cite{brmw}, including: (1) eBooks, based on the number of mentions in a corpus of 904 art-related books; (2) The New York Times (NYT), using 4,525 abstracts of art-related reviews from 1981–2019; (3) Wikipedia Mentions, measuring the number of times an artist's name appears; (4) Wikipedia Links, measuring the number of links from other artists’ pages; (5) Wikipedia Pageviews, tracking the number of times an artist’s page was accessed; (6) Google Ngram, counting mentions of artists in books via Google Books Ngram Viewer (2019 Ed.)~\cite{google_ngram}; (7) Google Trends, monitoring search frequency in the ``painting'' category from 2017–2022; (8) Artfacts~\cite{www.artfacts.net}, which ranks artists based on exhibition data; and (9) Artprice~\cite{www.artprice.com}, which reports the top 500 artists by auction revenue annually from 2006–2021. We also construct an overall ranking, (10) Aggregate Ranking, by aggregating the individual rankings using the Borda count system~\cite{borda}. Further details are provided in Section~~\ref{sec:wikiart_seq2rank} of the supplementary material.

\paragraph{Train/Val/Test splits} We evaluate our methods using two split strategies: stratified and time series. Each sample represents an artist's career, with static methods using full sets of artworks and sequential methods using sequences. For the stratified split, the dataset is divided into training (70\%), validation (10\%), and test (20\%) subsets. The time series split assigns artists to training (pre-1930), validation (1930–1951), and test (post-1951) sets, reflecting real-world scenarios. Detailed dataset statistics and results for the time series split are provided in Section~\ref{sec:wikiart_seq2rank} of the supplementary material.

\subsubsection{Baseline methods} We compare our Set2Seq Transformer with several static and temporal methods for the task of predicting artistic success. The static baselines include: (1) Vanilla linear regression; (2) Gradient Boosting (XGBoost)~\cite{xgboost}, an ensemble method for learning-to-rank tasks; (3) DeepSets~\cite{deep_sets}, using mean and max aggregation with a final fully connected layer; and (4) Set Transformer~\cite{set_transformer}. For temporal-based baselines, we consider: (5) Hierarchical DeepSets, which first learns representations of individual sets at each timestep and then aggregates these representations across timesteps; (6) LSTM, a bidirectional LSTM~\cite{lstm} with attention~\cite{attention}, using two stacked layers and 512 hidden units; and (7) Transformer~\cite{transformer}, which flattens temporal representations and applies positional encoding for each instance. Note that the static methods do not model the temporal structure of artists' careers, treating the entire body of work as a single unordered set, while the temporal baselines explicitly model the sequence of artwork sets over time. Full implementation details are provided in Section~\ref{sec:static_temporal_aware_baselines} of the supplementary material.

\subsubsection{Evaluation metrics and leaderboard} For evaluation on WikiArt-Seq2Rank, we report Kendall’s $\tau$~\cite{kendalls_tau} and Mean Absolute Error (MAE), reflecting performance in rank prediction. Definitions of both metrics are included in Section~\ref{subsec:evaluation_metrics} of the supplementary material. To provide an overall comparison of methods across the multiple ranking criteria in WikiArt-Seq2Rank, we construct a leaderboard by aggregating performance using two complementary scoring approaches. Inspired by recent advancements in large-scale model evaluation (e.g., Chiang et al.~\cite{chatbot_arena}), we first report Bradley–Terry (BT) scores~\cite{bt_scores}, based on pairwise win–loss comparisons between methods across all metric evaluations (see Chiang et al.~\cite{chatbot_arena} for details on BT scores). Additionally, we report a simple Borda count–based score~\cite{nlp_benchmarking, adversarial_dueling_bandits}, in which methods are ranked per metric and dataset, and rank-based points are summed to produce a global ordering. The leaderboard scores (BT and Borda) are reported in Table~\ref{tab:results_random_split} alongside raw metrics.

\begin{table}[t]
\centering
\caption{Dataset statistics for WikiArt-Seq2Rank and Mesogeos.}
\label{tab:dataset_statistics}
\input Tables/dataset_statistics.tex
\end{table}

\begin{table*}[t]
\caption{Results for the predicting artistic success task using the WikiArt-Seq2Rank stratified split subset. We report Kendall’s $\tau$ and Mean Absolute Error (MAE) for different rankings, as well as overall performance in the Aggregate Ranking and leaderboard scores (BT and Borda). PE denotes utilizing positional encodings; TE denotes utilizing temporal embeddings. (mean) and (max) refer to pooling operations. For Set2Seq Transformer using Set Transformer as the set-based module, the specific variant is denoted in parentheses. For Kendall’s $\tau$, BT, and Borda, higher ($\bigtriangleup$) is better; for MAE, lower ($\bigtriangledown$) is better. Best results in \textbf{bold}; second best \underline{underlined}.}
\label{tab:results_random_split}
\input Tables/results_random_split.tex
\end{table*}

\subsection{Short-Term Wildfire Danger Forecasting}\label{subsec:experimental_setup_short_term_wildfire_danger_forecasting}

We now describe the short-term wildfire danger forecasting task using the Mesogeos dataset~\cite{mesogeos}. We first present the task formulation and dataset characteristics, then introduce an early-forecasting setting, and describe the baseline methods and evaluation metrics.

\subsubsection{Task formulation} For short-term wildfire danger forecasting, each sequence $\mathcal{T}$ corresponds to a 30-day period of fire-driving environmental observations. Each set $\mathcal{S}_i$ contains the variables observed at timestep $t_i$, or aggregated over small groups of days in multiple-instance configurations. The task is formulated as a binary classification problem, where the model predicts whether a wildfire exceeding 30 hectares occurs on the final timestep of the sequence. Following~\cite{mesogeos}, we address class imbalance using a weighted cross-entropy loss scaled by a logarithmic transformation of fire size.

\subsubsection{Mesogeos dataset} The Mesogeos dataset spans 2006--2022 and includes 25,722 samples, of which 8,574 are positive fire cases and 17,148 are negative. The data is split temporally into training (2006--2019), validation (2020), and test (2021--2022) sets. Each sample corresponds to a 30-day period and includes dynamic features (e.g., vegetation indices and weather variables) normalized over time, as well as static geographical features (e.g., topography and elevation). Additional details on dataset characteristics and preprocessing are provided in Section~\ref{sec:mesogeos_dataset} of the supplementary material.

\paragraph{Single-instance and multiple-instance learning settings} Originally structured for single-instance learning, where each sample consists of 30 daily observations, we extend Mesogeos to support multiple-instance learning via temporal grouping. Observations are aggregated into fixed-size sets, resulting in five input configurations: (a) thirty 1-day sets, (b) fifteen 2-day sets, (c) ten 3-day sets, (d) six 5-day sets, and (e) three 10-day sets. Each configuration spans the full 30-day period, and the input to the model is a sequence of sets of fire-driving variables. The objective is to predict the probability of a fire event on the final day based on the full 30-day input. Table~\ref{tab:dataset_statistics} reports statistics for Mesogeos aggregated over all five input configurations.

\paragraph{Early short-term wildfire danger forecasting task} To simulate realistic forecasting conditions, we define an early short-term wildfire danger forecasting task, where models predict wildfire occurrence at progressively earlier timesteps with limited temporal context. For each input configuration consisting of $N$ timesteps, the model receives a sequence of sets and is evaluated at each timestep $t_i$, using only input available up to that timestep. This setup enables us to assess performance under incomplete information and track how predictive accuracy evolves as more temporal context becomes available. For this task, we use the Set2Seq Transformer with the Set Transformer (ISAB + PMA) module as the set encoder.

\subsubsection{Baseline methods} We compare our Set2Seq Transformer against several baselines that reflect different modeling assumptions. For set-based methods, we include: (1) DeepSets~\cite{deep_sets}, with max pooling; and (2) Set Transformer~\cite{set_transformer}, using ISAB and PMA modules. For temporal methods, we consider: (3) a bidirectional LSTM~\cite{lstm, attention} with two stacked layers and 512 hidden units; and (4) a Transformer encoder~\cite{transformer}, which flattens instance-level representations over time and applies positional encodings. For the Set2Seq Transformer, we report results using two variants: one with DeepSets (max pooling) and one with Set Transformer (ISAB + PMA) as the set encoder. A full training epoch of the Set2Seq Transformer takes approximately 1--2 minutes with approximately 27M trainable parameters. All experiments are implemented in PyTorch~\cite{pytorch} and conducted on a single NVIDIA A100 GPU.

\subsubsection{Evaluation metrics} We report Precision–Recall Area Under the Curve (PR-AUC)~\cite{roc_analysis} and F1-score, both commonly used in high-risk event forecasting. Detailed definitions of both metrics are included in Section~\ref{subsec:evaluation_metrics} of the supplementary material.

\begin{table*}[t]
\caption{F1-Score and Precision-Recall Area Under Curve (PR-AUC) for the short-term wildfire danger forecasting task using Mesogeos~\cite{mesogeos}. PE denotes positional encodings. TE denotes temporal embeddings. $K$ denotes the number of instances per set $\mathcal{S}$. $N$ denotes the number of timesteps in a sequence $\mathcal{T}$. For both metrics, higher is better. Best results in \textbf{bold}; second best \underline{underlined}.}
\label{tab:mesogeos_results}
\input Tables/mesogeos_results.tex
\end{table*}

%% file: Tables/dataset_statistics.tex
\begin{tabular*}{\columnwidth}{@{\extracolsep{\fill}}l
  c c @{}}
\toprule
Statistic & WikiArt-Seq2Rank & Mesogeos \\
\midrule
Number of sequences              & 849      & 25,916 \\
Number of instances              & 59,458   & 777,480 \\
Number of timestamps      & 581      & 6,026  \\
\midrule
Sequence length range (min--max) & 1--77    & 3--30  \\
Average sequence length             & 18.87    & 12.80  \\
\midrule
Set size range (min--max) & 1--326   & 1--10  \\
Average set size             & 3.71     & 4.20   \\
\bottomrule
\end{tabular*}

%% file: Tables/results_random_split.tex
\resizebox{\textwidth}{!}{\begin{tabular}{@{}l >{\columncolor[gray]{.92}}c>{\columncolor[gray]{.92}}c>{\columncolor[gray]{.92}}c>{\columncolor[gray]{.92}}ccccccccccccccccccc@{}}
% \toprule Method & \multicolumn{2}{>{\columncolor[gray]{.92}}c}{\thead{\phantom{}\\\quad\quad Learderboard\quad\quad}} & \multicolumn{2}{>{\columncolor[gray]{.92}}c}{\thead{ Overall\quad\quad\\ \quad\quad Aggregate\quad\quad\quad\quad}}    & \multicolumn{2}{c}{\thead{\phantom{}\\eBooks}} & \multicolumn{2}{c}{\thead{The New\\York Times}} & \multicolumn{2}{c}{\thead{Wikipedia\\Mentions}} & \multicolumn{2}{c}{\thead{Wikipedia\\Links}} & \multicolumn{2}{c}{\thead{Wikipedia\\Pageviews}} & \multicolumn{2}{c}{\thead{Google\\Ngram}} & \multicolumn{2}{c}{\thead{Google\\Trends}} & \multicolumn{2}{c}{\thead{\phantom{}\\Artfacts}} & \multicolumn{2}{c}{\thead{\phantom{}\\Artprice}}       \\
\toprule Method & \multicolumn{2}{c}{\thead{\phantom{}\\Learderboard}} & \multicolumn{2}{c}{\thead{Aggregate\\Ranking}}    & \multicolumn{2}{c}{\thead{\phantom{}\\eBooks}} & \multicolumn{2}{c}{\thead{The New\\York Times}} & \multicolumn{2}{c}{\thead{Wikipedia\\Mentions}} & \multicolumn{2}{c}{\thead{Wikipedia\\Links}} & \multicolumn{2}{c}{\thead{Wikipedia\\Pageviews}} & \multicolumn{2}{c}{\thead{Google\\Ngram}} & \multicolumn{2}{c}{\thead{Google\\Trends}} & \multicolumn{2}{c}{\thead{\phantom{}\\Artfacts}} & \multicolumn{2}{c}{\thead{\phantom{}\\Artprice}}       \\
\cmidrule(lr){2-3}
\cmidrule(lr){4-5}
\cmidrule(lr){6-7}
\cmidrule(lr){8-9}
\cmidrule(lr){10-11}
\cmidrule(lr){12-13}
\cmidrule(lr){14-15}
\cmidrule(lr){16-17}
\cmidrule(lr){18-19}
\cmidrule(l){20-21}
\cmidrule(l){22-23}
& \cellcolor{white} BT $\bigtriangleup$ & \cellcolor{white} Borda $\bigtriangleup$ & \cellcolor{white} $\tau$ $\bigtriangleup$ & \cellcolor{white} MAE $\bigtriangledown$ & $\tau$ $\bigtriangleup$ & MAE $\bigtriangledown$ & $\tau$ $\bigtriangleup$ & MAE $\bigtriangledown$ & $\tau$ $\bigtriangleup$ & MAE $\bigtriangledown$ & $\tau$ $\bigtriangleup$ & MAE $\bigtriangledown$ & $\tau$ $\bigtriangleup$ & MAE $\bigtriangledown$ & $\tau$ $\bigtriangleup$ & MAE $\bigtriangledown$ & $\tau$ $\bigtriangleup$ & MAE $\bigtriangledown$ & $\tau$ $\bigtriangleup$ & MAE $\bigtriangledown$ & $\tau$ $\bigtriangleup$ & MAE $\bigtriangledown$ \\ \midrule  
Vanilla (mean) & 436 & 25 & 0.098 & 0.734 & 0.148 & 0.777 & 0.076 & 0.510 & 0.013 & 0.444 & 0.037 & 0.464 & 0.132 & 0.686 & 0.151 & 0.681 & 0.083 & 0.740 & 0.090 & 0.614 & 0.077 & 0.919 \\
Vanilla (max) & 703 & 43 & 0.146 & 0.520 & 0.038 & 0.469 & 0.075 & 0.337 & 0.094 & 0.284 & 0.050 & 0.290 & 0.121 & 0.466 & 0.052 & 0.653 & 0.205 & 0.527 & 0.236 & 0.434 & 0.148 & 0.515 \\
Gradient Boosting (mean) & 1,046 & 83 & 0.245 & 0.229 & 0.216 & 0.227 & 0.125 & 0.129 & 0.055 & 0.113 & 0.075 & 0.115 & 0.145 & 0.261 & 0.146 & 0.283 & 0.230 & 0.235 & 0.450 & 0.195 & 0.112 & 0.252 \\
Gradient Boosting (max) & 1,407 & 195 & 0.281 & 0.215 & 0.306 & 0.209 & 0.290 & 0.093 & 0.216 & 0.088 & 0.234 & 0.096 & 0.315 & 0.221 & 0.183 & 0.269 & 0.316 & 0.205 & 0.421 & 0.208 & 0.252 & 0.219 \\\midrule
DeepSets (mean) & 1,296 & 152 & 0.290 & 0.223 & 0.308 & 0.205 & 0.125 & 0.100 & 0.123 & 0.095 & 0.201 & 0.088 & 0.177 & 0.249 & 0.249 & 0.263 & 0.120 & 0.244 & 0.456 & 0.190 & 0.119 & 0.219 \\
DeepSets (max) & 1,837 & 419 & 0.411 & 0.189 & 0.316 & 0.179 & \textbf{0.323} & \underline{0.083} & 0.262 & 0.080 & 0.288 & \textbf{0.076} & \textbf{0.382} & \textbf{0.205} & 0.296 & 0.251 & \textbf{0.399} & \textbf{0.178} & 0.423 & 0.180 & 0.230 & \underline{0.184} \\
Set Transformer (SAB + PMA) & 1,404 & 195 & 0.374 & 0.206 & 0.310 & 0.183 & 0.162 & 0.089 & 0.184 & 0.083 & 0.202 & 0.098 & 0.272 & 0.242 & 0.214 & 0.272 & 0.285 & 0.208 & 0.418 & 0.199 & 0.227 & 0.205 \\
Set Transformer (ISAB + PMA) & 1,346 & 169 & 0.314 & 0.212 & 0.299 & 0.203 & 0.162 & 0.088 & 0.187 & 0.081 & 0.198 & 0.091 & 0.273 & 0.232 & 0.213 & 0.267 & 0.233 & 0.219 & 0.441 & 0.196 & 0.105 & 0.245 \\
Set Transformer (ISAB + PMA + SAB) & 1,293 & 149 & 0.314 & 0.216 & 0.199 & 0.237 & 0.091 & 0.112 & 0.213 & 0.081 & 0.235 & 0.087 & 0.263 & 0.235 & 0.168 & 0.285 & 0.372 & 0.186 & 0.416 & 0.204 & 0.095 & 0.229 \\\midrule
Hierarchical DeepSets (max) & 1,547 & 266 & 0.378 & 0.201 & 0.298 & 0.215 & 0.275 & 0.099 & 0.223 & 0.088 & 0.211 & 0.089 & 0.307 & 0.222 & 0.265 & 0.259 & 0.340 & 0.200 & 0.478 & 0.193 & 0.263 & 0.215 \\
LSTM (mean) & 1,574 & 282 & 0.387 & 0.194 & 0.323 & 0.197 & 0.230 & 0.088 & 0.250 & 0.080 & 0.298 & 0.084 & 0.327 & 0.211 & 0.203 & 0.272 & 0.330 & 0.200 & 0.455 & 0.189 & 0.165 & 0.223 \\
LSTM (max) & 1,668 & 332 & 0.331 & 0.204 & \textbf{0.431} & 0.179 & 0.143 & 0.103 & 0.290 & 0.077 & 0.284 & 0.081 & 0.335 & 0.215 & 0.247 & 0.254 & 0.331 & 0.193 & 0.395 & 0.181 & \underline{0.305} & 0.189 \\
Transformer w/o PE + TE & 1,577 & 286 & 0.354 & 0.199 & 0.329 & 0.203 & 0.150 & 0.089 & 0.253 & 0.078 & 0.252 & 0.083 & 0.336 & 0.210 & 0.178 & 0.281 & 0.283 & 0.208 & \textbf{0.496} & \underline{0.174} & 0.236 & 0.211 \\
Transformer + PE & 1,619 & 304 & 0.344 & 0.207 & 0.322 & 0.205 & 0.246 & 0.089 & 0.286 & 0.077 & 0.242 & 0.080 & 0.347 & \underline{0.209} & 0.203 & 0.279 & 0.307 & 0.200 & 0.474 & \textbf{0.173} & 0.249 & 0.210 \\
Transformer + TE & 1,516 & 248 & 0.281 & 0.221 & 0.333 & 0.200 & 0.215 & 0.088 & 0.279 & 0.078 & 0.287 & \underline{0.079} & 0.244 & 0.247 & 0.177 & 0.282 & 0.378 & 0.183 & 0.441 & 0.179 & 0.083 & 0.215 \\
Transformer + PE + TE & 1,752 & 378 & 0.346 & 0.169 & 0.386 & 0.189 & 0.265 & 0.088 & \textbf{0.303} & \underline{0.073} & \textbf{0.336} & \textbf{0.076} & 0.323 & 0.220 & 0.201 & 0.272 & 0.339 & 0.181 & 0.478 & 0.181 & \textbf{0.314} & 0.204 \\\midrule
Set2Seq LSTM (DeepSets - mean) & 1,604 & 298 & 0.204 & 0.237 & 0.322 & 0.178 & 0.208 & \underline{0.083} & 0.284 & 0.075 & 0.250 & 0.080 & 0.308 & 0.224 & 0.242 & 0.267 & 0.346 & 0.184 & 0.451 & 0.182 & 0.204 & 0.208 \\
Set2Seq LSTM  (DeepSets - max) & 1,716 & 361 & 0.319 & 0.213 & 0.296 & 0.186 & 0.304 & \textbf{0.082} & 0.296 & 0.074 & 0.320 & \underline{0.079} & 0.318 & 0.228 & 0.255 & 0.262 & 0.352 & \underline{0.180} & 0.420 & 0.193 & 0.268 & 0.188 \\
Set2Seq LSTM  (SAB + PMA) & 1,718 & 359 & 0.294 & 0.213 & 0.387 & \underline{0.175} & 0.281 & 0.085 & 0.261 & 0.075 & \underline{0.324} & 0.082 & 0.304 & 0.231 & \textbf{0.314} & \textbf{0.239} & 0.352 & 0.185 & 0.419 & 0.186 & 0.290 & 0.194 \\
Set2Seq LSTM  (ISAB + PMA) & 1,609 & 306 & 0.189 & 0.234 & 0.389 & \underline{0.175} & \underline{0.312} & 0.085 & 0.252 & 0.075 & 0.287 & 0.086 & 0.281 & 0.239 & 0.257 & 0.254 & 0.293 & 0.200 & 0.459 & 0.185 & 0.226 & 0.191 \\
Set2Seq LSTM  (ISAB + PMA + SAB) & 1,542 & 267 & 0.295 & 0.217 & 0.338 & 0.184 & 0.229 & 0.089 & \underline{0.301} & 0.074 & 0.165 & 0.088 & 0.235 & 0.241 & 0.290 & \underline{0.250} & 0.319 & 0.208 & 0.457 & 0.193 & 0.211 & 0.213 \\
Set2Seq Transformer  (DeepSets - mean) & 1,711 & 360 & 0.376 & 0.203 & 0.362 & 0.185 & 0.234 & 0.090 & 0.267 & 0.074 & 0.314 & \underline{0.079} & 0.275 & 0.250 & 0.281 & 0.260 & 0.367 & 0.192 & 0.462 & \underline{0.174} & 0.261 & 0.191 \\
Set2Seq Transformer (DeepSets - max) & \underline{1,844} & \textbf{425} & 0.383 & 0.205 & \underline{0.411} & \underline{0.175} & 0.197 & 0.084 & 0.256 & 0.076 & 0.310 & 0.081 & \underline{0.354} & \textbf{0.205} & 0.285 & 0.251 & 0.387 & 0.182 & 0.474 & 0.176 & 0.283 & 0.189 \\
Set2Seq Transformer (SAB + PMA) & 1,782 & 398 & \textbf{0.425} & \underline{0.187} & 0.347 & 0.180 & 0.296 & 0.084 & 0.296 & 0.080 & 0.316 & 0.081 & 0.344 & 0.213 & 0.231 & 0.261 & 0.350 & 0.199 & 0.476 & 0.186 & 0.250 & \textbf{0.181} \\
Set2Seq Transformer (ISAB + PMA) & \textbf{1,857} & \underline{423} & \underline{0.414} & \textbf{0.184} & 0.406 & \textbf{0.174} & 0.277 & 0.084 & 0.288 & \textbf{0.071} & 0.310 & 0.084 & 0.288 & 0.233 & \underline{0.303} & 0.252 & \underline{0.390} & 0.184 & \underline{0.493} & 0.180 & 0.282 & 0.191 \\
Set2Seq Transformer (ISAB + PMA + SAB) & 1,598 & 297 & 0.360 & 0.202 & 0.369 & 0.198 & 0.191 & 0.086 & 0.275 & 0.080 & 0.294 & 0.080 & 0.274 & 0.231 & 0.243 & 0.263 & 0.348 & 0.197 & 0.453 & 0.190 & 0.189 & 0.212 \\

\bottomrule
\end{tabular}}

%% file: Tables/mesogeos_results.tex
\resizebox{\textwidth}{!}{\begin{tabular}{@{}lcccccccccc@{}}\toprule
\multirow{2}{*}{Method}              & \multicolumn{2}{c}{$K = 1,\ N = 30$} & \multicolumn{2}{c}{$K = 2,\ N = 15$} & \multicolumn{2}{c}{$K = 3,\ N = 10$} & \multicolumn{2}{c}{$K = 5,\ N = 6$} & \multicolumn{2}{c}{$K = 10,\ N = 3$} \\
\cmidrule(lr){2-3}
\cmidrule(lr){4-5}
\cmidrule(lr){6-7}
\cmidrule(lr){8-9}
\cmidrule(l){10-11}

                                        & F1  & PR-AUC   & F1  & PR-AUC   & F1  & PR-AUC   & F1  & PR-AUC    & F1  & PR-AUC  \\ \midrule
        DeepSets                                &             0.724 &             0.794 &             0.692 &             0.769 &             0.699 &             0.769 &             0.692 &             0.770 &             0.706 &               0.784  \\
        SetTransformer                          &             0.731 &             0.798 &             0.721 &             0.781 &             0.719 &             0.781 &             0.720 &             0.780 &             0.716 &              0.781   \\ \midrule
        LSTM                                    &             0.777 &             0.856 &             0.762 &             0.844 &             0.765 &             0.845 &             0.762 &             0.833 &             0.744 &               0.814  \\
         Transformer                           &    \textbf{0.788} &             0.851 & \underline{0.777} &             0.843 &             0.762 &             0.842 &             0.762 &             0.828 & \underline{0.754} &              0.824   \\ \midrule
            Set2Seq Transformer (Deep Sets) + PE  &             0.771 &             0.855 &             0.773 &             0.856 & \underline{0.768} &             0.846 & \underline{0.764} & \underline{0.836} &             0.749 &               0.826  \\
            Set2Seq Transformer (Deep Sets) + PE + TE & \underline{0.784} &             0.864 &             0.761 &    \textbf{0.863} & \underline{0.768} &             0.833 &             0.756 &             0.830 &             0.747 &               \underline{0.834}  \\
      Set2Seq Transformer (ISAB + PMA) + PE  &             0.776 & \underline{0.865} &             0.772 &             0.848 &             0.741 & \underline{0.851} &             0.755 &             0.832 &             0.736 &               0.816  \\
           Set2Seq Transformer (ISAB + PMA) + PE + TE & \underline{0.784} &    \textbf{0.866} &    \textbf{0.783} & \underline{0.861} &    \textbf{0.772} &    \textbf{0.856} &    \textbf{0.771} &    \textbf{0.850} &    \textbf{0.761} &               \textbf{0.835}  \\
\bottomrule

\end{tabular}}

%% file: Sections/results.tex
\section{Results}\label{sec:results}

In this section, we present the experimental results for the predicting artistic success and short-term wildfire danger forecasting tasks.

\subsection{Predicting Artistic Success}\label{subsec:predicting_artistic_success}

The main results of the evaluation of our methods using the WikiArt-Seq2Rank dataset are summarized in Table~\ref{tab:results_random_split}. For improved clarity, the second column reports the leaderboard BT and Borda scores, which summarize method comparisons across all evaluation settings. The third column shows performance on the Aggregate Ranking, constructed by aggregating the individual success rankings into a single ranking, as mentioned in Section~\ref{subsec:experimental_setup_predicting_artistic_success}.

\subsubsection{Performance of static methods} Max pooling improves most static baselines by leveraging salient features from the ResNet-34 backbone. DeepSets and Set Transformer outperform Vanilla and Gradient Boosting baselines across most settings. We have to note that DeepSets with max pooling comprise a very strong baseline performing almost on-par with both temporal-based baseline methods and our Set2Seq framework across all rankings, while achieving best results in several cases. Nevertheless, the performance of Hierarchical DeepSets is notably inferior, indicating that DeepSets can leverage the sufficient information in larger static sets, while not being able to model positional and temporal aspects.

\subsubsection{Performance of temporal methods} Table~\ref{tab:results_random_split} shows that leveraging temporal information improves performance on WikiArt-Seq2Rank, with both bidirectional LSTMs and Transformers outperforming static baselines. Similar to our static methods, maximum pooling at the set-level improves the performance of our bidirectional LSTM baselines across almost all rankings. Finally, it can be clearly seen that our Transformer baseline is the strongest temporal-based baseline method especially when combined with both positional encodings and temporal embeddings.

\subsubsection{Performance of Set2Seq Transformer} Our proposed Set2Seq Transformer outperforms almost all baselines across the different rankings and on average. It can be seen that the Set2Seq framework demonstrates improvements over set-based static methods by effectively modeling the temporal dimension. It is also worth noting that our Set2Seq Transformer generally outperforms temporal-based baselines across different settings. In particular, it achieves the best performance on the Aggregate Ranking, confirming its consistent advantage across different criteria of artistic success.

\subsubsection{Performance based on different rankings} We observe variance in the performance of our methods across different rankings. That is expected since these rankings differ based on the criteria that they incorporate. Additionally, the number of relevant artists in different rankings and other minor effects based on, e.g., ties, have a significant impact in the performance of our evaluated methods. However, our Set2Seq Transformer maintains strong results across the board, while outperforming almost all baseline methods in the Aggregate Ranking.

\subsubsection{Performance of position-aware encoding and temporal embeddings} We observe that combining position-aware encoding and temporal embeddings notably enhances Transformer performance across variants. Another important observation is that Transformers consistently outperform the Bi-LSTM methods across all different rankings. Additionally, we observe that in our Set2Seq framework Transformer-based methods outperform their bidirectional LSTM counterpart.

\begin{figure*}[t]
    \centering
    \includegraphics[width=\textwidth]{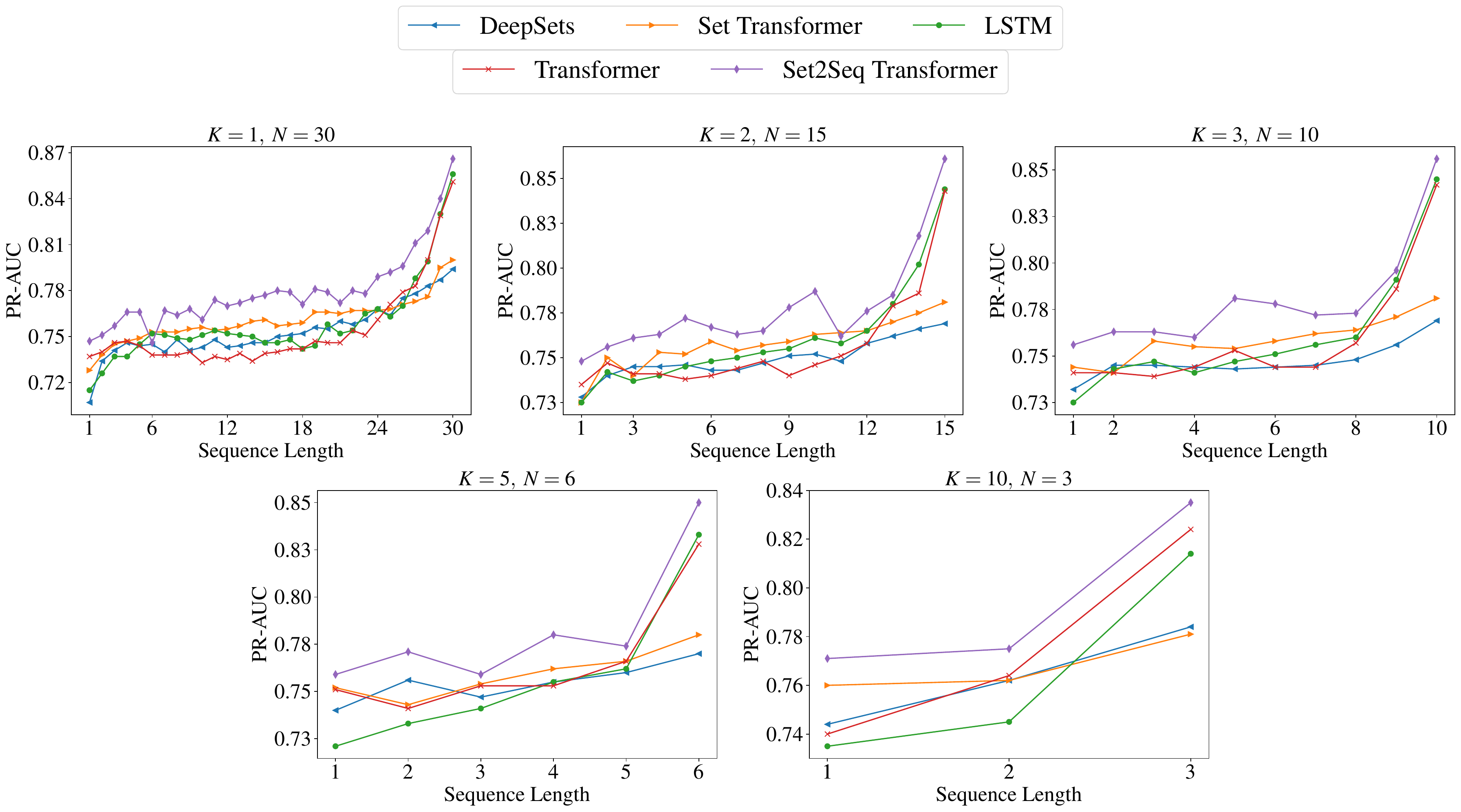}
    \caption{Performance on early short-term wildfire danger forecasting using Mesogeos~\cite{mesogeos}. Given progressively earlier timesteps and limited temporal context, the task is to predict wildfire danger with incomplete information. PR-AUC scores are reported for various models and settings across different sequence lengths.}
    \label{fig:early_wildfire_forecasting}
\end{figure*}

\subsubsection{Performance on leaderboard} Table~\ref{tab:results_random_split} also reports two aggregate leaderboard scores---the Bradley–Terry (BT) score and a Borda count–based score---which summarize performance across all rankings and evaluation metrics. The Set2Seq Transformer achieves the highest scores under both schemes, indicating consistent improvements over static and temporal baselines. Interestingly, DeepSets with max pooling also ranks highly, suggesting that the aggregated set of artworks may already contain strong expressive cues about artistic success, even without temporal modeling. However, the performance of Hierarchical DeepSets is once again consistently worse than that of all temporal-based baselines, highlighting the limitations of shallow temporal aggregation and the importance of explicitly modeling positional and sequential dependencies. Additional results for the predicting artistic success task, including results for the WikiArt-Seq2Rank time series split, using alternative visual backbones, and qualitative analysis of learned representations and attention mechanisms, are provided in Section~\ref{sec:wikiart_seq2rank} of the supplementary material.

\subsection{Short-Term Wildfire Danger Forecasting}\label{subsec:short_term_wildfire_danger_forecasting}

Table~\ref{tab:mesogeos_results} summarizes the main results of our methods on the Mesogeos dataset for short-term wildfire forecasting.

\subsubsection{Performance on sequential single-instance learning} Table~\ref{tab:mesogeos_results} presents results for sequential single-instance (i.e., $K = 1$) and multiple-instance learning (i.e., $K > 1$), alongside LSTM and Transformer baselines re-implemented to align with Mesogeos~\cite{mesogeos}. Static methods, such as DeepSets and Set Transformer, which do not account for temporality, are outperformed by sequential alternatives, with our Set2Seq Transformer consistently surpassing most baseline models. Notably, the Set2Seq Transformer performs strongly in single-instance learning, outperforming sequential methods that process raw input signals without transforming them into sets. This result indicates that our Set2Seq Transformer can effectively handle sets of single instances while maintaining high performance.

\subsubsection{Sensitivity analysis on sequential multiple-instance learning} To simulate real-world conditions with noisy or multi-day observations, we created a controlled setting by grouping consecutive days using a sliding time window. For this task, we evaluated Hierarchical DeepSets and Set Transformer baselines, which first learn set representations per timestep and then aggregate them across timesteps. As shown in Table~\ref{tab:mesogeos_results}, our Set2Seq Transformer (ISAB + PMA) utilizing position encodings and temporal embeddings, outperforms most baselines in this setting. The model adapts well to varying sequence lengths and set sizes, with positional and temporal representations consistently improving PR-AUC scores, underscoring the model’s effectiveness for short-term wildfire forecasting. Overall, our Set2Seq Transformer effectively integrates set, positional, and temporal representations, outperforming baselines and generalizing across modalities and tasks.

\subsubsection{Early short-term wildfire danger forecasting} In contrast to the standard forecasting setup in Table~\ref{tab:mesogeos_results}, which assumes the full 30-day context, Figure~\ref{fig:early_wildfire_forecasting} presents results for a proactive early forecasting task. In this setting, models predict wildfire occurrence on day 31 based on progressively increasing context, starting from day 1 and adding observations incrementally up to the 30-day window. Across the five temporal configurations, this yields a total of 64 experiments. As described in Section~\ref{subsec:experimental_setup_short_term_wildfire_danger_forecasting}, we use the Set2Seq Transformer with the Set Transformer (ISAB + PMA) module in this setup. Figure~\ref{fig:early_wildfire_forecasting} shows that our Set2Seq Transformer outperforms all baselines in 62 out of 64 cases, demonstrating its robustness under limited context and its ability to model temporal dependencies effectively.

Static set-based methods, such as DeepSets and Set Transformer, perform comparably to temporal models at early timesteps, where limited context diminishes the benefits of temporal modeling. However, unlike temporal baselines, their performance plateaus as more timesteps are added. In contrast, dynamic models---including LSTM, Transformer, and Set2Seq---exhibit consistent performance gains when later timesteps become available, highlighting their ability to exploit evolving patterns in the input sequence. Additional ablation studies, including Set Transformer variants and temporal encoding strategies, are provided in Section~\ref{sec:mesogeos_dataset} of the supplementary material.

%% file: Sections/discussion.tex
\section{Conclusion}\label{sec:conclusion}

We introduce Set2Seq Transformer, which advances sequential multiple-instance learning by jointly modeling two complementary forms of structure: permutation-invariant representations within sets and temporal position-aware relationships across sequences. Unlike prior work that handles these aspects in isolation or ignores one entirely, our framework integrates them within a unified end-to-end architecture that models relative temporal progression alongside absolute temporal context, enabling joint optimization across these dimensions. Evaluation on two semantically distinct tasks---predicting artistic success using the newly curated WikiArt-Seq2Rank dataset and short-term wildfire danger forecasting using Mesogeos---demonstrates that Set2Seq Transformer effectively captures these temporal relationships while preserving permutation-invariant set representations. Our findings underscore the importance of jointly modeling permutation-invariant set representations and temporal dynamics, an often underexplored aspect of diverse real-world applications where observations are naturally organized as unordered set-structured inputs that evolve and interact over time.

%% file: Sections/appendix.tex
\section*{Supplementary Material}\label{sec:supplementary_material}

In this supplementary material, we provide additional details to complement the main paper. In Section~\ref{sec:objectives_evaluation_metrics}, we present the formulas for the training objectives and evaluation metrics used. In Section~\ref{sec:static_temporal_aware_baselines}, we provide implementation details and visualizations of the static and temporal-aware baseline methods. In Section~\ref{sec:wikiart_seq2rank}, we describe the creation process of the WikiArt-Seq2Rank dataset in detail and include additional results. In Section~\ref{sec:mesogeos_dataset}, we present a detailed overview of the input variables and temporal embeddings used for the short-term wildfire danger task, along with ablation studies and additional results for an early short-term wildfire danger forecasting task. Finally, in Section~\ref{sec:limitations_broader_impact}, we discuss the limitations of this work and its broader impact.

\section{Loss Functions and Evaluation Metrics}\label{sec:objectives_evaluation_metrics}

In this section, we provide formulas for the training objectives and evaluation metrics.

\subsection{Loss Functions}\label{subsec:objectives}

We present the formulas for training our methods for the WikiArt-Seq2Rank predicting artistic success, and the Mesogeos short-term wildfire danger forecasting tasks.

\subsubsection{Pointwise Learning-to-Rank} For the WikiArt-Seq2Rank predicting artistic success task, we are adopting pointwise learning-to-rank and we optimize our methods using the Mean Squared Error (MSE) loss:
\begin{equation}\label{eq:mse}
    \mathcal{L}(Y, \hat{Y}) = \frac{1}{B} \sum_{b=1}^B (y_b - \hat{y}_b)^2,
\end{equation}
where $Y$ denotes the target output, $\hat{Y}$ the predicted output, $B$ denotes the total number of the samples within a batch, $y_b$ the ground-truth label for the sample $b$, and $\hat{y}_b$ the predicted output for the sample $b$. For all rankings, we scale the outputs to the $[0,1]$ range.

\subsubsection{Cross-Entropy Loss} For the Mesogeos~\cite{mesogeos} short-term wildfire danger forecasting task, we treat the problem as a binary classification task with two output logits, one for each class. To train the models, we optimize a weighted Cross-Entropy (CE) loss, designed to emphasize larger wildfires while maintaining adequate attention to low-danger instances. The loss is defined as:

\begin{equation}\label{eq:weighted_ce}
    \mathcal{L}(Y, \hat{Y}) = - \frac{1}{B} \sum_{b=1}^B w_b \log \hat{y}_{b,y_b},
\end{equation}
where $Y$ denotes the target labels, $\hat{Y}$ the predicted probabilities obtained after applying the log-softmax function to the model's output logits, $B$ the total number of samples within a batch, $y_b \in \{0, 1\}$ the ground-truth label for sample $b$, and $\hat{y}_{b,y_b}$ the predicted probability for the correct class $y_b$ for sample $b$.

Following the same approach as in~\cite{mesogeos}, the weight $w_b$ for each sample is calculated based on the burned area size of the corresponding wildfire. To ensure that larger fires contribute proportionally more to the training objective, the CE loss is weighted using the logarithmic transformation of the burned area size:

\begin{equation}\label{eq:weight_term}
w_b =
\begin{cases}
\log(\text{burned\_area}_b + 1), & \text{if } y_b = 1, \\
\min_{b':\,y_{b'}=1} \log(\text{burned\_area}_{b'} + 1), & \text{if } y_b = 0,
\end{cases}
\end{equation}
where $\text{burned\_area}_b$ is the burned area size for sample $b$. This ensures that negative samples retain a non-zero weight while preserving the relative importance of larger fires. The logarithmic transformation mitigates the risk of larger fires dominating the learning process by compressing the range of burned area values, as described in~\cite{mesogeos}.

\begin{figure*}[t]
\centering
\input{Figures/static_temporal_aware_baselines}
\caption{Visualization of three different static and temporal-aware baselines. $\hat{y}$ denotes the predicted output. (a) Static Baselines illustrates the procedure of our static baselines. Given an unordered set $\mathcal{S}$, first we extract representations of the instances $x_j$ in the set $\mathcal{S}$, thereafter we aggregate them utilizing a merging operation and finally we feed the representation forward to a regression or classification model. (b) Recurrent Neural Networks (RNNs) illustrates the procedure of the RNN-based baselines. Given an ordered input sequence $\mathcal{T}$, we aggregate the instances within a set $\mathcal{S}_i$ at timestep $t_i$ and we feed the sequential-structured representations forward to an RNN-based temporal-aware baseline. (c) Transformer with Positional Encoding (PE) and Temporal Embedding (TE) illustrates the procedure of the Transformer-based baselines. Given an ordered input sequence $\mathcal{T}$, we first extract representations of the different instances within a set $\mathcal{S}_i$ at timestep $t_i$, and thereafter we combine them with the positional encoding and temporal embedding of timestep $t_i$. Finally, we feed the obtained representation forward to a Transformer encoder module.}
\label{fig:static_temporal_aware_baselines}
\end{figure*}

\subsection{Evaluation Metrics}\label{subsec:evaluation_metrics}

Here, we detail the evaluation metrics for the WikiArt-Seq2Rank predicting artistic success, and the Mesogeos short-term wildfire danger forecasting tasks.

\subsubsection{Evaluation metrics for predicting artistic success} During inference, we process each artist individually by setting the batch size equal to 1 to alleviate issues arising from the varying size of sets and length of sequences. Inspired by research in information ordering~\cite{reorder_bart, learning_to_order_sentences, automatic_evaluation_information_ordering, sentence_ordering_rnn, han_sentence_ordering}, we report Kendall's $\tau$~\cite{kendalls_tau} and Mean Absolute Error (MAE)~\cite{age_estimation_mae} to evaluate the performance of the proposed methods.

\paragraph{Kendall's $\tau$} We compute Kendall's $\tau$ as follows:
\begin{equation}\label{al:kendalls_tau}
\tau(Y, \hat{Y}) = 1 - \frac{2(\text{\# inversions})}{B(B-1)/2},
\end{equation}
where $B$ denotes the total number of samples and \# number of inversions of consecutive items in the predicted ranking $\hat{Y}$ necessary to arrange them to match the target ranking $Y$.

\paragraph{Mean Absolute Error (MAE)} We compute the Mean Absolute Error (MAE) as follows: 
\begin{equation}\label{al:mae}
MAE(Y, \hat{Y}) = \frac{1}{B} \sum_{b=1}^B |y_b - \hat{y}_b|,
\end{equation}
where $Y$ denotes the ground-truth and $\hat{Y}$ the predicted ranking as in Equation~\eqref{eq:mse}.

\subsubsection{Evaluation metrics for short-term wildfire danger forecasting} Following~\cite{mesogeos}, we report the F1-score and Precision-Recall Area Under the Curve (PR-AUC) to evaluate the performance of the proposed methods.

\paragraph{F1-score} The F1-score provides a balance between precision and recall, capturing overall performance in binary classification tasks. It is computed as follows:
\begin{equation}\label{al:f1_score}
F1-score = 2 \cdot \frac{\text{Precision} \cdot \text{Recall}}{\text{Precision} + \text{Recall}},
\end{equation}
where Precision is defined as:
\begin{equation}\label{al:precision}
\text{Precision} = \frac{\text{True Positives}}{\text{True Positives} + \text{False Positives}},
\end{equation}
and Recall is defined as:
\begin{equation}\label{al:recall}
\text{Recall} = \frac{\text{True Positives}}{\text{True Positives} + \text{False Negatives}}.
\end{equation}

\paragraph{Precision-Recall Area Under the Curve (PR-AUC)} The PR-AUC measures the relationship between precision and recall across varying thresholds. It is defined as:
\begin{equation}\label{al:pr_auc_integral}
PR\text{-}AUC = \int_{0}^{1} \text{Precision}(\text{Recall}) \, d(\text{Recall}),
\end{equation}
where $\text{Precision}(\text{Recall})$ represents precision as a function of recall. In practice, this integral is approximated as:
\begin{equation}\label{al:pr_auc_abbrev}
PR\text{-}AUC = \sum_{r=1}^{R-1} (\text{Recall}_{r+1} - \text{Recall}_r) \cdot \text{AvgPrecision}_r,
\end{equation}
with $\text{AvgPrecision}_r$ as the average precision at successive recall thresholds.

\section{Static and Temporal-Aware Baselines} \label{sec:static_temporal_aware_baselines}

Figure~\ref{fig:static_temporal_aware_baselines} illustrates three static and temporal-aware baseline methods considered in this work.

\subsubsection{Static baselines} Figure~\ref{fig:static_temporal_aware_baselines}(a) illustrates the procedure of our static baselines. First, we extract feature vectors to represent each instance $x_j$ for a set $\mathcal{S}$ of unordered instances $x$. We aggregate the feature vectors of all instances in $\mathcal{S}$ to obtain its representation. Finally, we use a regression or classification model to obtain the predicted output $\hat{y}$ for the set $\mathcal{S}$. Note that for the WikiArt-Seq2Rank task, for our static baselines each artist in our corpus correspond to one unordered set $\mathcal{S}$ of visual instances. We consider the following static baselines:
\begin{itemize}
\item Vanilla: For the WikiArt-Seq2Rank task, we use a plain linear regression-based method that uses simple aggregation operations over all the available instances, artworks, associated to a focal artist to obtain its representation.
\item Gradient Boosting: For the WikiArt-Seq2Rank task, we use a Gradient Boosting an ensemble method with strong performance on different learning-to-rank tasks. Similar to our Vanilla baseline, we use aggregation operations to obtain an artist representation. For implementation we use the XGBoost~library~\cite{xgboost}.
\item DeepSets~\cite{deep_sets}: Static DeepSets employing mean and max aggregation operations as described in Section~\ref{sec:approach} of the main paper, and adding a final fully connected-layer to obtain the final output. For DeepSets, we are using the implementation provided in~\cite{set_transformer}.
\item Set Transformer~\cite{set_transformer}: Static Set Transformer using three different variants as described in Section~\ref{sec:approach} of the main paper, using the implementation provided in~\cite{set_transformer}. For each variant, we add a final fully connected-layer to obtain the final output.
\end{itemize}

\subsubsection{Temporal-aware baselines} Figure~\ref{fig:static_temporal_aware_baselines}(b-c) illustrates the procedure of two temporal-aware baseline methods.

\paragraph{Recurrent Neural Networks (RNNs)} Figure~\ref{fig:static_temporal_aware_baselines}(b) illustrates the procedure of our temporal-aware RNN-based baseline. Given an ordered input sequence $\mathcal{T}$ of sets $\mathcal{S}$, we first aggregate the feature vectors of instances observed at the same timestep $t_i$ to obtain a set representation $\mathbf{s}_i$ for the timestep $t_i$. Thereafter, we feed the sequential-structured representations forward to an RNN-based module. In this work, we utilize Long Short Term Memory networks (LSTMs)~\cite{lstm} as our RNN-based module. Specifically, we employ a bidirectional LSTM coupled with attention-mechanism as described in Section~\ref{sec:experimental_setup} of the main paper.

\paragraph{Transformer} Figure~\ref{fig:static_temporal_aware_baselines}(c) illustrates the procedure of our temporal-aware Transformer-based baseline. Given an ordered input sequence $\mathcal{T}$ of sets $\mathcal{S}$, we first extract the feature vector representations $\mathbf{x}_{ij}$ of the instances observed at a timestep $t_i$. We combine the feature vector representations for each instance at timestep $t_i$ with the positional encoding $\mathbf{u}_i$ and temporal embedding $\mathbf{v}_i$ of timestep $t_i$. Finally, the obtained representation, i.e., the sum of the feature vector $\mathbf{s}_i$, positional encoding $\mathbf{u}_i$ and temporal embedding $\mathbf{v}_i$ is fed forward to a Transformer encoder. In this work, we utilize a Transformer encoder module~\cite{transformer} as described in Section~\ref{sec:experimental_setup} of the main paper.

For our temporal-based baselines, we consider the following:
\begin{itemize}
\item Hierarchical DeepSets: An extension of DeepSets for temporal-aware learning of set representations, i.e., we learn a set representation for each timestep $t_i$ for an ordered sequence $\mathcal{T}$, and a set representation for the sequence $\mathcal{T}$.
\item BiLSTM + ATTN: A bidirectional LSTM~\cite{lstm} method coupled with attention mechanism~\cite{attention} that learns sequential representations over aggregates of sets. We use two stacked layers and set the hidden dimensionality to 512.
\item Transformer: A Transformer encoder module~\cite{transformer} as described in Section~\ref{sec:approach} of the main paper with the modification of flattening the temporal representations over the sequence of sets, i.e., all instances in the set $\mathcal{S}_i$ at timestep $t_i$ have a $i^{th}$ indexed positional encoding and a $i^{th}$ indexed temporal embedding, but their own representation $\mathbf{x}_{ij}$.
\end{itemize}

\begin{table}[t]
\centering
\caption{Statistics of the WikiArt-Seq2Rank dataset.}
\label{tab:dataset_collection_properties}
\input Tables/dataset_collection_properties.tex
\end{table}

\begin{table*}[t]
\caption{Statistics on rankings creation and different splitting strategies. For dataset splits, the number of artists and paintings (in parentheses) are given. Ranked denotes the number of artists ranked per individual ranking (out of the 849 artists in our corpus). Ties denotes the number of rank positions that consist of more than one artist. Range denotes the minimum and maximum ranked positions (all not-ranked artists are tied in the last position). Time Period denotes the time period in which the different source provides information regarding the ranked artists. Aggregate denotes the final aggregate operation for constructing a single ranking per source. time series split row show the number of artists and paintings (in parentheses) for all rankings considered. N/A denotes Not Applicable.}
\label{tab:rankings_statistics}
\input Tables/rankings_statistics.tex
\end{table*}

\begin{table*}[t]
\caption{Top 10 artists per ranking (in ascending order).}
\label{tab:rankings_top_10}
\input Tables/top_10_artists_per_ranking.tex
\end{table*}

\section{WikiArt-Seq2Rank} \label{sec:wikiart_seq2rank}

Here, we describe in full detail how our WikiArt-Seq2Rank dataset was created.

\subsection{Dataset Statistics} \label{subsec:dataset_statistics}

We make use of the WikiArt collection~\cite{www.wikiart.org}, which has been extensively studied in automatic fine art analysis~\cite{artemis_v1, cetinic_dl_art, artsagenet, vlkge, elgammal2018shape, artemis_v2, tan_artgan, learning_to_rank_for_refining_image_retrieval}. Following~\cite{artsagenet, elgammal2018shape}, we use the publicly available version of the WikiArt dataset, which consists of 75,921 paintings created by 1,111 artists from 1401 to 2012, spanning 20 stylistic movements ranging from Early Renaissance to contemporary art. We omit artworks without a known creation date and artists with fewer than 10 artworks. This results in a dataset of 59,458 artworks created by 849 artists. Each artist is represented as a temporally ordered sequence, where artworks are grouped by creation year to form sets at each timestep. The curated dataset also provides annotations based on 4,513 unique tags associated with the artworks, as well as annotations of the most popular artworks for different artists, comprising 4,711 artworks in total. In this work, we model only the visual content and creation period of each artwork. Table~\ref{tab:dataset_collection_properties} summarizes the statistics of the WikiArt-Seq2Rank dataset. We extend the publicly available WikiArt dataset using additional information from various sources to construct artist appreciation rankings for the 849 artists. This extended dataset enables systematic evaluation of sequence-level prediction methods for artistic career modeling across multiple success indicators and evaluation protocols.

\subsection{Rankings Construction} \label{subsec:rankings}

We construct artist appreciation rankings building upon the framework established in~\cite{brmw}, which gathered data from various publicly available sources to study the relationship between visual originality and career success. While that work used these rankings for static analysis and regression modeling, we apply them to a sequential multiple-instance learning problem, organizing artworks temporally to model career progression. Following previous work in rank aggregation~\cite{models_metasearch,rank_aggregation_methods_web,rank_aggregation_effective_recommender_systems, bordarank}, we use the Borda count system~\cite{borda} to aggregate different rankings when necessary. In this work, we consider the following rankings:

\subsubsection{eBooks} \label{subsubsec:ebooks}
Following~\cite{brmw}, we use rankings based on mentions in 904 art-related digitized books gathered from four different sources. For each source, eBooks were collected in Portable Document Format (PDF). Scanned eBooks were converted to editable format using Optical Character Recognition (OCR) with Tesseract~\cite{tesseract}. For each eBook, Named Entity Recognition (NER) was performed using the spaCy~\cite{spacy} implementation of RoBERTa~\cite{roberta}, keeping all named entities belonging to the label ``PERSON''. The number of eBooks mentioning each artist in our corpus was counted to construct the ranking.

\paragraph{The Met Collection} \label{subsubsec:met_collection}
Following~\cite{brmw}, we use 601 digitized eBooks that were collected from the MetPublications publicly available service~\cite{www.metmuseum.org/met-publications}. For this source, 47\% (396 out of 894) of the artists in our corpus have at least one mention in any of the 601 available eBooks.

\paragraph{The Guggenheim Collection} \label{subsubsec:guggenheim_collection}
Following~\cite{brmw}, we use 192 digitized eBooks that were collected from the Guggenheim Publications publicly available service~\cite{www.guggenheim.org/publications}. For this source, 40\% (341 out of 894) of the artists in our corpus have at least one mention in any of the 192 available eBooks.

\paragraph{The Getty Collection} \label{subsubsec:getty_collection}
Following~\cite{brmw}, we use 70 digitized eBooks that were collected from the Getty Publications publicly available service~\cite{www.getty.edu/publications}. For this source, 32\% (270 out of 894) of the artists in our corpus have at least one mention in any of the 70 available eBooks.

\paragraph{University Library} \label{subsubsec:university_library}
Following~\cite{brmw}, we use 41 art-related eBooks that were manually collected from an online university library. For this source, 30\% (251 out of 894) of the artists in our corpus have at least one mention in any of the 41 available eBooks. Using the Borda count system, we create a final aggregate ranking, which we refer to as eBooks. In this ranking, 65\% (550 out of 849) of the artists in our dataset are mentioned at least once in any of the available eBooks of the four collections.

\subsubsection{The New York Times (NYT)} \label{subsubsec:nyt}
Following~\cite{brmw}, we use the publicly available archive of The New York Times~\cite{www.nytimes.com}, from which 4,525 total abstracts of art-related reviews published from 1981--2019 were retrieved. The same procedure as in eBooks was used to extract entities and count the number of times an artist is mentioned in any abstract. In this ranking, 47\% (401 out of 849) of the artists in our data have at least one mention in an abstract.

\subsubsection{Wikipedia Mentions} \label{subsubsec:wikipedia_mentions}
Following~\cite{brmw}, we use all available pages in the English section of Wikipedia~\cite{www.wikipedia.org} for the artists in our corpus. The same procedure as in eBooks and The New York Times rankings was used to extract entities and count the number of times an artist is mentioned in another artist's Wikipedia page. This resulted in a ranking where 86\% (732 out of 849) of the artists are mentioned at least once in another artist's Wikipedia page.

\subsubsection{Wikipedia Links} \label{subsubsec:wikipedia_links}
We constructed an additional ranking by counting the number of times an artist's Wikipedia page uniform resource locator (URL) is linked to another artist's Wikipedia page, using all available pages in the English section of Wikipedia for the artists in our corpus. This resulted in a ranking where 66\% (560 out of 849) of the artists' Wikipedia pages are linked to at least another artist's Wikipedia page.

\subsubsection{Wikipedia Pageviews} \label{subsubsec:wikipedia_pageviews}
Following~\cite{brmw}, we use the pageviews statistics as provided by Wikipedia to create a ranking by collecting artists' English Wikipedia pages that were accessed by the public between July 2015 and December 2020. A ranking was created by aggregating the number of times an artist's English Wikipedia page was accessed by an individual. This resulted in a ranking where 96\% (817 out of 849) of the artists' Wikipedia pages were accessed at least once.

\subsubsection{Google Ngram} \label{subsubsec:google_ngram}
Following~\cite{brmw}, we use the Google Books Ngram Viewer~\cite{google_ngram} on a yearly basis, counting the number of times an artist was mentioned in any book using the American English Google Books corpus (2019 version). This resulted in a ranking where 81\% (691 out of 849) of the artists in our corpus are mentioned at least once.

\subsubsection{Google Trends} \label{subsubsec:google_trends}
Following~\cite{brmw}, we use the Google Trends service~\cite{trends.google.com/trends}, utilizing the relative frequency of an artist being searched using Google Search for the Google category ``Painting'' from April 23, 2017, up to April 23, 2022. We identified 554 artists in our corpus that have a valid Google topic identifier and compared their respective frequency of being searched to the topic ``Painter''. This resulted in a ranking where 65\% (554 out of 849) of the artists in our corpus were searched by the public using the Google Search service.

\subsubsection{Artfacts} \label{subsubsec:artfacts}
Following~\cite{brmw}, we use artist-specific rankings through Artfacts, a web service~\cite{www.artfacts.net} that provides art-related rankings based on artists' exhibitions at museums and art galleries, as provided for the year 2016. This results in a ranking where 52\% (441 out of 849) of the artists in our corpus were ranked.

\subsubsection{Artprice} \label{subsubsec:artprice}
Following~\cite{brmw}, we use publicly available reports from Artprice, a web service that provides information on art auction sales~\cite{www.artprice.com}. These reports provide information about the top 500 artists by revenue on a yearly basis from 2006--2021. We use the Borda count system and aggregate the annual rankings. This resulted in a ranking where 43\% (366 out of 849) of the artists in our corpus were ranked.

\subsubsection{Aggregate Ranking} \label{subsubsec:aggregate_ranking}
Further to the individual rankings, we created an overall ranking by aggregating them using the Borda count system. This resulted in a ranking where 98\% (833 out of 849) of the artists in our corpus were ranked.

\subsubsection{Train/Val/Test splits} We provide two separate settings to evaluate our proposed methods using stratified and times series split strategies. For both settings, each training, validation, and test sample consists of the full career of an artist, either as a static set of multiple instances (artworks) for static methods or as a sequence of sets of multiple instances for sequential methods.

\begin{figure*}[t]
\centering
\begin{minipage}{0.2\textwidth}
\centering
\includegraphics[width=1.1\textwidth]{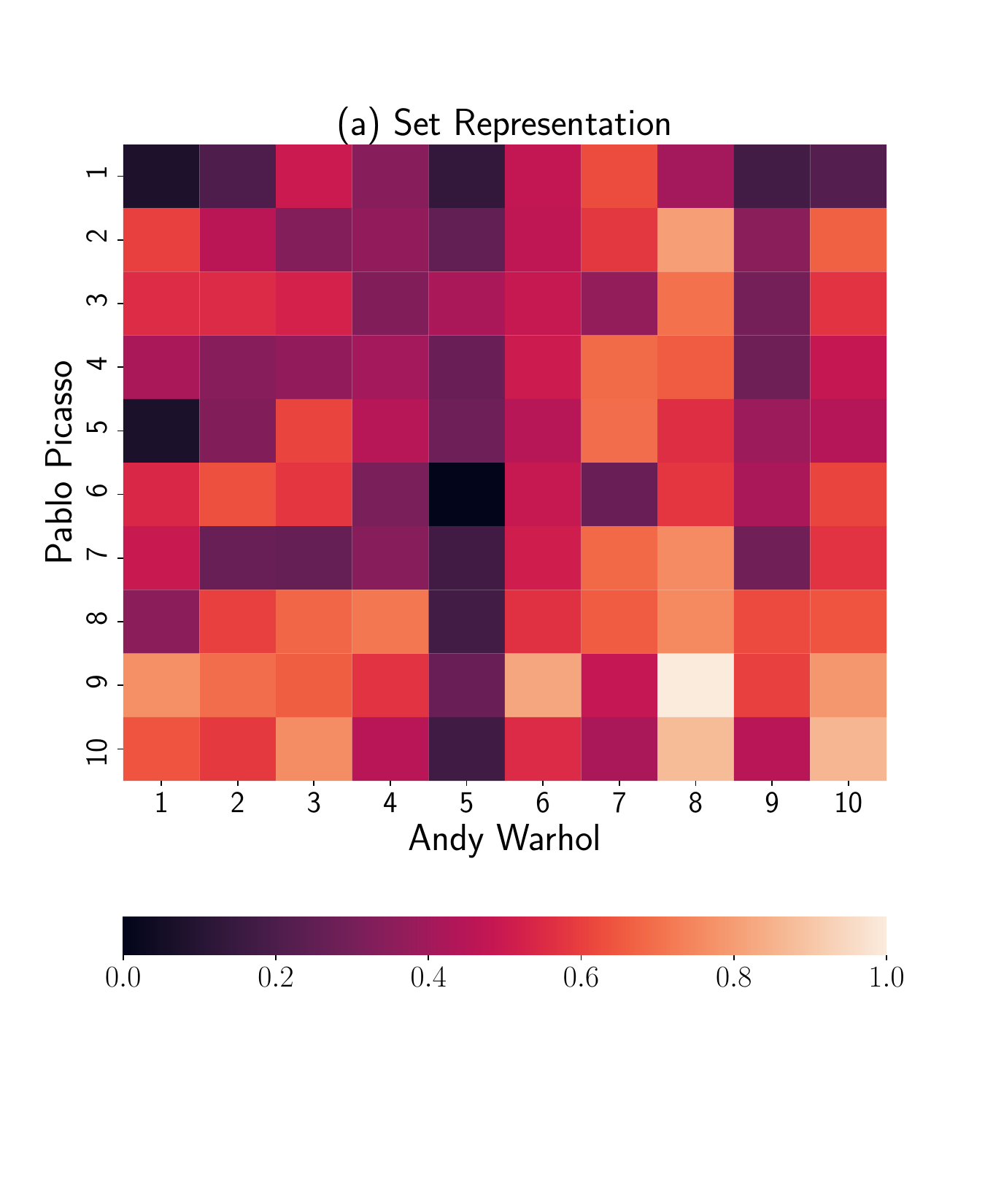}
\end{minipage}\hfill
\begin{minipage}{0.2\textwidth}
\centering
\includegraphics[width=1.1\textwidth]{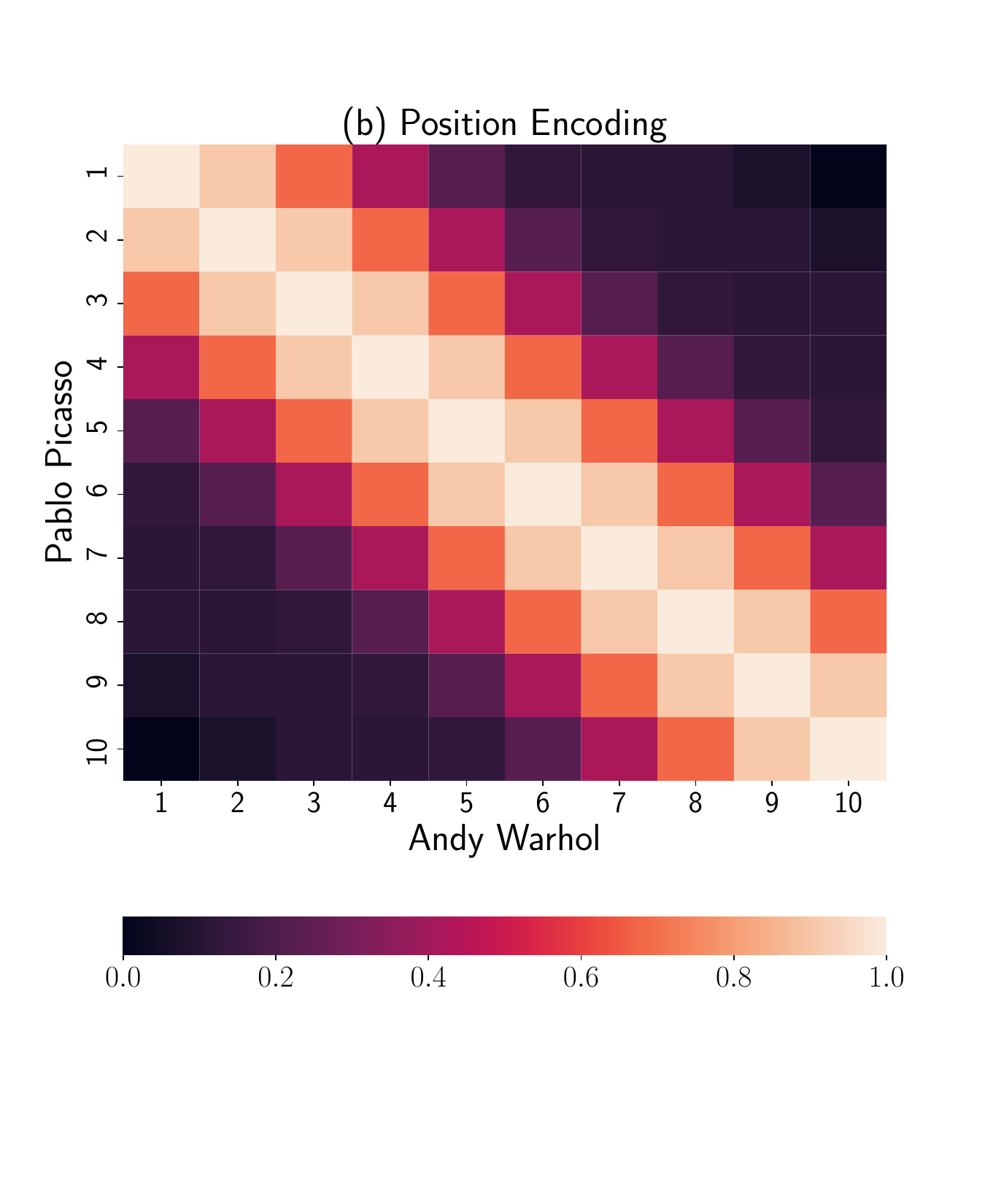}
\end{minipage}\hfill
\begin{minipage}{0.2\textwidth}
\centering
\includegraphics[width=1.1\textwidth]{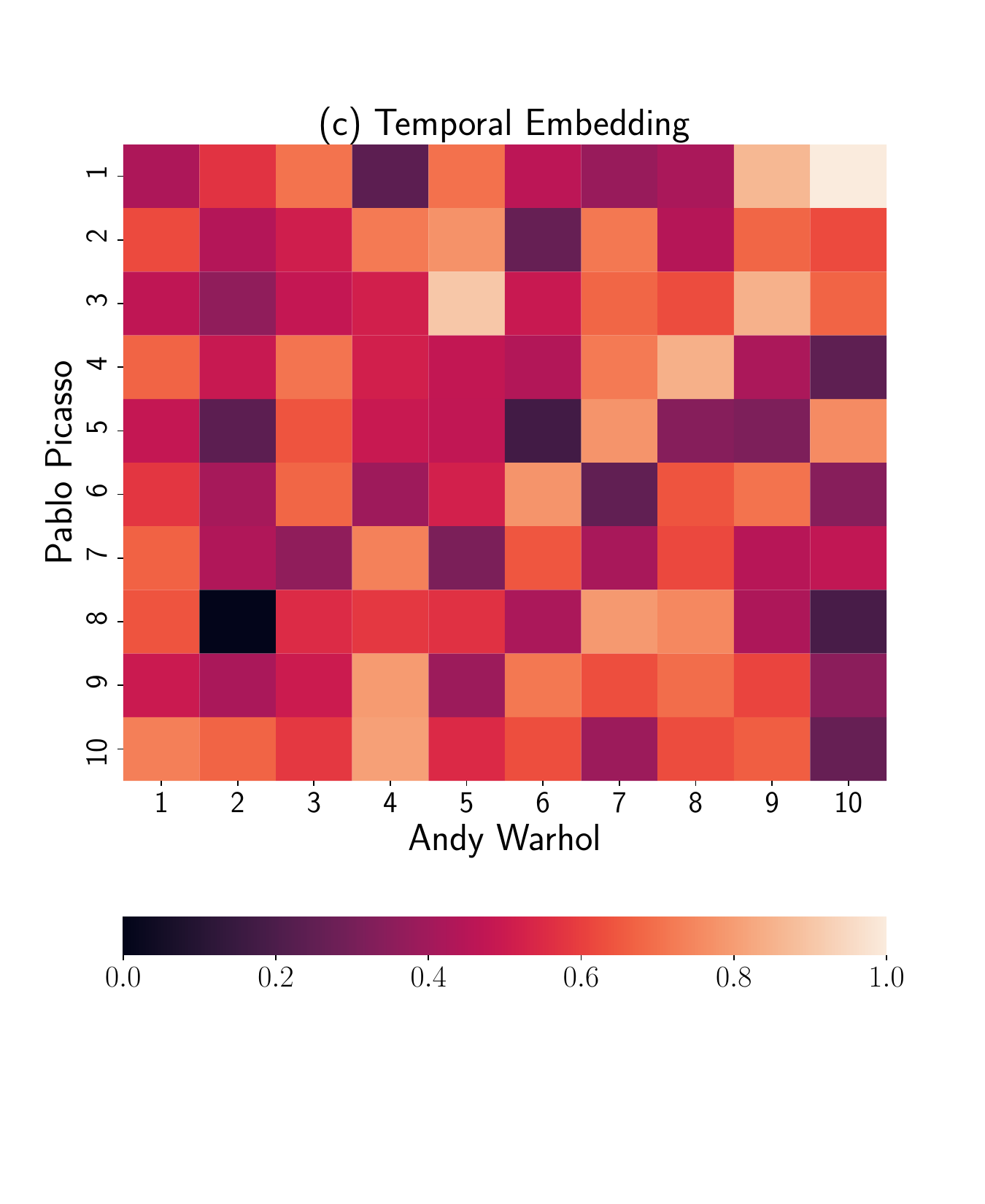}
\end{minipage}\hfill
\begin{minipage}{0.2\textwidth}
\centering
\includegraphics[width=1.1\textwidth]{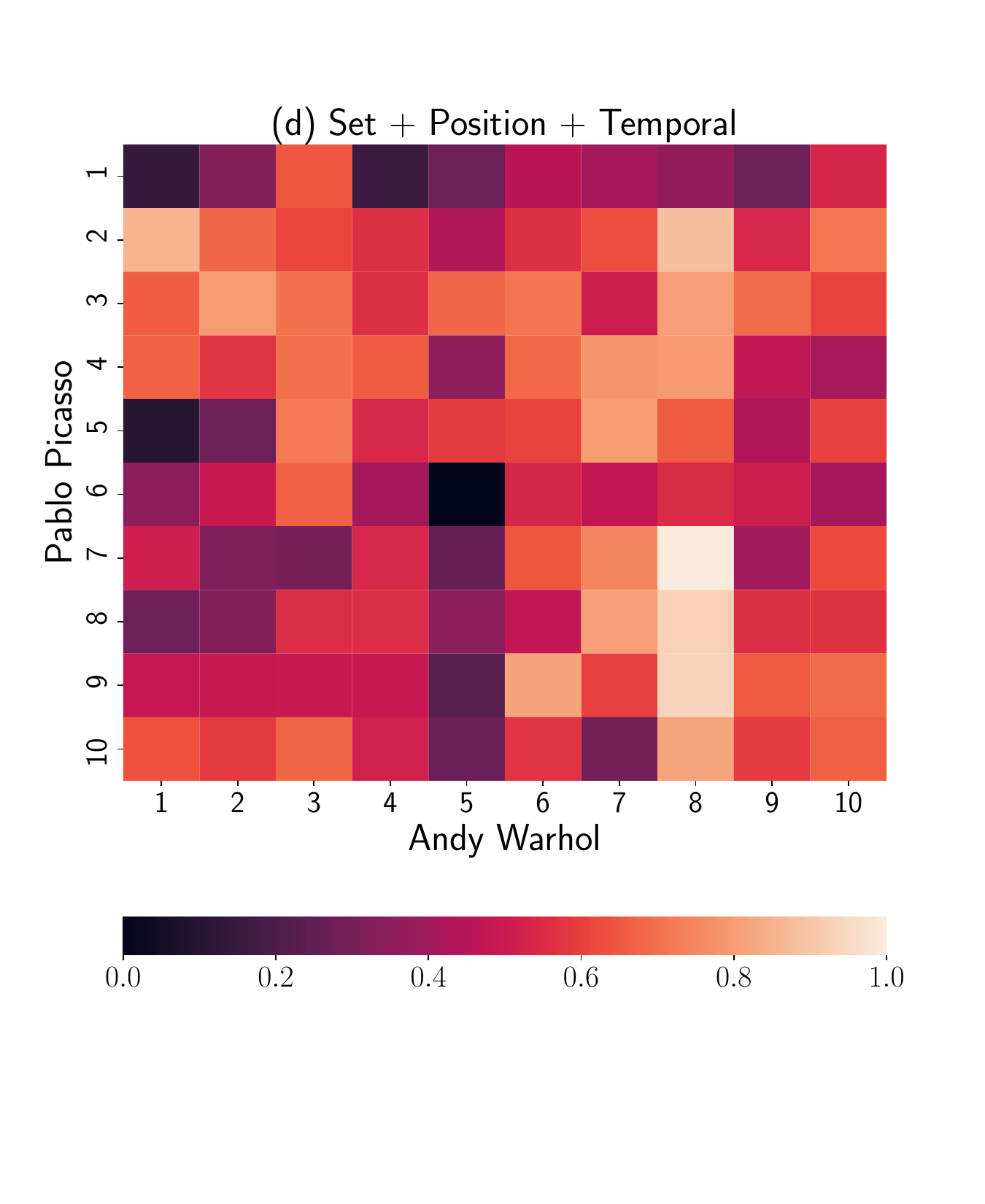}
\end{minipage}\hfill
\begin{minipage}{0.2\textwidth}
\centering
\includegraphics[width=1.1\textwidth]{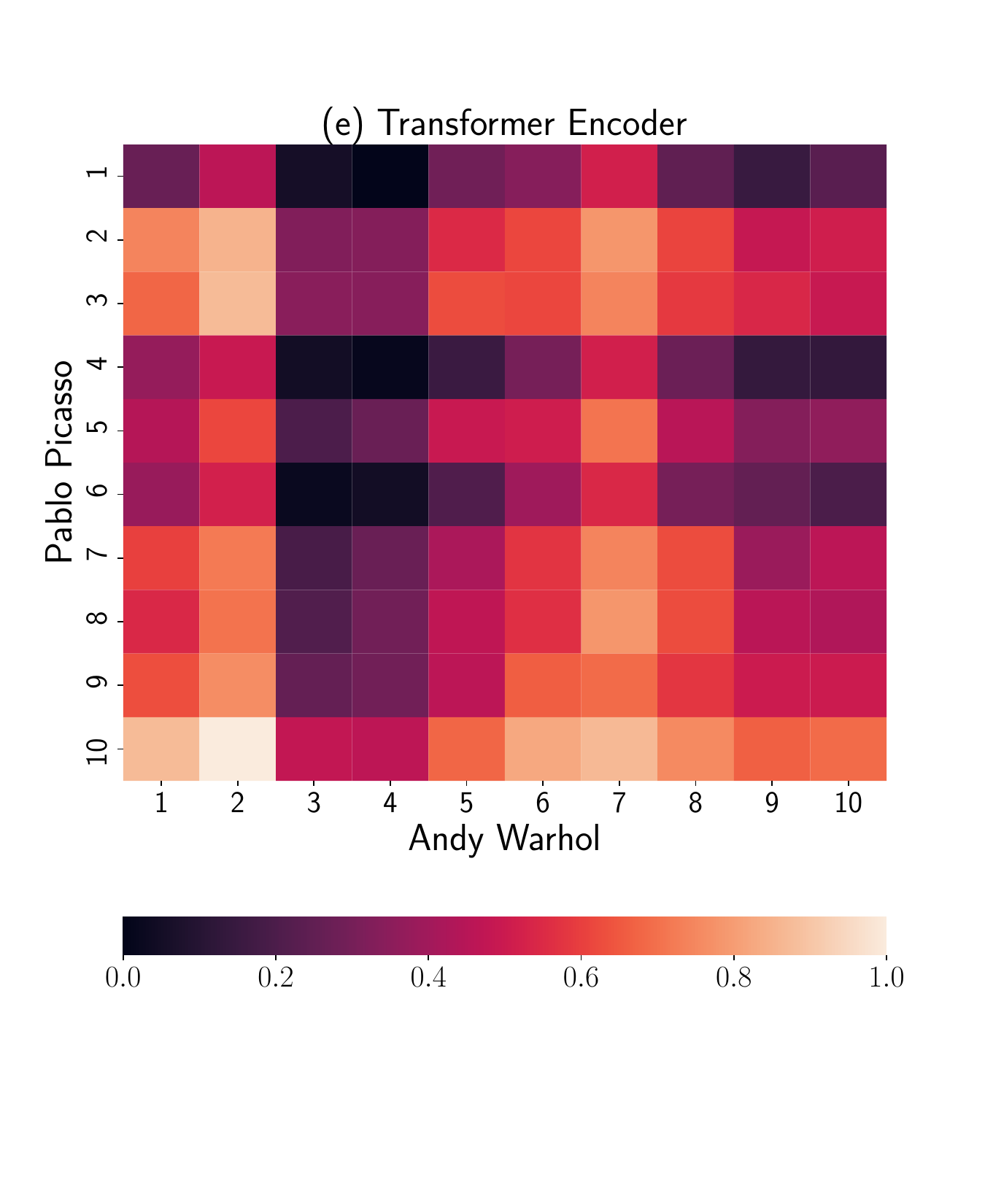}
\end{minipage}
\caption{Visualization of pairwise cosine distances between the first ten years of the top-2 ranked artists' careers based on the Aggregate Ranking (Pablo Picasso and Andy Warhol). (a) illustrates the pairwise cosine distance between the set representations of the first ten years of Pablo Picasso and Andy Warhol's careers. (b) illustrates the Position Encoding. (c) illustrates the Temporal Embeddings. (d) illustrates the aggregate of the set representations, positional encodings and temporal embeddings. (e) illustrates the final output obtained from the Transformer module. Lighter denotes higher cosine similarity.}
\label{fig:pairwise_cosine_distances}
\end{figure*}

\begin{figure*}[t]
    \includegraphics[width=\textwidth]{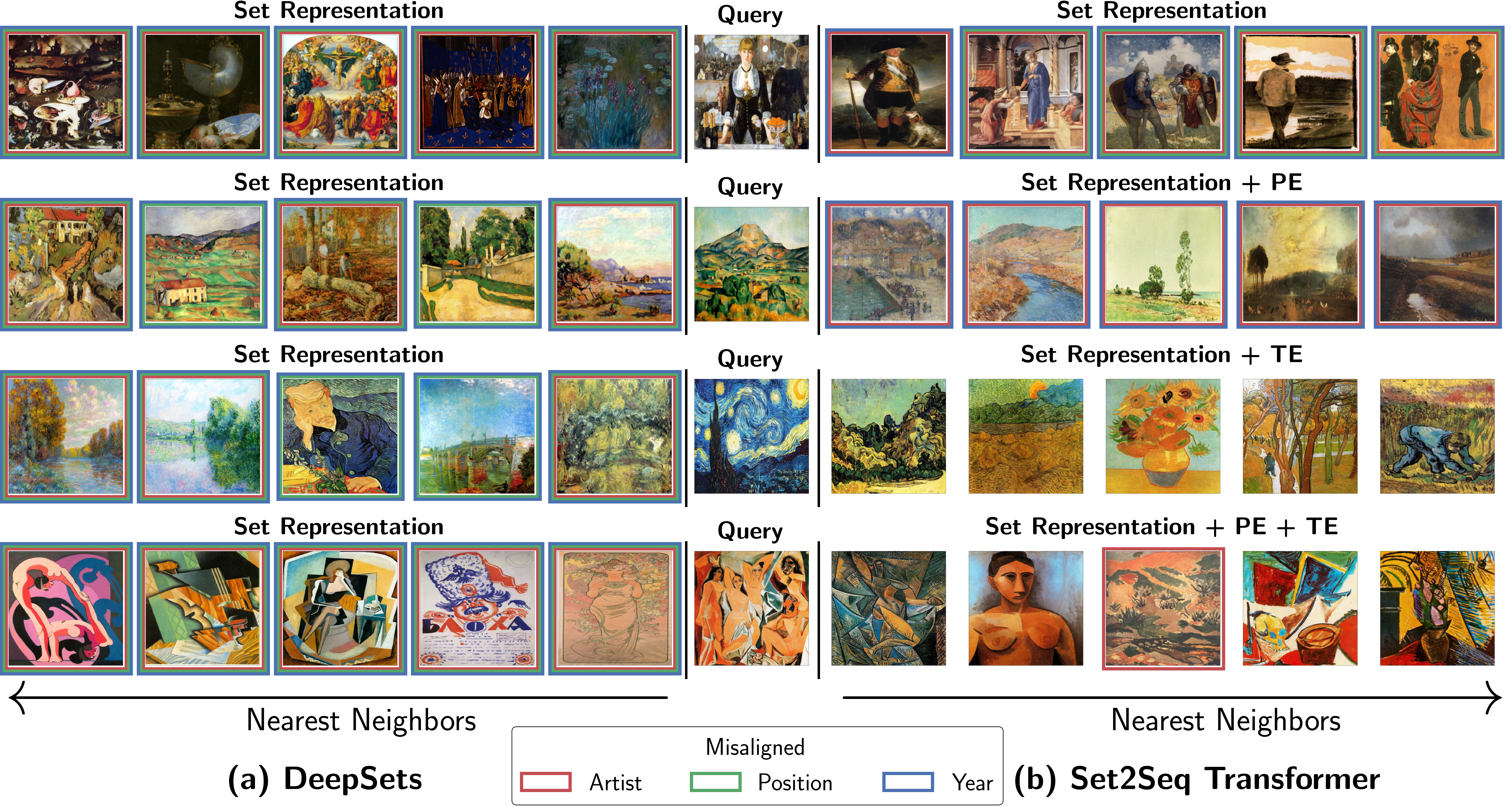}
    \caption{Qualitative analysis of learned visual representations for single-instance painting retrieval. Given a reference painting (middle), the top-5 nearest neighbors of DeepSets (left) and Set2Seq Transformer were retrieved (right). Misaligned patches denote paintings attributed with different \textcolor[HTML]{C44E52}{artist}, \textcolor[HTML]{55A868}{position} or \textcolor[HTML]{4C72B0}{year} annotations from the reference painting.}
   \label{fig:qualitative_analysis_embeddings}
\end{figure*}

\paragraph{Stratified split} First, we use a ranking-based stratified split strategy to split the dataset into training (70\% of the artists), validation (10\% of the artists) and test (20\% of the artists) subsets. 

\paragraph{Time series split} Inspired by real-world applications, in which based on the assessments of known (seen) artists the goal is to predict the artistic performance of unknown (unseen) recently active artists, we split the data at fixed time intervals. We use artists that started their careers before 1930 (exclusive) as the training set (70\% of the dataset), the artists that started their careers in the period $[1930,1951)$ as the validation set (10\% of the dataset), and the remaining artists that started their careers after the year 1951 (inclusive) as the test set (20\% of the dataset). Table~\ref{tab:rankings_statistics} summarizes the statistics for the different rankings and Table~\ref{tab:rankings_top_10} provides a list of the top-10 artists per individual ranking.

\begin{figure*}[t]
\centering
\subfloat[{Pablo Picasso}\label{fig:attention_plot_pablo_picasso}]{
    \includegraphics[width=\textwidth]{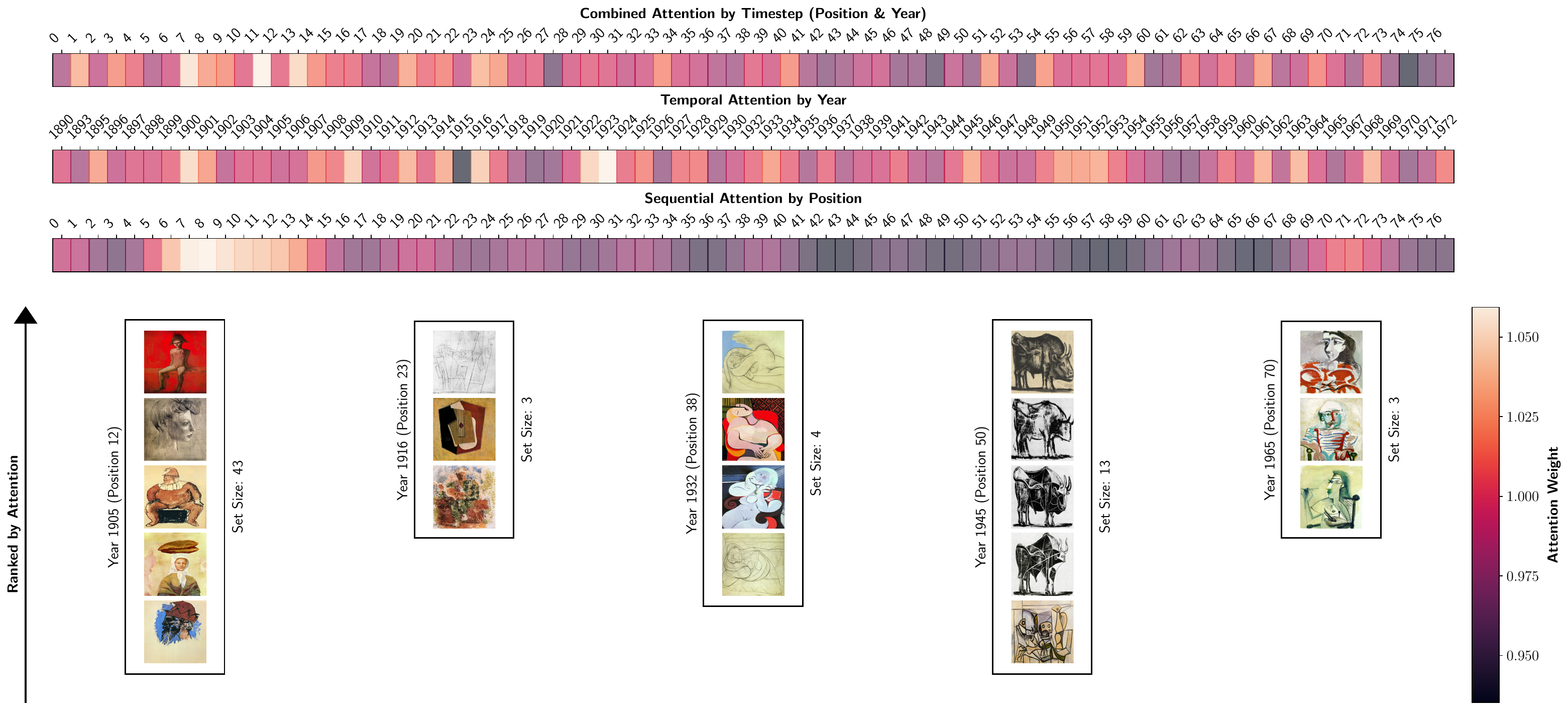}
}

\vspace{1em}

\subfloat[Andy Warhol\label{fig:attention_plot_andy_warhol}]{
    \includegraphics[width=\textwidth]{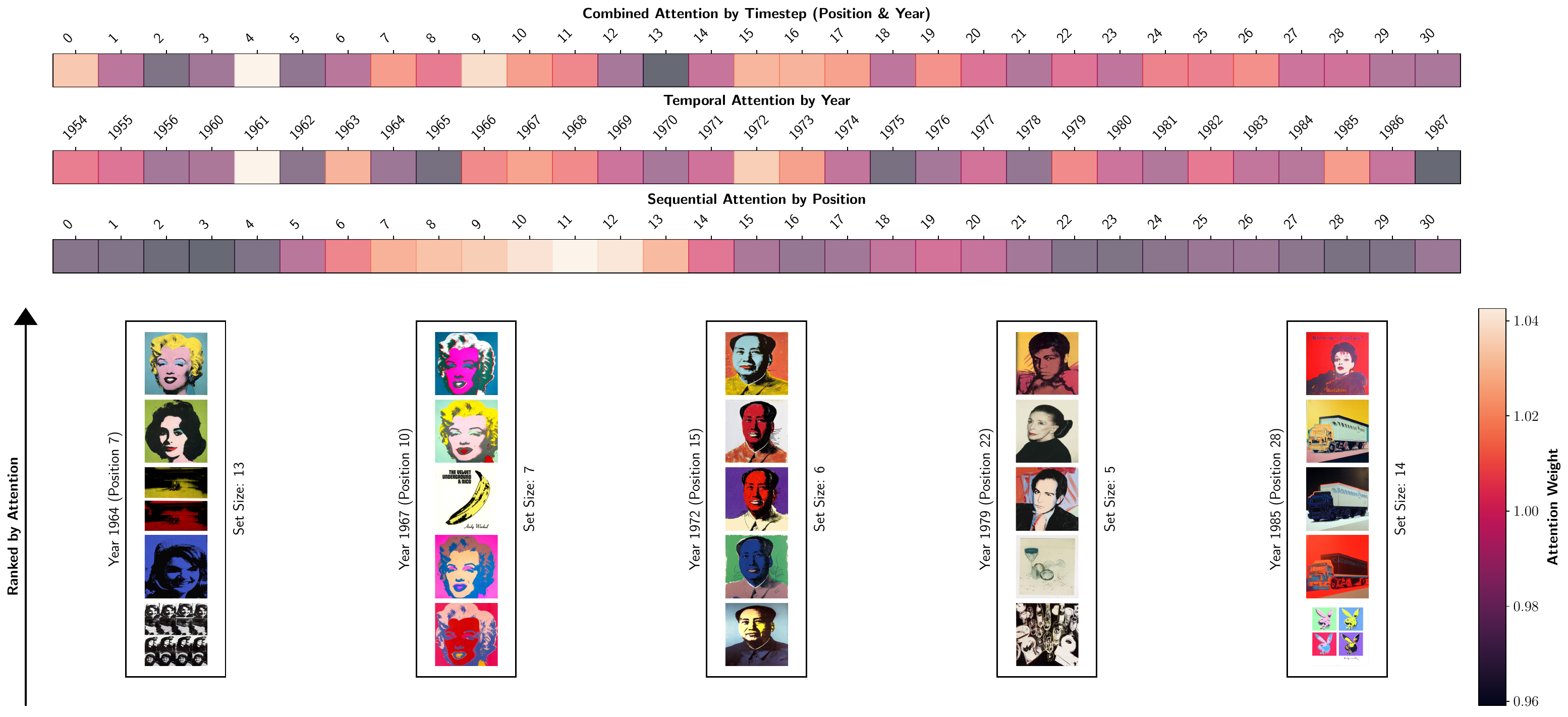}
}
\caption{Hierarchical attention visualization for artist career analysis. We visualize the learned attention patterns of the Set2Seq-Transformer model (ISAB + PMA for set aggregation, multi-head self-attention for sequence modeling) trained to predict aggregate artist rankings for Pablo Picasso and Andy Warhol. The timeline displays three x-axes showing sequence-level attention from the transformer operating on learned set representations: combined attention using both position and year embeddings (top), temporal attention using only year embeddings (middle), and positional attention using only position embeddings (bottom), where attention importance is computed from the final transformer layer. For each manually selected year (columns in main panel), we display up to five artworks ranked by set-level PMA attention (averaged across heads and seeds).}
\label{fig:attention_plots}
\end{figure*}

\begin{table*}[t]
\caption{Kendall's $\tau$ and MAE for different methods and rankings using the time series split test set. For this setting we omit results with learnable temporal embeddings due to the misalignment of the training and test set. PE denotes utilizing position encodings. (mean) and (max) refer to pooling operations. For Set2Seq Transformer using Set Transformer as the set-based module, the specific variant is denoted in parentheses. For Kendall’s $\tau$, BT, and Borda, higher ($\bigtriangleup$) is better; for MAE, lower ($\bigtriangledown$) is better. Best results in \textbf{bold}; second best \underline{underlined}.}
\label{tab:results_time_series_split}
\input Tables/results_time_series_split.tex
\end{table*}

\begin{figure*}[t]
\centering
\begin{minipage}{0.5\textwidth}
\centering
\includegraphics[width=\textwidth]{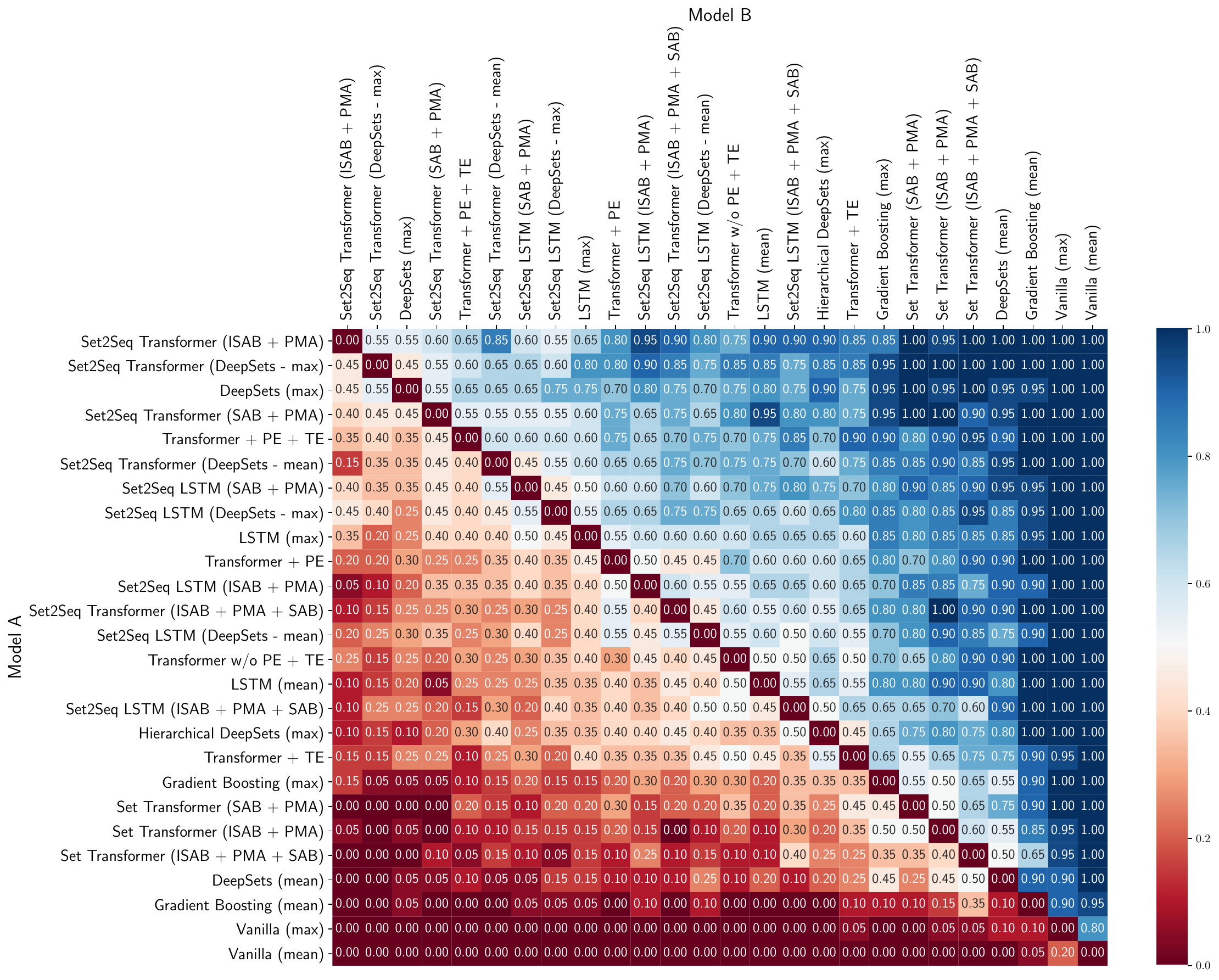}
\end{minipage}\hfill
\begin{minipage}{0.5\textwidth}
\centering
\includegraphics[width=\textwidth]{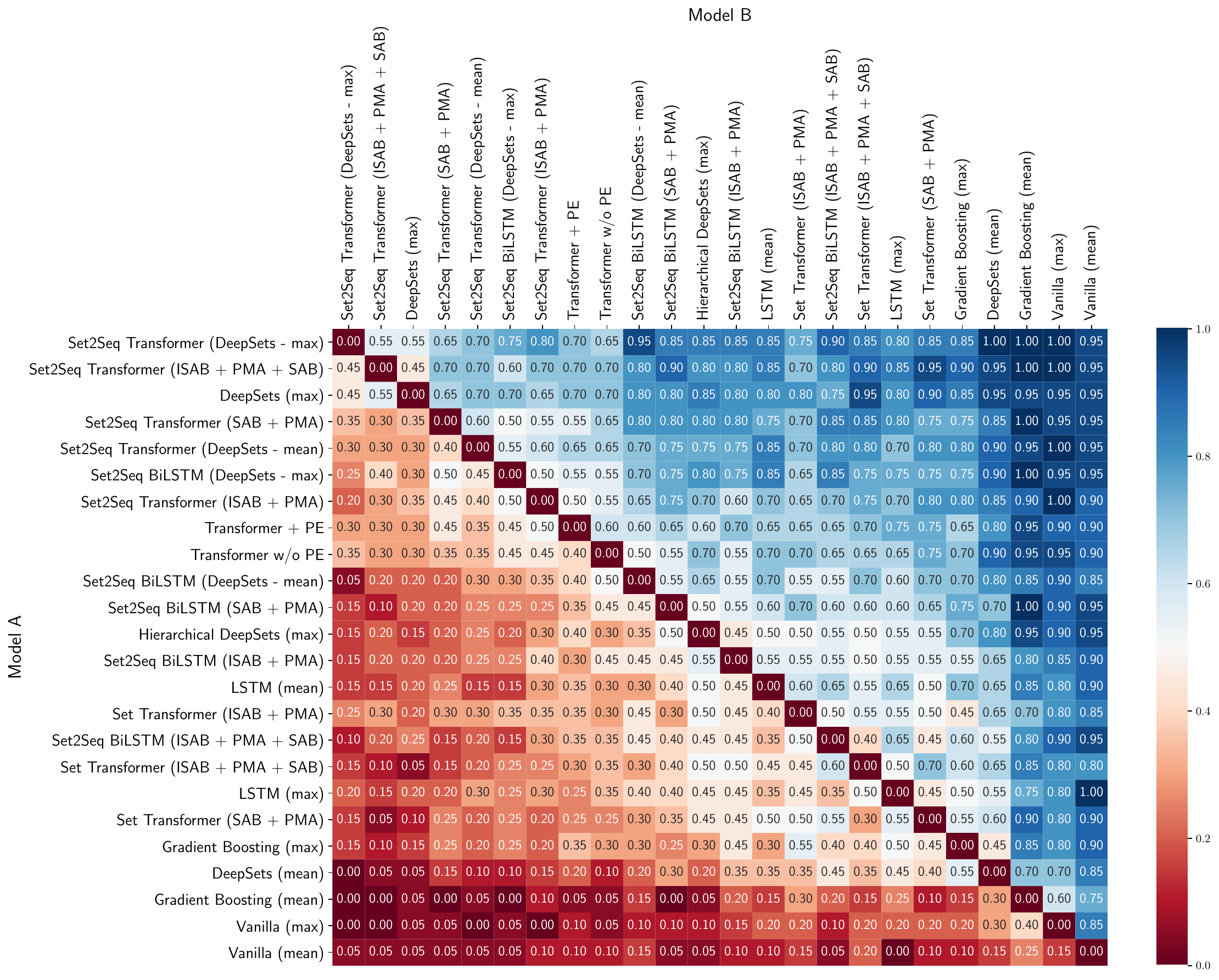}
\end{minipage}
\caption{Pairwise win-rate heatmap showing the fraction of
times Model A (rows) defeats Model B (columns) in head-to-
head comparisons for the stratified split (left) and time series split (right). Models are ordered by their average win-rate across all comparisons.}
\label{fig:pairwise_win_fraction}
\end{figure*}

\subsection{Analysis of Learned Representations} 
\label{subsec:qualitative_analysis_of_set2seq_transformer_learned_representations}

In addition to our experimental results, we provide qualitative analysis of our Set2Seq Transformer.

\subsubsection{Qualitative analysis of temporal and position-aware representations} Inspired by~\cite{position_embeddings_learn}, we provide qualitative analysis of the learned temporal and position-aware representations of our Set2Seq Transformer. Figure~\ref{fig:pairwise_cosine_distances} illustrates the pairwise cosine distances between Pablo Picasso and Andy Warhol for the first ten years of their respective careers, with point at (\textit{i, j}) illustrating the cosine distance between the \textit{i}-th horizontal position with the \textit{j}-th vertical position. An interesting observation based on Figure~\ref{fig:pairwise_cosine_distances}(e), is that the most similar year of Pablo Picasso respective career to Andy Warhol's is the latest one. In contrast, Andy Warhol's initial two years are generally closer, especially the second year, to the first ten years of Pablo Picasso's career. This is to be expected to a certain degree, given the time difference between the two artists' careers.

\subsubsection{Qualitative analysis on single-instance retrieval} Following ~\cite{artsagenet}, we perform qualitative analysis on a paintings retrieval task. In particular, we extract learned representations for DeepSets and Set2Seq Transformer and use k-nearest neighbors (k-NN) to retrieve the top-5 neighbors by cosine distance to a query image. Figure~\ref{fig:qualitative_analysis_embeddings} illustrates the results for DeepSets (on the left side) and our Set2Seq Transformer (on the right side). It can be clearly seen that the single-instance DeepSets representations capture some relevant characteristics, while they lack temporal information. In contrast, Set2Seq Transformer representations with position-aware and temporal embeddings retrieve images with similar characteristics, successfully integrating temporal position-aware information with visual content. An interesting observation is that the only misalignment in our Set2Seq Transformer retrieval for Pablo Picasso's Les Demoiselles d’Avignon (1907) is Henri Matisse's Landscape with Brook (1907), an artist with a great influence from and on Pablo Picasso~\cite{matisse_picasso}.

\subsubsection{Qualitative analysis of attention mechanisms} To gain insight into how the Set2Seq Transformer prioritizes information at both set and sequence levels, we visualize the learned attention patterns for the careers of Pablo Picasso and Andy Warhol in Figure~\ref{fig:attention_plots}. We extract sequence-level attention weights from the final transformer layer, which processes the learned set representations (from ISAB + PMA) augmented with positional and temporal information, using three separate forward passes: one with both positional and temporal embeddings (combined), one with only temporal embeddings (years), and one with only positional embeddings (positions), allowing us to isolate the contribution of each embedding type to the model's predictions. For manually selected years, we rank artworks by their PMA attention weights (averaged across heads and seeds), revealing which instances within each set the model deems most important. The visualization reveals that the model learns to distribute attention relatively uniformly within thematic series (e.g., Warhol's Marilyn variations in 1967, Picasso's Bull plates in 1945), suggesting that the model recognizes variations of the same work as equally representative of the artist's output during that period. While absolute attention weights vary with set size, the ranking patterns indicate that the model successfully integrates temporal and positional information with visual content to capture career-level artistic importance.

\begin{table*}[t]
\caption{Kendall's $\tau$ and MAE for different visual feature extractors using the stratified split validation set with Gradient Boosting and max aggregation. For Kendall's $\tau$, higher ($\bigtriangleup$) is better. For MAE, lower ($\bigtriangledown$) is better. Best results in \textbf{bold}; second best \underline{underlined}.}
\label{tab:visual_embeddings_random_split}
\input Tables/visual_embeddings_random_split.tex
\end{table*}

\begin{table*}[t]
\caption{Kendall's $\tau$ and MAE for different visual feature extractors using the time series split validation set with Gradient Boosting and max aggregation. For Kendall's $\tau$, higher ($\bigtriangleup$) is better. For MAE, lower ($\bigtriangledown$) is better. Best results in \textbf{bold}; second best \underline{underlined}.}
\label{tab:visual_embeddings_time_series_split}
\input Tables/visual_embeddings_time_series_split.tex
\end{table*}

\subsection{Additional Results}

This section presents additional experimental results and analyses supporting the findings reported in the main paper.

\subsubsection{Results using the time series split}

Table~\ref{tab:results_time_series_split} presents results under the time series split. All methods exhibit performance degradation compared to the stratified split, which is expected given the challenging task of predicting the success of modern artists based on historical observations. Despite this distributional shift, Set2Seq Transformer variants remain robust, outperforming most baselines and maintaining strong performance across ranking criteria. Methods explicitly modeling temporal structure (LSTM, Transformer, Set2Seq) show greater resilience than static baselines.

\paragraph{Leaderboard and aggregate performance} To summarize performance across artist ranking criteria, we report Bradley--Terry (BT) and Borda leaderboard scores following the evaluation protocol in Section~\ref{subsec:experimental_setup_predicting_artistic_success} of the main paper. BT scores are derived from pairwise win--loss comparisons~\cite{bt_scores,chatbot_arena}, while Borda scores aggregate rank-based points across metrics and datasets~\cite{nlp_benchmarking}. Under the time series split, Set2Seq Transformer variants achieve the strongest aggregate performance. The variant with DeepSets and max pooling achieves the highest Borda score and ranks second in BT score, while the variant with Set Transformer (ISAB + PMA + SAB) achieves the highest Kendall's $\tau$ and ties for the lowest MAE. DeepSets with max pooling also achieves the highest BT score and a Borda score comparable to top Set2Seq variants, confirming the strength of expressive set representations.

\paragraph{Pairwise win-rate heatmaps} Figure~\ref{fig:pairwise_win_fraction} presents pairwise win-rate heatmaps under both the stratified and time series splits. Each cell shows the fraction of ranking criteria where the row model outperforms the column model. Following~\cite{chatbot_arena}, models are ordered by average pairwise win rate. Set2Seq Transformer variants consistently achieve top rankings, with the DeepSets-based variant achieving the highest average win rate under the time series split and ranking second under the stratified split, demonstrating robustness under temporal distribution shift. Across both splits, Set2Seq variants dominate the top positions, with only DeepSets with max pooling and the Transformer with positional and temporal embeddings achieving comparable performance.

\subsubsection{Results using alternative visual backbones}

In this work, we utilize a frozen ResNet-34~\cite{resnet} backbone pretrained on ImageNet~\cite{imagenet} to extract feature vectors for each visual instance $x_j$, as defined in Equation~\eqref{eq:visual_embedding}. Here, we evaluate alternative visual feature extractors, including: (i) multimodal vision-language models (CLIP~\cite{clip}, BLIP~\cite{blip}, BLIP-2~\cite{blip_2}); (ii) unimodal vision models pretrained on ImageNet (VGG-19~\cite{vgg}, ResNet variants~\cite{resnet}, ConvNeXt~\cite{convnext}, ViT-B/16~\cite{vit}); and (iii) art-specific fine-tuned architectures (ResNet-152 and ArtSAGENet variants fine-tuned for tag prediction and multi-task learning~\cite{artsagenet}). Tables~\ref{tab:visual_embeddings_random_split} and~\ref{tab:visual_embeddings_time_series_split} report results obtained using these feature extractors with the Gradient Boosting baseline. Art-specific fine-tuned models consistently improve performance over generic ImageNet-pretrained models. Multimodal models further improve performance in the stratified split, while unimodal and multimodal models perform comparably under the time series split, reflecting increased temporal distribution shift. For consistency and fair comparison across all methods, we adopt ResNet-34 as the default visual backbone in the main experiments.

\section{Mesogeos Dataset}\label{sec:mesogeos_dataset}

This section presents the variables used for the Mesogeos short-term wildfire danger forecasting task, details of the extended temporal embedding implementation and additional ablation studies.

\begin{table*}[t]
\caption{Numerical variables for the short-term wildfire danger forecasting task using Mesogeos~\cite{mesogeos}. The values in the column ‘Feature Name’ correspond to the variable names in the dataset.}
\label{tab:mesogeos_dataset_variables}
\input Tables/mesogeos_dataset_variables.tex
\end{table*}

\subsection{Wildfire Danger Forecasting Variables}\label{subsec:input_features_mesogeos}

Table~\ref{tab:mesogeos_dataset_variables} summarizes the numerical variables for the short-term wildfire danger forecasting task, selected to capture key wildfire drivers~\cite{mesogeos}. To ensure a fair comparison of methods, we strictly adhere to the preprocessing and variable definitions outlined in~\cite{mesogeos}. The target variable is a binary indicator of wildfire danger, derived from the feature burned\_area\_has, which represents the burned area in hectares within a designated buffer zone. That is, if the burned area exceeds 30 hectares, the wildfire danger variable is set to 1; otherwise, it is set to 0.

Meteorological variables, such as Max Temperature (t2m), Min Relative Humidity (rh), and Total Precipitation (tp), are sourced from ERA5-Land~\cite{era5_land}. Vegetation and land surface properties, including Normalized Difference Vegetation Index (ndvi), Leaf Area Index (lai), and Land Surface Temperatures (lst\_day and lst\_night), are obtained from MODIS~\cite{MOD13A2, MOD15A2, MOD11A1}. Other environmental features include the Soil Moisture Index (smi) derived from the European Drought Observatory (EDO)~\cite{european_drought_observatory}, and land cover fractions, such as fractions of agriculture (lc\_agriculture) and forest (lc\_forest), sourced from Copernicus Climate Change Services (CCS) products~\cite{copernicus_climate_change_service}. Human-related variables include population density (population) and roads distance (roads\_distance), obtained from WorldPop~\cite{wordpop}. Elevation data (dem) and its derivatives, such as slope, are provided by COP-DEM~\cite{cop_dem}.

\subsection{Temporal Embedding}\label{subsec:temporal_embedding}

To effectively capture temporal dynamics in wildfire danger forecasting, we extend the temporal embedding approach from the WikiArt-Seq2Rank task. While WikiArt used only creation years, the Mesogeos dataset provides full timestamps, enabling more flexible embeddings with day, month, and year components. In this work, we focus on month-year embeddings for their balance of granularity and noise reduction. As described in Section~\ref{sec:approach} of the main paper, the temporal embedding $\mathbf{v}_i \in \mathbb{R}^D$ for an absolute temporal value $a_i = (c_i, z_i)$, where $c_i$ denotes the cyclic component (month) and $z_i$ denotes the linear component (year), is computed by concatenating the cyclic component $\mathbf{v}_i^{c} \in \mathbb{R}^{D/2}$ and the linear component $\mathbf{v}_i^{z} \in \mathbb{R}^{D/2}$ (see Equation~\eqref{eq:temporal_embedding_timestamp} in the main paper). Here we provide the detailed implementation of these components.

\begin{table*}[t]
\caption{Results using different Set Transformer modules for set representation learning. PE denotes utilizing positional encodings. TE denotes utilizing temporal embeddings. $K$ denotes the number of instances per set $\mathcal{S}$. $N$ denotes the number of timesteps in a sequence $\mathcal{T}$. For both metrics, higher is better. Best results in \textbf{bold}; second best \underline{underlined}.}
\label{tab:mesogeos_results_set_transformer}
\input Tables/mesogeos_results_set_transformer.tex
\end{table*}

\begin{table*}[t]
\caption{Results using different positional encodings and temporal embeddings. PE denotes utilizing positional encodings. TE denotes utilizing temporal embeddings. (DM) denotes using the day and month, (MY) denotes using the month and year, and (DMY) denotes using the day, month and year as the timestamp feature. $K$ denotes the number of instances per set $\mathcal{S}$. $N$ denotes the number of timesteps in a sequence $\mathcal{T}$. For both metrics, higher is better. Best results in \textbf{bold}; second best \underline{underlined}.}
\label{tab:mesogeos_results_pe_ce}
\input Tables/mesogeos_results_pe_ce.tex
\end{table*}

\subsubsection{Cyclic month encoding ($\mathbf{v}_i^{c}$)}

Our cyclic encoding for months is inspired by the Time2Vec framework~\cite{time2vec}, which uses learnable sine and cosine functions to represent periodic temporal patterns. We employ learnable frequency and phase parameters to adaptively encode cyclic patterns. The month encoding uses sine and cosine functions with learnable parameters for frequency ($\boldsymbol{\omega}_c$) and phase ($\boldsymbol{\phi}_c$):

\begin{equation}\label{al:month_embedding}
\mathbf{v}_i^{c} =
\big[\sin(\boldsymbol{\omega}_c c_i + \boldsymbol{\phi}_c),
\cos(\boldsymbol{\omega}_c c_i + \boldsymbol{\phi}_c)\big].
\end{equation}
Here, $c_i$ represents the normalized month, computed as:

\begin{equation}\label{al:month_normalization}
c_i = \frac{\text{month}}{12} \cdot 2\pi,
\end{equation}
where $\text{month} \in \{1, 2, \dots, 12\}$, $\boldsymbol{\omega}_c \in \mathbb{R}^{D/4}$ is a learnable frequency vector, and $\boldsymbol{\phi}_c \in \mathbb{R}^{D/4}$ is a learnable phase vector. The final cyclic embedding, combining sine and cosine components, is of size $D/2$.

\subsubsection{Linear year encoding ($\mathbf{v}_i^{z}$)}

Unlike Time2Vec~\cite{time2vec}, which uses a simple linear term to encode temporal progression, our year embedding employs a feed-forward network (FFN) to capture complex, non-linear trends in temporal data. This flexibility is crucial for wildfire forecasting, where temporal patterns often involve abrupt shifts and non-linear trends due to climate and human activity. The linear temporal embedding is calculated using a feed-forward network $FFN(\cdot)$ combined with learnable weights and biases:

\begin{equation}\label{al:learnable_year_representation}
\mathbf{v}_i^{z} = FFN(z_i) \mathbf{w}_z + \mathbf{b}_z,
\end{equation}
where $z_i$ is the normalized year value, computed as:

\begin{equation}\label{al:year_normalization}
z_i =
\frac{\text{year} - \text{min\_year}}
{\text{max\_year} - \text{min\_year}}.
\end{equation}
The feed-forward network $FFN(z_i)$ is defined as:

\begin{equation}\label{al:ffn_year}
FFN(z_i) =
\sigma\big(\mathbf{W}_1 \cdot z_i + \mathbf{b}_1\big)
\mathbf{W}_2 + \mathbf{b}_2,
\end{equation}
where $\sigma$ is a non-linearity, such as the Rectified Linear Unit (ReLU), defined as:

\begin{equation}\label{al:relu}
\text{ReLU}(x) = \max(0, x).
\end{equation}
Here, $\mathbf{W}_1, \mathbf{W}_2 \in \mathbb{R}^{D/4}$ are the weights of the feed-forward network, $\mathbf{b}_1, \mathbf{b}_2 \in \mathbb{R}^{D/4}$ are its biases, $\mathbf{w}_z \in \mathbb{R}^{D/2}$ is the learnable linear weight, and $\mathbf{b}_z \in \mathbb{R}^{D/2}$ is the learnable linear bias. The final temporal embedding is computed as the concatenation of the cyclic and linear embeddings, as shown in Equation~\eqref{eq:temporal_embedding_timestamp} of the main paper.

\subsection{Additional Results}\label{subsec:additional_results_mesogeos}

Here, we present additional results from evaluations of the Set2Seq Transformer, focusing on variations in its Set Transformer module configurations and temporal representations.

\subsubsection{Set Transformer module configurations} Table~\ref{tab:mesogeos_results_set_transformer} summarizes the performance of the Set2Seq Transformer with different Set Transformer modules for set representation learning. The results indicate that the Set Transformer (ISAB + PMA + SAB) configuration achieves superior performance on the short-term wildfire danger forecasting task in single-instance learning settings. Moreover, integrating temporal embeddings consistently improves performance over positional encoding-only variants. It is worth noting that the Set Transformer (ISAB + PMA) variant performs comparably to the other configurations in single-instance learning tasks, while outperforming them in multiple-instance learning scenarios.

\subsubsection{Temporal representation strategies} Table~\ref{tab:mesogeos_results_pe_ce} reports the results of ablation studies exploring alternative temporal position-aware representations for the Set2Seq Transformer. We experiment with different strategies for timestamp-based temporal embeddings integrated into set-based methods, such as Deep Sets and Set Transformer. The results demonstrate that incorporating temporal position-aware representations enhances performance compared to methods relying solely on positional encodings. The temporal embedding that utilizes month-year information derived from timestamps outperforms alternatives using day-month or day-month-year information, potentially due to its balance between granularity and reduced noise.

\section{Limitations and Broader Impact}\label{sec:limitations_broader_impact}

In this section we discuss the methodological challenges encountered in this work, highlighting directions for future improvement, and exploring the broader implications of our contributions.

\subsection{Limitations}\label{subsec:limitations}

\subsubsection{Rankings construction} The artist appreciation rankings in WikiArt-Seq2Rank provide a framework for analyzing artistic success by compiling data from diverse publicly available sources such as eBooks, Wikipedia, Artfacts, and Artprice, grounded in cultural industries theory~\cite{deep_learning_based_product_distinctiveness, relations_aesthetic_space, selection_systems}. These rankings are central to understanding artistic success as they capture multiple dimensions of recognition, including public interest, institutional acknowledgment, and commercial success, offering a unique perspective on the complex dynamics of cultural impact. However, as a proxy rather than an absolute measure, these rankings may reflect notoriety or controversy instead of positive recognition, and biases inherent in the source data---such as geographic, cultural, or institutional priorities---can influence the results. Therefore, while these rankings offer valuable insights, they should be interpreted cautiously and contextualized within the broader scope of artistic practices and historical influences to ensure meaningful conclusions.

\subsubsection{Learning-to-Rank} In this work, we adopt a pointwise learning-to-rank approach by optimizing the Mean Squared Error (MSE) loss. While MSE captures ranking trends, it does not directly optimize metrics such as Kendall’s $\tau$ or NDCG. Future work could explore pairwise and listwise approaches~\cite{listwise_learning_to_rank, pairwise_learning_to_rank}, as well as other recent methods~\cite{learning_to_rank_grades_matter}, to directly improve ranking~performance.

\subsubsection{Performance evaluation and benchmarking} In this work, we benchmark various methods across multiple datasets and metrics. For fair evaluation, all methods were trained using the same hyperparameters, including batch size, optimizer, learning rate, and early stopping criteria, with minor adjustments for task-specific requirements. While exhaustive hyperparameter optimization was beyond the scope of this work, task-specific tuning and advanced training strategies could further improve results. For instance, leveraging widely adopted techniques such as pretraining with masked inputs~\cite{bert}, autoregressive methods~\cite{gpt3}, and learning rate schedules (e.g., cosine warm-up~\cite{sgdr}) could enhance the ability of Transformer-based methods to better capture temporal dependencies and improve training stability. Systematically investigating these techniques represents a promising avenue for further advancing sequential multiple-instance learning. From an evaluation perspective, given the extensive benchmarking across several baselines, multiple datasets, and diverse metrics, we focus on large-scale comparative analysis to assess consistency and robustness across evaluation settings, rather than exhaustive pairwise statistical significance testing. Future work may incorporate additional statistical analyses to further characterize variability in specific pairwise comparisons.

\subsubsection{Complexity of learned temporal embeddings} The temporal embeddings in this work are designed to balance simplicity and expressiveness, adapting to the granularity of the datasets. For the WikiArt-Seq2Rank task, where only creation years are available, learnable year-based embeddings provide an effective and straightforward solution. For the Mesogeos dataset, a more advanced embedding leverages full timestamps to capture fine-grained temporal dynamics, which is critical for accurately modeling wildfire danger. Although the embedding approach for Mesogeos requires greater computational effort, it demonstrates substantial benefits in capturing temporal patterns and enhancing predictive performance. This dual framework underscores the adaptability of the temporal embeddings to diverse datasets and tasks, providing a foundation for future investigations into the trade-offs between complexity and efficiency.

\subsection{Broader Impact}\label{subsec:broader_impact}

\subsubsection{Interpretability in Transformer-based sequential multiple-instance learning} Our Set2Seq Transformer leverages attention mechanisms in both its Set Transformer module and Transformer layer, offering interpretability into the importance of individual set elements and the evolution of sequential temporal dependencies. For the WikiArt-Seq2Rank task, these mechanisms can uncover relationships between artists or artworks that drive rankings, revealing patterns of influence and recognition. Similarly, for the Mesogeos dataset, temporal attention can highlight critical periods or environmental factors influencing wildfire danger predictions. While this work does not explicitly explore these applications, the interpretability provided by attention weights and learned positional and temporal representations lays a solid foundation for future analyses in both artistic and environmental domains.

\subsubsection{Applications to different domains} Modeling set-to-sequence relationships offers valuable opportunities for applications across creative domains such as literature, music, and film, where understanding temporal progression and patterns of creativity is essential. These methods could be used to analyze how authors, composers, or filmmakers evolve their craft over time, shedding light on artistic trajectories and innovation~\cite{attention_evolution_films, macroanalysis_digital_methods_literary_history, evolution_popular_music, computational_music_analysis, literary_change}. Furthermore, the positional and temporal embeddings developed in this work can be adapted to study broader phenomena in creativity and cultural influence, capturing an artist's career stage or uncovering patterns shaped by historical and cultural contexts. By providing a flexible framework for integrating diverse types of input, these methods enable novel analyses and contribute to a deeper understanding of temporal and contextual dynamics in creative fields.

\subsubsection{Effect of climate change} This work focuses on short-term wildfire danger forecasting using existing datasets and methodologies, without directly addressing the impact of climate change. However, the temporal embedding strategies introduced here, designed to capture fine-grained temporal dynamics, could be adapted to study long-term trends in wildfire patterns and their relationship to evolving climate conditions. Future research could build on these embeddings to analyze temporal trends in environmental variables, providing a framework for investigating connections between wildfire danger and climate change with appropriate validation and adaptation.

%% file: Figures/static_temporal_aware_baselines.tex
\definecolor{AntiqueWhite}{RGB}{250,235,215}
\definecolor{LightSteelBlue}{RGB}{176,196,222}
\definecolor{DarkSeaGreen}{RGB}{143,188,143}

\newcommand\HUGE{\fontsize{40}{60}\selectfont}
\newcommand\HUGER{\fontsize{60}{60}\selectfont}

\begin{adjustbox}{max totalsize={\textwidth}{\textheight},center}
\begin{tikzpicture}

\tikzset{
  -|-/.style={
    to path={
      (\tikztostart) -| ($(\tikztostart)!#1!(\tikztotarget)$) |- (\tikztotarget)
      \tikztonodes
    }, rounded corners=2pt
  },
  -|-/.default=0.5,
  |-|/.style={
    to path={
      (\tikztostart) |- ($(\tikztostart)!#1!(\tikztotarget)$) -| (\tikztotarget)
      \tikztonodes
    }
  },
  |-|/.default=0.5,
}

\tikzstyle{group}= [rectangle, draw, rounded corners=10pt, inner sep=10pt]
\tikzstyle{group_dashed}= [rectangle, draw, dashed, rounded corners=4pt, inner sep=50pt]
\tikzstyle{images} = [text height=1.5ex, text depth=.25ex, text width=11em, text centered, minimum height=2em, anchor=mid]

%%%%%%%%%%%%%%%%%%%%%%%%%%%%%%%%%%%%%%%%%
% DEFINE FIXED POSITIONS
%%%%%%%%%%%%%%%%%%%%%%%%%%%%%%%%%%%%%%%%%
\def\imageY{0}
\def\outputY{21}

%%%%%%%%%%%%%%%%%%%%%%%%%%%%%%%%%%%%%%%%%
% SHARED INPUT IMAGES AT BOTTOM (y=0)
%%%%%%%%%%%%%%%%%%%%%%%%%%%%%%%%%%%%%%%%%
% DeepSets images (left section)
\node[images] at (0,\imageY) (image0) {\includegraphics[width=\textwidth]{Images/images_sequence_0/image_5.png}};
\node[images, right=.2cm of image0] (image1) {\includegraphics[width=\textwidth]{Images/images_sequence_0/image_0.png}};
\node[images, right=.2cm of image1] (image2) {\includegraphics[width=\textwidth]{Images/images_sequence_0/image_19.png}};
\node[images, right=-1cm of image2, yshift=1.5cm] (image3) {\HUGE$\cdots$};
\node[images, right=2.5cm of image2] (image4) {\includegraphics[width=\textwidth]{Images/images_sequence_0/image_6.png}};

% RNN images (middle section) - start after DeepSets
\node[images, right=7.5cm of image4] (rnn_text0) {\includegraphics[width=\textwidth]{Images/images_sequence_0/image_0.png}};
\node[below=-.1cm of rnn_text0] (rnn_text0_label) {\HUGE $t_1$};
\node[images, right=1.5cm of rnn_text0] (rnn_text1) {\includegraphics[width=\textwidth]{Images/images_sequence_0/image_5.png}};
\node[images, right=0cm of rnn_text1] (rnn_text1b) {\includegraphics[width=\textwidth]{Images/images_sequence_0/image_6.png}};
\node[group_dashed, fit={([yshift=2.5cm, xshift=1.5cm]rnn_text1.north west) ([yshift=1.5cm, xshift=-1.5cm]rnn_text1b.south east)}, inner sep=50pt] (rnn_group_t1) {};
\node[below=1cm of $(rnn_text1)!0.5!(rnn_text1b)$] (rnn_text1_label) {\HUGE $t_2$};
\node[images, right=1.8cm of rnn_text1b, yshift=1.5cm] (rnn_text2) {\HUGER $\cdots$};
\node[images, right=7.5cm of rnn_text1b] (rnn_text3) {\includegraphics[width=\textwidth]{Images/images_sequence_0/image_19.png}};
\node[below=-.1cm of rnn_text3] (rnn_text3_label) {\HUGE $t_N$};

% Transformer images (right section) - start after RNN
\node[images, right=10.5cm of rnn_text3] (tr_text0) {\includegraphics[width=\textwidth]{Images/images_sequence_0/image_0.png}};
\node[below=-.1cm of tr_text0] (tr_text0_label) {\HUGE $t_1$};
\node[images, right=0.8cm of tr_text0] (tr_text1) {\includegraphics[width=\textwidth]{Images/images_sequence_0/image_5.png}};
\node[below=-.1cm of tr_text1] (tr_text1_label) {\HUGE $t_2$};
\node[images, right=0.8cm of tr_text1] (tr_text1b) {\includegraphics[width=\textwidth]{Images/images_sequence_0/image_6.png}};
\node[below=-.1cm of tr_text1b] (tr_text1b_label) {\HUGE $t_2$};
\node[images, right=-0.8cm of tr_text1b, yshift=1.5cm] (tr_text2) {\HUGER $\cdots$};
\node[images, right=8cm of tr_text1] (tr_text3) {\includegraphics[width=\textwidth]{Images/images_sequence_0/image_19.png}};
\node[below=-.1cm of tr_text3] (tr_text3_label) {\HUGE $t_N$};

%%%%%%%%%%%%%%%%%%%%%%%%%%%%%%%%%%%%%%%%%
% OUTPUTS AT TOP (all at same y)
%%%%%%%%%%%%%%%%%%%%%%%%%%%%%%%%%%%%%%%%%
\coordinate (ds_output_pos) at ($(image0)!0.5!(image4)+(0,\outputY)$);
\coordinate (rnn_output_pos) at ($(rnn_text3)+(0,\outputY)$);
\coordinate (tr_output_pos) at ($(tr_text0)!0.5!(tr_text3)+(0,\outputY)$);

\node at (ds_output_pos) (ds_output) {\HUGE \textbf{$\hat{y}$}}; 
\node at ([xshift=1cm]rnn_output_pos) (rnn_output) {\HUGE \textbf{$\hat{y}$}}; 
\node at ([xshift=.2cm]tr_output_pos) (tr_output) {\HUGE \textbf{$\hat{y}$}};

%%%%%%%%%%%%%%%%%%%%%%%%%%%%%%%%%%%%%%%%%
% DEEPSETS ARCHITECTURE (LEFT)
%%%%%%%%%%%%%%%%%%%%%%%%%%%%%%%%%%%%%%%%%
\node[group_dashed,fit={([yshift=2.5cm, xshift=1.5cm]image0.north west) ([yshift=1.5cm, xshift=-1.5cm]image4.south east)}, label={[yshift=-0.5cm]below:\HUGE Unordered Set}] (ds_paintings) {};
\node[draw, rectangle, minimum height=2cm, minimum width=21.5cm, fill=AntiqueWhite!75, above=2.5cm of ds_paintings] (ds_feature) {\HUGE Feature Extractor};
\path[-{Latex[length=6mm, width=5mm]}, line width=3pt] (ds_paintings) edge (ds_feature);
\node[draw, rectangle, fill=LightSteelBlue!50, minimum height=2cm, minimum width=21.5cm, above=2.5cm of ds_feature] (ds_aggregate) {\HUGE Aggregate Operation};
\path[-{Latex[length=6mm, width=5mm]}, line width=3pt] (ds_feature) edge (ds_aggregate);
\node[draw, rectangle, fill=DarkSeaGreen!50, minimum height=2cm, minimum width=21.5cm, above=2.5cm of ds_aggregate] (ds_model) {\HUGE Fully Connected Layers};
\path[-{Latex[length=6mm, width=5mm]}, line width=3pt] (ds_aggregate) edge (ds_model);

\path[-{Latex[length=6mm, width=5mm]}, line width=3pt] (ds_model) edge (ds_output);

\coordinate (ds_caption_pos) at ($(image0)!0.5!(image4)+(0,-6)$);
\node at (ds_caption_pos) {\HUGER (a) Static Baselines};

%%%%%%%%%%%%%%%%%%%%%%%%%%%%%%%%%%%%%%%%%
% RNN ARCHITECTURE (MIDDLE)
%%%%%%%%%%%%%%%%%%%%%%%%%%%%%%%%%%%%%%%%%
% Build RNN layers between images and output
% Position RNN components to fill the vertical space

% Time step 0 - above rnn_text0
\foreach \i / \j in {0/0,1/1.4,2/2.8,3/4.2}
    \node[circle,fill=black!75,draw,inner sep=0pt,minimum size=25pt] at ($(rnn_text0)+(-1.8+\j,7.5)$) (w_0_\i) {};
\node[group,fit={(w_0_0) (w_0_3)}] (wgr_0) {};
\node[circle, draw, minimum size=10pt, above=2cm of wgr_0](rnn_0){\HUGE $f_{rnn}$};
\foreach \hidden / \j in {0/0,1/1.4}
    \node[circle,fill=black!25,draw,inner sep=0pt,minimum size=25pt, above=3cm of rnn_0, yshift=\j cm] (h_0_\hidden) {};
\node[group,fit={(h_0_0) (h_0_1)}, label=north west:\HUGE $\mathbf{h}_1$] (hgr_0) {};
\coordinate (shifted_point) at ($(rnn_text0.north)+(0,4.2)$);
\draw[{Latex[length=6mm, width=5mm]}-, line width=3pt] (wgr_0.south) -- (wgr_0.south |- shifted_point);
\draw[-{Latex[length=6mm, width=5mm]}, line width=3pt] (wgr_0) -- (rnn_0);
\draw[-{Latex[length=6mm, width=5mm]}, line width=3pt] (rnn_0) -- (hgr_0);

% h_-1 (initial hidden state)
\foreach \hidden / \j in {0/0,1/1.4}
    \node[circle,fill=white,draw,inner sep=0pt,minimum size=25pt, left=3cm of h_0_\hidden] (h_init_\hidden) {};
\node[group,fit={(h_init_0) (h_init_1)}, label=north west:\HUGE $\mathbf{h}_0$] (hgr_init) {};
\draw[-{Latex[length=6mm, width=5mm]}, line width=3pt] (hgr_init) -| ($(hgr_init.east)!0.5!(rnn_0.west)$) |- (rnn_0);

% Time step 1 - positioned relative to wgr_0
\foreach \i / \j in {0/0,1/1.4,2/2.8,3/4.2}
    \node[circle,fill=black!75,draw,inner sep=0pt,minimum size=25pt] at ($(wgr_0.east)+(2+\j,0)$) (w_1_\i) {};
\node[group,fit={(w_1_0) (w_1_3)}] (wgr_1) {};
\node[circle, draw, minimum size=10pt, above=2cm of wgr_1](rnn_1){\HUGE $f_{rnn}$};
\foreach \hidden / \j in {0/0,1/1.4}
    \node[circle,fill=black!25,draw,inner sep=0pt,minimum size=25pt, above=3cm of rnn_1, yshift=\j cm] (h_1_\hidden) {};
\node[group,fit={(h_1_0) (h_1_1)}, label=north west:\HUGE $\mathbf{h}_2$] (hgr_1) {};
\coordinate (shifted_point) at ($(rnn_text1.north)+(0,4.2)$);
\draw[{Latex[length=6mm, width=5mm]}-, line width=3pt] (wgr_1.south) -- (wgr_1.south |- shifted_point);
\draw[-{Latex[length=6mm, width=5mm]}, line width=3pt] (wgr_1) -- (rnn_1);
\draw[-{Latex[length=6mm, width=5mm]}, line width=3pt] (rnn_1) -- (hgr_1);
\draw[-{Latex[length=6mm, width=5mm]}, line width=3pt] (hgr_0) -| ($(hgr_0.east)!0.5!(rnn_1.west)$) |- (rnn_1);

% Time step dots - positioned relative to wgr_1
\foreach \i / \j in {0/0,1/1.4,2/2.8,3/4.2}
    \node[circle,fill=black!75,draw,inner sep=0pt,minimum size=25pt] at ($(wgr_1.east)+(3+\j,0)$) (w_2_\i) {};
\node[group,fit={(w_2_0) (w_2_3)}] (wgr_2) {};
\node[circle, draw, minimum size=10pt, above=2cm of wgr_2](rnn_2){\HUGE $f_{rnn}$};
\foreach \hidden / \j in {0/0,1/1.4}
    \node[circle,fill=black!25,draw,inner sep=0pt,minimum size=25pt, above=3cm of rnn_2, yshift=\j cm] (h_2_\hidden) {};
\node[group,fit={(h_2_0) (h_2_1)}] (hgr_2) {};
\node[images, left=.8cm of hgr_2] (hgr_dots) {\HUGER $\cdots$};
\coordinate (shifted_point) at ($(rnn_text2.north)+(0,2.5)$);
\draw[{Latex[length=6mm, width=5mm]}-, line width=3pt] (wgr_2.south) -- (wgr_2.south |- shifted_point);
\draw[-{Latex[length=6mm, width=5mm]}, line width=3pt] (wgr_2) -- (rnn_2);
\draw[-{Latex[length=6mm, width=5mm]}, line width=3pt] (rnn_2) -- (hgr_2);

% Time step N - positioned relative to wgr_2
\foreach \i / \j in {0/0,1/1.4,2/2.8,3/4.2}
    \node[circle,fill=black!75,draw,inner sep=0pt,minimum size=25pt] at ($(wgr_2.east)+(2+\j,0)$) (w_N_\i) {};
\node[group,fit={(w_N_0) (w_N_3)}] (wgr_N) {};
\node[circle, draw, minimum size=10pt, above=2cm of wgr_N](rnn_N){\HUGE $f_{rnn}$};
\foreach \hidden / \j in {0/0,1/1.4}
    \node[circle,fill=black!25,draw,inner sep=0pt,minimum size=25pt, above=3cm of rnn_N, yshift=\j cm] (h_N_\hidden) {};
\node[group,fit={(h_N_0) (h_N_1)}, label=north west:\HUGE $\mathbf{h}_N$] (hgr_N) {};
\coordinate (shifted_point) at ($(rnn_text3.north)+(0,4.2)$);
\draw[{Latex[length=6mm, width=5mm]}-, line width=3pt] (wgr_N.south) -- (wgr_N.south |- shifted_point);
\draw[-{Latex[length=6mm, width=5mm]}, line width=3pt] (wgr_N) -- (rnn_N);
\draw[-{Latex[length=6mm, width=5mm]}, line width=3pt] (rnn_N) -- (hgr_N);
\draw[-{Latex[length=6mm, width=5mm]}, line width=3pt] (hgr_2) -| ($(hgr_2.east)!0.5!(rnn_N.west)$) |- (rnn_N);
\draw[{Latex[length=6mm, width=5mm]}-, line width=3pt] (rnn_output.south) -- (rnn_output.south |- hgr_N.north);

\coordinate (rnn_caption_pos) at ($(rnn_text0)!0.5!(rnn_text3)+(0,-6)$);
\node at (rnn_caption_pos) {\HUGER (b) Recurrent Neural Networks (RNNs)};

%%%%%%%%%%%%%%%%%%%%%%%%%%%%%%%%%%%%%%%%%
% TRANSFORMER ARCHITECTURE (RIGHT)
%%%%%%%%%%%%%%%%%%%%%%%%%%%%%%%%%%%%%%%%%
% Build Transformer layers between images and output

\node[draw, rectangle, fill=AntiqueWhite!75, inner sep=0pt,minimum size=60pt, above=6.5cm of tr_text0] (e_0) {\HUGE $\mathbf{u}_1$};
\node[left=4cm of e_0, text width=3cm] (pos_emb_label) {\HUGE Positional \\ Encoding};
\node[draw, rectangle, fill=AntiqueWhite!75, inner sep=0pt,minimum size=60pt, above=6.5cm of tr_text1] (e_1) {\HUGE $\mathbf{u}_2$};
\node[draw, rectangle, fill=AntiqueWhite!75, inner sep=0pt,minimum size=60pt, above=6.5cm of tr_text1b] (e_1b) {\HUGE $\mathbf{u}_2$};
\node[above=5.5cm of tr_text2] (e_dots) {\HUGER $\cdots$};
\node[draw, rectangle, fill=AntiqueWhite!75, inner sep=0pt,minimum size=60pt, above=6.5cm of tr_text3] (e_N) {\HUGE $\mathbf{u}_N$};

\node[above=0.8cm of e_0] (sign_0) {\HUGE +};
\node[above=0.8cm of e_1] (sign_1) {\HUGE +};
\node[above=0.8cm of e_1b] (sign_1b) {\HUGE +};
\node[above=1.6cm of e_dots] (sign_dots) {\HUGER $\cdots$};
\node[above=0.8cm of e_N] (sign_N) {\HUGE +};

\node[draw, rectangle, fill=LightSteelBlue!50, inner sep=0pt,minimum size=60pt, above=.5cm of sign_0] (v_0) {\HUGE $\mathbf{v}_1$};
\node[left=4cm of v_0, text width=3cm] (temp_emb_label) {\HUGE Temporal \\ Embedding};
\node[draw, rectangle, fill=LightSteelBlue!50, inner sep=0pt,minimum size=60pt, above=.5cm of sign_1] (v_1) {\HUGE $\mathbf{v}_2$};
\node[draw, rectangle, fill=LightSteelBlue!50, inner sep=0pt,minimum size=60pt, above=.5cm of sign_1b] (v_1b) {\HUGE $\mathbf{v}_2$};
\node[above=1.5cm of sign_dots] (v_dots) {\HUGER $\cdots$};
\node[draw, rectangle, fill=LightSteelBlue!50, inner sep=0pt,minimum size=60pt, above=.5cm of sign_N] (v_N) {\HUGE $\mathbf{v}_N$};

\node[draw, rectangle, fill=DarkSeaGreen!50, minimum height=2.cm, minimum width=24cm, right=41cm of ds_model] (transformer) {\HUGE Transformer Encoder};

\path[-{Latex[length=6mm, width=5mm]}, line width=3pt] ([yshift=3.8cm]tr_text0.north) edge (e_0.south);
\path[-{Latex[length=6mm, width=5mm]}, line width=3pt] ([yshift=3.8cm]tr_text1.north) edge (e_1.south);
\path[-{Latex[length=6mm, width=5mm]}, line width=3pt] ([yshift=3.8cm]tr_text1b.north) edge (e_1b.south);
\path[-{Latex[length=6mm, width=5mm]}, line width=3pt] ([yshift=3.8cm]tr_text3.north) edge (e_N.south);

\path[-{Latex[length=6mm, width=5mm]}, line width=3pt] (v_0.north) edge (v_0.north|-transformer.south);
\path[-{Latex[length=6mm, width=5mm]}, line width=3pt] (v_1.north) edge (v_1.north|-transformer.south);
\path[-{Latex[length=6mm, width=5mm]}, line width=3pt] (v_1b.north) edge (v_1b.north|-transformer.south);
\path[-{Latex[length=6mm, width=5mm]}, line width=3pt] (v_N.north) edge (v_N.north|-transformer.south);
\draw[{Latex[length=6mm, width=5mm]}-, line width=3pt] (tr_output.south) -- (tr_output.south |- transformer.north);

\coordinate (tr_caption_pos) at ($(tr_text0)!0.5!(tr_text3)+(0,-6)$);
\node at (tr_caption_pos) {\HUGER (c) Transformer with PE + TE};

\end{tikzpicture}
\end{adjustbox}

%% file: Tables/dataset_collection_properties.tex
\begin{tabular*}{\columnwidth}{@{}l@{\extracolsep{\fill}}r@{}}\\
\toprule
Statistic & Value \\
\midrule
Number of artists                       & 849 \\ 
Number of paintings                     & 59,458  \\
Time period                      & 1409--2012  \\
\midrule
Max paintings/artist             & 1,884 \\
Average paintings/artist            & 70.03 \\
\midrule
Max sequence length              & 77 \\
Average sequence length             & 18.87 \\
\midrule
Max set size                     & 326 \\
Average set size       & 3.71 \\
Average set size/timestep         & 5.06 \\
\midrule
Number of rankings                      & 10 \\
\bottomrule
\end{tabular*}

%% file: Tables/rankings_statistics.tex
\begin{adjustbox}{max totalsize={\textwidth}{\textheight}, center}
\begin{tabular}{@{}lcccccccc@{}}\\
\toprule  & \multicolumn{3}{c}{Stratified split} & & & & \\
\cmidrule(lr){2-4}
Ranking    & {Train} & {Validation} & {Test} & Ranked & Ties & Range & Time period & Aggregate  \\ \midrule
eBooks & 591 (40,369) & 81 (6,647) & 177 (12,442) & 550 & 63 & [1 , 346] & N/A & Borda count on four publicly available sources \\ 
The New York Times & 591 (37,261) & 83 (7,752) & 175 (14,445) & 401 & 39 & [1 , 62] & January, 1981 - December, 2019 & Sum over time  \\
Wikipedia Mentions & 590 (40,480) & 81 (4,612) & 178 (14,366) & 732 & 26 & [1 , 44] & N/A & Sum \\
Wikipedia Links & 591 (43,030) & 82 (4,834) & 176 (11,594) & 560 & 30 & [1 , 40] & N/A & Sum \\
Wikipedia Pageviews & 589 (41,046) & 80 (4,830) & 180 (13,582) & 817 & 1 & [1 , 818] & July, 2015 - December, 2020 & Sum over time \\
Google Ngram & 589 (39,819) & 80 (4,915) & 180 (14,724) & 691 & 1 & [1 , 692] & 16th Century - 21st Century & Sum over time \\
Google Trends & 591 (42,542) & 81 (4,164) & 177 (12,752) & 554 & 66 & [1 , 385] & April, 2017 - April, 2022 & Sum over time\\
Artfacts & 591 (41,155) & 81 (7,009) & 177 (11,294) & 441 & 2 & [1 , 441] & January, 2016 - December, 2016 & N/A  \\
Artprice & 592 (42,122) & 82 (5,422) & 175 (11,914) & 366 & 18 & [1 , 349] & January, 2006 - December, 2021 & Borda count on the top-500 annual rankings \\
Overall & 589 (42,439) & 80 (6,023) & 180 (10,996) & 833 & 26 & [1 , 809] & N/A & Borda count on all the individual rankings\\   \midrule
Time series split & 601 (51,928) & 83 (3,407) & 165 (4,123) & N/A & N/A & N/A & N/A & N/A\\
\bottomrule
\end{tabular}
\end{adjustbox}

%% file: Tables/top_10_artists_per_ranking.tex
\begin{adjustbox}{max totalsize={\textwidth}{\textheight}, center}
\begin{tabular}{@{}l l l l l l l l l l@{}}
\toprule
 eBooks & The New York Times & Wikipedia Mentions & Wikipedia Links & Wikipedia Pageviews & Google Ngram & Google Trends & Artfacts & Artprice &  Aggregate Ranking \\ \midrule
Michelangelo & Andy Warhol & Pablo Picasso & Pablo Picasso & Winston Churchill & Titian & Vincent van Gogh & Andy Warhol & Pablo Picasso & Pablo Picasso  \\
Raphael & Michelangelo & Michelangelo & Henri Matisse & Leonardo da Vinci & Raphael & Pablo Picasso & Pablo Picasso & Andy Warhol & Andy Warhol  \\
Leonardo da Vinci & Roy Lichtenstein & Henri Matisse & Paul Cézanne  & Vincent van Gogh & Rembrandt & Salvador Dalí & Bruce Nauman & Claude Monet &  Vincent van Gogh   \\
Paul Cézanne & John Russell & Raphael & Jackson Pollock&  Pablo Picasso & Winston Churchill & Andy Warhol & Gerhard Richter & Gerhard Richter & Jackson Pollock \\
Pablo Picasso & Willem De Kooning & Titian &  Michelangelo& Andy Warhol & Leonardo da Vinci & Leonardo da Vinci & Sol Lewitt & Roy Lichtenstein & Claude Monet \\
Vincent van Gogh & Richard Serra & Jackson Pollock & Leonardo da Vinci & Michelangelo & Michelangelo & Michelangelo & Man Ray & Marc Chagall & Henri Matisse \\
Claude Monet & Jackson Pollock & Paul Cézanne & Titian& Salvador Dalí & Joshua Reynolds & Claude Monet & Roy Lichtenstein & Joan Miro & Mark Rothko \\
Rembrandt & David Smith & Rembrandt & Raphael &  Claude Monet & Tintoretto & Rembrandt & Marcel Duchamp & Willem De Kooning  & Paul Klee \\
Édouard Manet & Sol Lewitt & Mark Rothko & Mark Rothko & Jackson Pollock & Donatello & Caravaggio & Richard Serra  & Alexander Calder & Paul Gauguin\\
Edgar Degas & Mark Rothko & Georges Braque & Vincent van Gogh  & Rembrandt & Le Corbusier & Wassily Kandinsky & 
Paul Klee & Henri Matisse & Edgar Degas \\\bottomrule
\end{tabular}
\end{adjustbox}

%% file: Tables/results_time_series_split.tex
\resizebox{\textwidth}{!}{\begin{tabular}{@{}l >{\columncolor[gray]{.92}}c>{\columncolor[gray]{.92}}c>{\columncolor[gray]{.92}}c>{\columncolor[gray]{.92}}ccccccccccccccccccccc@{}}
\toprule Method & \multicolumn{2}{c}{\thead{\phantom{}\\Learderboard}} & \multicolumn{2}{c}{\thead{Aggregate\\Ranking}}    & \multicolumn{2}{c}{\thead{\phantom{}\\eBooks}} & \multicolumn{2}{c}{\thead{The New\\York Times}} & \multicolumn{2}{c}{\thead{Wikipedia\\Mentions}} & \multicolumn{2}{c}{\thead{Wikipedia\\Links}} & \multicolumn{2}{c}{\thead{Wikipedia\\Pageviews}} & \multicolumn{2}{c}{\thead{Google\\Ngram}} & \multicolumn{2}{c}{\thead{Google\\Trends}} & \multicolumn{2}{c}{\thead{\phantom{}\\Artfacts}} & \multicolumn{2}{c}{\thead{\phantom{}\\Artprice}}       \\
\cmidrule(r){2-3}
\cmidrule(lr){4-5}
\cmidrule(lr){6-7}
\cmidrule(lr){8-9}
\cmidrule(lr){10-11}
\cmidrule(lr){12-13}
\cmidrule(lr){14-15}
\cmidrule(lr){16-17}
\cmidrule(lr){18-19}
\cmidrule(l){20-21}
\cmidrule(l){22-23}
& \cellcolor{white} BT $\bigtriangleup$ & \cellcolor{white} Borda $\bigtriangleup$ & \cellcolor{white} $\tau$ $\bigtriangleup$ & \cellcolor{white} MAE $\bigtriangledown$ & $\tau$ $\bigtriangleup$ & MAE $\bigtriangledown$ & $\tau$ $\bigtriangleup$ & MAE $\bigtriangledown$ & $\tau$ $\bigtriangleup$ & MAE $\bigtriangledown$ & $\tau$ $\bigtriangleup$ & MAE $\bigtriangledown$ & $\tau$ $\bigtriangleup$ & MAE $\bigtriangledown$ & $\tau$ $\bigtriangleup$ & MAE $\bigtriangledown$ & $\tau$ $\bigtriangleup$ & MAE $\bigtriangledown$ & $\tau$ $\bigtriangleup$ & MAE $\bigtriangledown$ & $\tau$ $\bigtriangleup$ & MAE $\bigtriangledown$ \\ \midrule  
                   Vanilla (mean) & 1,070 & 64 & 0.062 & 0.960 & 0.038 & 1.065 & 0.013 & 0.665 & 0.112 & 1.123 & 0.130 & 1.183 & 0.026 & 1.070 & 0.057 & 1.161 & 0.073 & 1.154 & -0.059 & 1.029 & 0.021 & 1.503 \\
Vanilla (max) & 1,170 & 92 & 0.124 & 0.571 & 0.107 & 0.640 & 0.154 & 0.477 & 0.081 & 0.299 & 0.094 & 0.312 & 0.051 & 0.635 & 0.062 & 0.792 & 0.006 & 0.698 & 0.118 & 0.628 & 0.056 & 0.672 \\
Gradient Boosting (mean) & 1,176 & 94 & 0.092 & 0.264 & 0.037 & 0.268 & 0.105 & 0.195 & -0.052 & 0.133 & 0.066 & 0.152 & 0.164 & 0.270 & 0.079 & 0.256 & -0.038 & 0.299 & 0.047 & 0.314 & 0.010 & 0.342 \\
Gradient Boosting (max) & 1,414 & 199 & 0.222 & 0.230 & 0.208 & 0.213 & 0.147 & 0.161 & 0.018 & 0.085 & 0.081 & 0.094 & 0.137 & 0.262 & 0.074 & 0.248 & 0.112 & 0.231 & 0.078 & 0.317 & 0.154 & 0.271 \\\midrule
DeepSets (mean) & 1,346 & 163 & 0.136 & 0.272 & 0.188 & 0.153 & 0.072 & 0.127 & -0.092 & 0.066 & -0.048 & 0.073 & 0.139 & 0.274 & 0.119 & 0.240 & -0.016 & 0.132 & 0.085 & 0.309 & 0.024 & 0.251 \\
DeepSets (max) & \underline{1,767} & \underline{387} & 0.218 & \textbf{0.216} & 0.236 & 0.114 & 0.171 & 0.113 & 0.159 & 0.051 & \underline{0.178} & \textbf{0.052} & 0.197 & 0.242 & \underline{0.179} & \textbf{0.219} & 0.071 & \textbf{0.096} & 0.100 & 0.361 & 0.165 & \underline{0.185} \\
Set Transformer (SAB + PMA) & 1,435 & 210 & 0.173 & 0.232 & 0.039 & 0.151 & 0.111 & 0.128 & 0.132 & 0.063 & 0.064 & 0.078 & 0.134 & 0.263 & 0.118 & 0.247 & 0.066 & 0.155 & 0.112 & 0.308 & 0.113 & 0.260 \\
Set Transformer (ISAB + PMA) & 1,465 & 226 & 0.105 & 0.260 & 0.110 & 0.121 & 0.103 & \underline{0.111} & 0.052 & \underline{0.048} & 0.118 & \textbf{0.052} & 0.093 & 0.265 & 0.063 & 0.267 & 0.070 & 0.115 & 0.029 & 0.327 & 0.162 & 0.201 \\
Set Transformer (ISAB + PMA + SAB) & 1,450 & 219 & 0.157 & 0.232 & 0.156 & 0.127 & -0.001 & 0.123 & 0.023 & 0.063 & 0.079 & 0.075 & 0.153 & 0.256 & 0.170 & 0.248 & 0.067 & 0.146 & 0.079 & 0.325 & 0.143 & 0.242 \\\midrule
Hierarchical DeepSets (max) & 1,483 & 238 & 0.135 & 0.264 & 0.216 & 0.153 & 0.087 & 0.123 & 0.191 & 0.065 & 0.159 & 0.073 & 0.182 & 0.262 & 0.155 & 0.244 & 0.019 & 0.215 & 0.129 & 0.299 & 0.049 & 0.305 \\
LSTM (mean) & 1,474 & 232 & 0.154 & 0.255 & 0.192 & 0.147 & 0.136 & 0.122 & 0.042 & 0.069 & 0.089 & 0.085 & 0.135 & 0.256 & 0.113 & 0.278 & \textbf{0.169} & 0.166 & 0.107 & 0.315 & \underline{0.190} & 0.191 \\
LSTM (max) & 1,443 & 217 & 0.074 & 0.279 & 0.059 & 0.119 & 0.083 & 0.143 & \textbf{0.246} & 0.057 & \textbf{0.206} & 0.071 & 0.130 & 0.267 & 0.107 & 0.249 & 0.106 & 0.125 & 0.056 & 0.317 & 0.086 & 0.236 \\
Transformer w/o PE & 1,584 & 293 & 0.141 & 0.250 & 0.151 & 0.149 & 0.143 & 0.117 & 0.083 & 0.058 & 0.173 & 0.073 & \underline{0.223} & 0.236 & 0.120 & 0.286 & 0.063 & \underline{0.104} & 0.119 & 0.308 & 0.165 & \underline{0.185} \\
Transformer + PE & 1,600 & 301 & 0.178 & 0.235 & 0.184 & 0.172 & 0.141 & 0.125 & 0.088 & 0.051 & 0.069 & 0.062 & 0.186 & \textbf{0.230} & 0.136 & 0.281 & 0.082 & 0.119 & \textbf{0.194} & 0.314 & \textbf{0.193} & 0.198 \\\midrule
Set2Seq BiLSTM (DeepSets - mean) & 1,527 & 261 & 0.193 & 0.253 & \underline{0.240} & 0.132 & 0.168 & 0.115 & 0.132 & 0.053 & 0.108 & 0.068 & 0.121 & 0.250 & -0.007 & 0.245 & 0.071 & 0.128 & 0.102 & 0.315 & 0.086 & 0.247 \\
Set2Seq BiLSTM  (DeepSets - max) & 1,656 & 334 & 0.180 & 0.239 & 0.236 & \textbf{0.108} & 0.241 & 0.122 & \underline{0.197} & 0.052 & 0.168 & 0.064 & 0.189 & 0.244 & 0.150 & 0.254 & 0.060 & 0.125 & 0.111 & 0.302 & 0.153 & 0.269 \\
Set2Seq BiLSTM  (SAB + PMA) & 1,514 & 256 & 0.167 & 0.240 & 0.117 & 0.134 & 0.132 & 0.157 & 0.091 & 0.076 & 0.144 & 0.090 & 0.183 & 0.256 & 0.137 & 0.246 & 0.100 & 0.149 & 0.110 & 0.301 & 0.176 & 0.231 \\
Set2Seq BiLSTM  (ISAB + PMA) & 1,476 & 230 & 0.157 & 0.239 & 0.172 & 0.140 & -0.016 & 0.124 & 0.074 & 0.051 & 0.160 & 0.073 & 0.158 & \underline{0.234} & 0.110 & 0.278 & 0.094 & 0.149 & 0.056 & 0.335 & 0.098 & 0.200 \\
Set2Seq BiLSTM  (ISAB + PMA + SAB) & 1,445 & 215 & 0.127 & 0.267 & 0.086 & \underline{0.109} & 0.176 & 0.137 & 0.125 & 0.051 & 0.121 & 0.071 & 0.130 & 0.262 & 0.080 & 0.270 & 0.122 & 0.187 & 0.049 & 0.335 & 0.100 & 0.214 \\
Set2Seq Transformer (DeepSets - mean) & 1,667 & 338 & \underline{0.237} & 0.225 & 0.202 & 0.115 & 0.184 & \textbf{0.109} & 0.117 & 0.049 & 0.136 & 0.061 & 0.195 & 0.243 & 0.120 & 0.265 & 0.060 & 0.133 & 0.145 & \underline{0.291} & 0.057 & 0.214 \\
Set2Seq Transformer (DeepSets - max) & \textbf{1,786} & \textbf{393} & 0.236 & 0.238 & \textbf{0.254} & 0.125 & \underline{0.253} & \underline{0.111} & 0.170 & \textbf{0.043} & 0.162 & 0.055 & 0.220 & 0.237 & \textbf{0.205} & \underline{0.222} & 0.062 & 0.111 & 0.122 & 0.307 & 0.104 & 0.209 \\
Set2Seq Transformer (SAB + PMA) & 1,676 & 344 & 0.169 & 0.244 & 0.174 & 0.115 & 0.236 & 0.119 & 0.111 & 0.049 & 0.166 & 0.057 & \textbf{0.253} & 0.236 & 0.090 & 0.243 & 0.104 & 0.120 & 0.107 & 0.294 & 0.128 & 0.210 \\
Set2Seq Transformer (ISAB + PMA) & 1,618 & 312 & 0.221 & \underline{0.222} & 0.135 & 0.131 & \textbf{0.255} & 0.113 & 0.095 & 0.057 & 0.100 & 0.069 & 0.141 & 0.240 & 0.100 & 0.231 & \underline{0.135} & 0.113 & 0.184 & 0.324 & 0.069 & 0.221 \\
Set2Seq Transformer (ISAB + PMA + SAB) & 1,758 & 382 & \textbf{0.259} & \textbf{0.216} & 0.118 & 0.110 & 0.249 & 0.112 & 0.127 & 0.053 & 0.096 & \underline{0.054} & 0.189 & 0.251 & 0.147 & 0.223 & 0.109 & 0.128 & \underline{0.192} & \textbf{0.284} & 0.160 & \textbf{0.183} \\\bottomrule
\end{tabular}}

%% file: Tables/visual_embeddings_random_split.tex
\resizebox{\textwidth}{!}{\begin{tabular}{@{}l @{\hskip .1in} cccccccccccccccccccccc@{}}
\toprule Method & \multicolumn{2}{c}{\thead{\phantom{}\\eBooks}} & \multicolumn{2}{c}{\thead{The New\\York Times}} & \multicolumn{2}{c}{\thead{Wikipedia\\Mentions}} & \multicolumn{2}{c}{\thead{Wikipedia\\Links}} & \multicolumn{2}{c}{\thead{Wikipedia\\Pageviews}} & \multicolumn{2}{c}{\thead{Google\\Ngram}} & \multicolumn{2}{c}{\thead{Google\\Trends}} & \multicolumn{2}{c}{\thead{\phantom{}\\Artfacts}} & \multicolumn{2}{c}{\thead{\phantom{}\\Artprice}} & \multicolumn{2}{c}{\thead{Aggregate\\Ranking}}   & \multicolumn{2}{c}{\thead{\phantom{}\\Average}}       \\
\cmidrule(r){2-3}
\cmidrule(lr){4-5}
\cmidrule(lr){6-7}
\cmidrule(lr){8-9}
\cmidrule(lr){10-11}
\cmidrule(lr){12-13}
\cmidrule(lr){14-15}
\cmidrule(lr){16-17}
\cmidrule(lr){18-19}
\cmidrule(lr){20-21}
\cmidrule(l){22-23}
& $\tau$ $\bigtriangleup$ & MAE $\bigtriangledown$ & $\tau$ $\bigtriangleup$ & MAE $\bigtriangledown$ & $\tau$ $\bigtriangleup$ & MAE $\bigtriangledown$ & $\tau$ $\bigtriangleup$ & MAE $\bigtriangledown$ & $\tau$ $\bigtriangleup$ & MAE $\bigtriangledown$ & $\tau$ $\bigtriangleup$ & MAE $\bigtriangledown$ & $\tau$ $\bigtriangleup$ & MAE $\bigtriangledown$ & $\tau$ $\bigtriangleup$ & MAE $\bigtriangledown$ & $\tau$ $\bigtriangleup$ & MAE $\bigtriangledown$ & $\tau$ $\bigtriangleup$ & MAE $\bigtriangledown$ & $\tau$ $\bigtriangleup$ & MAE $\bigtriangledown$\\ \midrule
                      CLIP  &             0.319 & \underline{0.170} &             0.179 &             0.111 &             0.291 &             0.092 &             0.251 &             0.088 &    \textbf{0.385} & \underline{0.202} &             0.272 &             0.252 &             0.258 &             0.205 &    \textbf{0.480} &    \textbf{0.198} &             0.323 &             0.203 &             0.261 & \underline{0.208} &             0.302 & \underline{0.173} \\
                      BLIP  &             0.350 &             0.172 &    \textbf{0.355} &             0.115 &             0.300 &             0.100 &             0.330 &             0.091 &             0.292 &             0.236 &    \textbf{0.361} &    \textbf{0.228} &             0.219 &             0.212 &             0.433 &             0.215 &             0.256 &             0.223 &             0.322 &             0.217 & \underline{0.322} &             0.181 \\
                     BLIP-2  &             0.331 &             0.185 &             0.247 &             0.117 &    \textbf{0.336} &    \textbf{0.083} &             0.216 &             0.101 &             0.307 &             0.230 &             0.128 &             0.269 &             0.198 &             0.211 &             0.380 &             0.229 &             0.271 &             0.210 &             0.289 &             0.220 &             0.270 &             0.186 \\\midrule
                     VGG-19  &             0.144 &             0.230 &             0.305 &    \textbf{0.097} &             0.184 &             0.104 &             0.200 &             0.098 &             0.331 &             0.225 &             0.271 &             0.257 &             0.315 & \underline{0.191} &             0.424 &             0.225 &             0.321 &    \textbf{0.185} &             0.183 &             0.237 &             0.268 &             0.185 \\
                  ResNet-18  &             0.208 &             0.215 &             0.300 &             0.112 &             0.108 &             0.101 &             0.293 &    \textbf{0.081} &             0.302 &             0.228 &             0.226 &             0.270 &             0.213 &             0.213 &             0.405 &             0.215 &             0.318 &             0.195 &             0.282 &             0.218 &             0.265 &             0.185 \\
                  ResNet-34  &             0.267 &             0.196 &             0.312 & \underline{0.107} &             0.202 &             0.092 &             0.232 &             0.088 &             0.348 &             0.209 &             0.191 &             0.272 &             0.252 &             0.204 &             0.356 &             0.239 &             0.238 &             0.211 &             0.244 &             0.223 &             0.264 &             0.184 \\
                  ResNet-50  &             0.341 &             0.206 &             0.210 &             0.129 &             0.201 &             0.108 &             0.319 &             0.093 &             0.363 &             0.210 &             0.246 &             0.251 &             0.221 &             0.215 &             0.412 &             0.220 &             0.272 &             0.218 &             0.293 &             0.210 &             0.288 &             0.186 \\
                 ResNet-101  & \underline{0.366} &             0.185 &             0.203 &             0.137 &             0.281 &             0.088 &             0.209 &             0.092 &             0.346 &             0.209 &             0.258 &             0.244 &             0.241 &             0.210 &             0.422 &             0.228 &             0.244 &             0.234 &             0.187 &             0.250 &             0.276 &             0.188 \\
                 ResNet-152  &             0.280 &             0.206 &             0.278 &             0.115 &             0.236 &             0.097 &             0.313 & \underline{0.085} & \underline{0.376} &             0.204 &             0.233 &             0.270 &             0.191 &             0.224 &             0.380 &             0.219 & \underline{0.338} & \underline{0.186} &             0.228 &             0.229 &             0.285 &             0.184 \\
             ConvNeXt  &             0.279 &             0.208 &             0.234 &             0.110 &             0.245 &             0.090 &             0.330 &             0.090 &             0.282 &             0.243 &             0.174 &             0.275 &             0.143 &             0.227 &             0.412 &             0.225 &             0.123 &             0.237 &             0.259 &             0.223 &             0.248 &             0.193 \\
                       ViT  &             0.267 &             0.205 &             0.301 &             0.111 &             0.248 &             0.092 &             0.302 &             0.086 &             0.254 &             0.231 &             0.234 &             0.261 &             0.285 &             0.194 &             0.350 &             0.235 &             0.325 &             0.201 & \underline{0.347} &             0.209 &             0.291 &             0.183 \\\midrule
 ResNet-152 (Tag Prediction)  &             0.323 &             0.190 &             0.340 &             0.112 &             0.198 &             0.109 &    \textbf{0.373} &             0.090 &             0.285 &             0.225 &             0.126 &             0.276 &             0.286 &             0.207 &             0.407 &             0.223 &             0.241 &             0.212 &             0.244 &             0.232 &             0.282 &             0.188 \\
 ArtSAGENet (Tag Prediction) &             0.327 &             0.179 &             0.289 &             0.109 & \underline{0.313} &             0.089 &             0.232 &             0.091 &             0.340 &             0.215 &             0.278 &             0.252 & \underline{0.337} &             0.211 &             0.275 &             0.250 &             0.186 &             0.221 &             0.336 &             0.209 &             0.291 &             0.183 \\
               ArtSAGENet (MTL)  &    \textbf{0.492} &    \textbf{0.138} & \underline{0.342} & \underline{0.107} &             0.204 & \underline{0.085} & \underline{0.342} &             0.090 &             0.364 &    \textbf{0.196} & \underline{0.343} & \underline{0.232} &    \textbf{0.375} &    \textbf{0.180} & \underline{0.438} & \underline{0.205} &    \textbf{0.347} &             0.197 &    \textbf{0.486} &    \textbf{0.176} &    \textbf{0.373} &    \textbf{0.161} \\

\bottomrule
\end{tabular}}

%% file: Tables/visual_embeddings_time_series_split.tex
\resizebox{\textwidth}{!}{\begin{tabular}{@{}l @{\hskip .1in} cccccccccccccccccccccc@{}}
\toprule Method & \multicolumn{2}{c}{\thead{\phantom{}\\eBooks}} & \multicolumn{2}{c}{\thead{The New\\York Times}} & \multicolumn{2}{c}{\thead{Wikipedia\\Mentions}} & \multicolumn{2}{c}{\thead{Wikipedia\\Links}} & \multicolumn{2}{c}{\thead{Wikipedia\\Pageviews}} & \multicolumn{2}{c}{\thead{Google\\Ngram}} & \multicolumn{2}{c}{\thead{Google\\Trends}} & \multicolumn{2}{c}{\thead{\phantom{}\\Artfacts}} & \multicolumn{2}{c}{\thead{\phantom{}\\Artprice}} & \multicolumn{2}{c}{\thead{Aggregate\\Ranking}}   & \multicolumn{2}{c}{\thead{\phantom{}\\Average}}       \\
\cmidrule(r){2-3}
\cmidrule(lr){4-5}
\cmidrule(lr){6-7}
\cmidrule(lr){8-9}
\cmidrule(lr){10-11}
\cmidrule(lr){12-13}
\cmidrule(lr){14-15}
\cmidrule(lr){16-17}
\cmidrule(lr){18-19}
\cmidrule(lr){20-21}
\cmidrule(l){22-23}
& $\tau$ $\bigtriangleup$ & MAE $\bigtriangledown$ & $\tau$ $\bigtriangleup$ & MAE $\bigtriangledown$ & $\tau$ $\bigtriangleup$ & MAE $\bigtriangledown$ & $\tau$ $\bigtriangleup$ & MAE $\bigtriangledown$ & $\tau$ $\bigtriangleup$ & MAE $\bigtriangledown$ & $\tau$ $\bigtriangleup$ & MAE $\bigtriangledown$ & $\tau$ $\bigtriangleup$ & MAE $\bigtriangledown$ & $\tau$ $\bigtriangleup$ & MAE $\bigtriangledown$ & $\tau$ $\bigtriangleup$ & MAE $\bigtriangledown$ & $\tau$ $\bigtriangleup$ & MAE $\bigtriangledown$ & $\tau$ $\bigtriangleup$ & MAE $\bigtriangledown$\\ \midrule
                                           CLIP  &             0.304 &             0.197 &             0.205 &             0.176 &             0.260 &             0.110 &             0.193 &             0.133 &             0.188 &             0.248 &             0.042 &             0.257 &             0.045 &             0.244 &             0.250 &             0.244 &             0.175 &             0.284 &             0.272 &             0.235 &             0.193 &             0.213 \\
                      BLIP  &             0.228 &             0.211 &             0.270 &             0.167 &             0.262 &             0.122 &             0.283 &             0.130 &             0.213 &             0.237 &             0.134 & \underline{0.237} &             0.093 &             0.221 &    \textbf{0.361} &    \textbf{0.218} &             0.060 &             0.318 & \underline{0.366} & \underline{0.208} &             0.227 &             0.207 \\
                     BLIP-2  &             0.300 &             0.225 &             0.284 &             0.167 &             0.298 &             0.109 &             0.311 &             0.129 &             0.124 &             0.257 &             0.111 &             0.246 &             0.032 & \underline{0.203} & \underline{0.323} & \underline{0.234} &             0.177 &             0.289 &             0.268 &             0.217 &             0.223 &             0.208 \\ \midrule
                     VGG-19  &             0.253 &    \textbf{0.183} &             0.166 &             0.159 &             0.152 &             0.120 &             0.235 &             0.125 &             0.193 &             0.239 &             0.140 &             0.248 &             0.110 &             0.223 &             0.176 &             0.265 &             0.163 &             0.272 &             0.276 &             0.231 &             0.186 &             0.206 \\
                  ResNet-18  &             0.281 &             0.201 &             0.296 & \underline{0.137} &             0.306 & \underline{0.100} &             0.275 &    \textbf{0.115} &             0.086 &             0.253 &            -0.004 &             0.263 &             0.072 &             0.222 &             0.176 &             0.261 &             0.154 & \underline{0.266} &             0.236 &             0.226 &             0.188 & \underline{0.204} \\
                  ResNet-34  &             0.246 &             0.228 &             0.303 &             0.152 &             0.218 &             0.117 &             0.189 &             0.128 &             0.166 &             0.260 &             0.004 &             0.252 &             0.144 &             0.224 &             0.221 &             0.250 &             0.132 &             0.301 &             0.176 &             0.246 &             0.180 &             0.216 \\
                  ResNet-50  &             0.295 &             0.199 &             0.287 &             0.158 &             0.238 &             0.120 & \underline{0.335} &             0.121 &             0.247 &    \textbf{0.218} &             0.105 &             0.245 &             0.145 &             0.222 &             0.224 &             0.253 &             0.130 &             0.303 &             0.123 &             0.252 &             0.213 &             0.209 \\
                 ResNet-101  &             0.106 &             0.212 &             0.209 &             0.160 &             0.177 &             0.115 &             0.195 &             0.124 &             0.195 &             0.235 &            -0.059 &             0.276 &             0.130 &             0.229 &             0.090 &             0.284 &             0.163 &             0.293 &             0.235 &             0.241 &             0.144 &             0.217 \\
                 ResNet-152  &             0.184 &             0.220 &             0.282 &             0.160 &             0.334 &             0.123 &             0.313 &             0.134 &             0.207 &             0.247 &             0.068 &             0.249 &             0.040 &             0.247 &             0.142 &             0.276 &             0.232 &             0.279 &             0.205 &             0.233 &             0.201 &             0.217 \\
             ConvNeXt  &             0.266 &             0.195 &             0.162 &             0.170 &             0.188 &             0.120 &             0.158 &             0.139 & \underline{0.282} & \underline{0.229} &             0.113 &             0.239 &             0.162 &             0.205 &             0.251 &             0.242 &             0.244 &             0.280 &             0.248 &             0.220 &             0.207 & \underline{0.204} \\
                       ViT  & \underline{0.345} &             0.192 &             0.283 &    \textbf{0.136} &    \textbf{0.413} &    \textbf{0.095} &             0.261 & \underline{0.119} &             0.141 &             0.247 &             0.115 &             0.242 & \underline{0.226} &             0.212 &             0.194 &             0.258 &             0.209 &             0.280 &             0.350 &             0.212 & \underline{0.254} &    \textbf{0.199} \\ \midrule
 ResNet-152 (Tag Prediction)   &             0.311 & \underline{0.188} & \underline{0.308} &             0.157 &             0.299 &             0.125 &             0.252 &             0.155 &             0.130 &             0.257 &             0.008 &             0.257 &             0.079 &             0.223 &             0.114 &             0.275 &    \textbf{0.298} &    \textbf{0.247} &             0.198 &             0.241 &             0.200 &             0.212 \\
 ArtSAGENet (Tag Prediction) &             0.172 &             0.195 &             0.220 &             0.182 &             0.262 &             0.116 &             0.161 &             0.152 &             0.220 &             0.241 & \underline{0.185} &             0.238 &    \textbf{0.253} &    \textbf{0.194} &             0.231 &             0.258 &             0.075 &             0.316 &             0.269 &             0.230 &             0.205 &             0.212 \\
      ArtSAGENet (MTL)   &    \textbf{0.357} &             0.231 &    \textbf{0.404} &             0.185 & \underline{0.369} &             0.119 &    \textbf{0.423} &             0.123 &    \textbf{0.337} &             0.232 &    \textbf{0.322} &    \textbf{0.216} &             0.139 &             0.228 &             0.268 &             0.242 & \underline{0.266} &             0.270 &    \textbf{0.393} &    \textbf{0.197} &    \textbf{0.328} & \underline{0.204} \\
\bottomrule
\end{tabular}}

%% file: Tables/mesogeos_dataset_variables.tex
\resizebox{\textwidth}{!}{\begin{tabular}{@{}lccccc@{}}
\toprule
Variable                                     & Spatial & Temporal       & Units      & Type      & Feature Name              \\ \midrule
Max Temperature                              & 9 km    & Hourly         & K          & Dynamic   & t2m               \\
Max Dewpoint Temperature                     & 9 km    & Hourly         & K          & Dynamic   & d2m               \\
Max Wind Speed                               & 9 km    & Hourly         & m/s        & Dynamic   & wind\_speed        \\
Max Surface Pressure                         & 9 km    & Hourly         & Pa         & Dynamic   & sp                \\
Min Relative Humidity                        & 9 km    & Hourly         & \%/100      & Dynamic   & rh                \\
Total Precipitation                          & 9 km    & Hourly         & m          & Dynamic   & tp                \\
Mean Surface Solar Radiation Downwards       & 9 km    & Hourly         & J/m²       & Dynamic   & ssrd              \\
Day Land Surface Temperature                 & 1 km    & Daily          & K          & Dynamic   & lst\_day           \\
Night Land Surface Temperature               & 1 km    & Daily          & K          & Dynamic   & lst\_night         \\
Normalized Difference Vegetation Index (NDVI)& 500 m   & 16-days        & -          & Dynamic   & ndvi              \\
Leaf Area Index (LAI)                        & 500 m   & 8-days         & -          & Dynamic   & lai               \\
Soil moisture                                & 5 km    & 10-days        & -          & Dynamic   & smi               \\
Timestamp                                    & -       & Daily   & Datetime   & Dynamic   & time         \\ \midrule
Population                                   & 1 km    & Yearly         & people/km² & Semi-Static    & population        \\
Fraction of agriculture                      & 300 m   & Yearly         & \%/100      & Semi-Static    & lc\_agriculture    \\
Fraction of forest                           & 300 m   & Yearly         & \%/100      & Semi-Static    & lc\_forest         \\
Fraction of grassland                        & 300 m   & Yearly         & \%/100      & Semi-Static    & lc\_grassland      \\
Fraction of settlements                      & 300 m   & Yearly         & \%/100      & Semi-Static    & lc\_settlement     \\
Fraction of shrubland                        & 300 m   & Yearly         & \%/100      & Semi-Static    & lc\_shrubland      \\
Fraction of sparse vegetation                & 300 m   & Yearly         & \%/100      & Semi-Static    & lc\_sparse\_vegetation \\
Fraction of water bodies                     & 300 m   & Yearly         & \%/100      & Semi-Static    & lc\_water\_bodies   \\
Fraction of wetland                          & 300 m   & Yearly         & \%/100      & Semi-Static    & lc\_wetland        \\
Roads distance                               & 1 km    & Static         & km         & Static    & roads\_distance    \\
Elevation                                    & 30 m    & Static         & m          & Static    & dem               \\
Slope                                        & 30 m    & Static         & rad        & Static    & slope             \\ \midrule
Wildfire Danger       & 1 km         & Daily         & Binary (0/1)  & Target    & burned\_area\_has         \\\bottomrule
\end{tabular}}

%% file: Tables/mesogeos_results_set_transformer.tex
\resizebox{\textwidth}{!}{\begin{tabular}{@{}lcccccccccc@{}}\toprule
\multirow{2}{*}{Method}              & \multicolumn{2}{c}{$K = 1,\ N = 30$} & \multicolumn{2}{c}{$K = 2,\ N = 15$} & \multicolumn{2}{c}{$K = 3,\ N = 10$} & \multicolumn{2}{c}{$K = 5,\ N = 6$} & \multicolumn{2}{c}{$K = 10,\ N = 3$} \\
\cmidrule(lr){2-3}
\cmidrule(lr){4-5}
\cmidrule(lr){6-7}
\cmidrule(lr){8-9}
\cmidrule(l){10-11}

                                        & F1  & PR-AUC   & F1  & PR-AUC   & F1  & PR-AUC   & F1  & PR-AUC    & F1  & PR-AUC  \\ \midrule
            Set2Seq Transformer (SAB + PMA) + PE   &             0.784 &             0.861 &             0.782 &             0.850 &             0.760 &             0.845 &             0.757 &             0.839 &             0.732 &                 0.812  \\
           Set2Seq Transformer (SAB + PMA) + PE + TE &             0.783 &             0.864 &    \textbf{0.789} & \underline{0.862} &    \textbf{0.773} &    \textbf{0.856} &             0.757 &             0.831 & \underline{0.751} &                 \underline{0.823}  \\ \midrule
           Set2Seq Transformer (ISAB + PMA) + PE   &             0.776 &             0.865 &             0.772 &             0.848 &             0.741 & \underline{0.851} &             0.755 &             0.832 &             0.736 &                0.816   \\
           Set2Seq Transformer (ISAB + PMA) + PE + TE &             0.784 &             0.866 & \underline{0.783} &             0.861 & \underline{0.772} &    \textbf{0.856} &    \textbf{0.771} &    \textbf{0.850} &    \textbf{0.761} &                \textbf{0.835}   \\ \midrule
     Set2Seq Transformer (ISAB + PMA +SAB) + PE   & \underline{0.790} & \underline{0.867} &             0.773 &             0.852 &             0.770 &             0.844 & \underline{0.766} &             0.835 &             0.747 &                 0.809  \\
           Set2Seq Transformer (ISAB + PMA +SAB) + PE + TE &    \textbf{0.792} &    \textbf{0.872} &             0.777 &    \textbf{0.864} &             0.762 &             0.849 &             0.764 & \underline{0.840} &             0.721 &                 0.822  \\ \bottomrule
\end{tabular}}

%% file: Tables/mesogeos_results_pe_ce.tex
\resizebox{\textwidth}{!}{\begin{tabular}{@{}lcccccccccc@{}}\toprule
\multirow{2}{*}{Method}              & \multicolumn{2}{c}{$K = 1,\ N = 30$} & \multicolumn{2}{c}{$K = 2,\ N = 15$} & \multicolumn{2}{c}{$K = 3,\ N = 10$} & \multicolumn{2}{c}{$K = 5,\ N = 6$} & \multicolumn{2}{c}{$K = 10,\ N = 3$} \\
\cmidrule(lr){2-3}
\cmidrule(lr){4-5}
\cmidrule(lr){6-7}
\cmidrule(lr){8-9}
\cmidrule(l){10-11}

                                        & F1  & PR-AUC   & F1  & PR-AUC   & F1  & PR-AUC   & F1  & PR-AUC    & F1  & PR-AUC  \\ \midrule
                 Set2Seq Transformer (DeepSets)   &             0.703 &             0.786 &             0.717 &             0.772 &             0.712 &             0.786 &             0.721 &             0.785 &             0.738 &                 0.796  \\
            Set2Seq Transformer (DeepSets) + PE   &             0.771 &             0.855 &             0.773 &             0.856 &             0.768 &             0.846 &             0.764 &             0.836 &             0.749 &                0.826   \\
             Set2Seq Transformer (DeepSets) + TE\textsuperscript{MY}  &             0.737 &             0.812 &             0.734 &             0.812 &             0.767 &             0.822 &             0.746 &             0.814 &             0.737 &                 0.813  \\
           Set2Seq Transformer (DeepSets) + PE + TE\textsuperscript{DM}  &             0.782 &             0.859 & \underline{0.779} &             0.856 & \underline{0.775} & \underline{0.851} &    \textbf{0.772} &             0.844 &    \textbf{0.763} &                 0.831  \\
           Set2Seq Transformer (DeepSets) + PE + TE\textsuperscript{MY}  &             0.784 &             0.864 &             0.761 &    \textbf{0.863} &             0.768 &             0.833 &             0.756 &             0.830 &             0.747 &                \underline{0.834}   \\
           Set2Seq Transformer (DeepSets) + PE + TE\textsuperscript{DMY} &             0.782 &             0.860 &             0.770 &             0.852 &    \textbf{0.783} & \underline{0.851} &             0.744 & \underline{0.849} &             0.755 &  0.831                 \\ \midrule
                 Set2Seq Transformer (ISAB + PMA)  &             0.755 &             0.853 &             0.756 &             0.850 &             0.748 &             0.837 &             0.769 &             0.834 &             0.727 &                 0.797  \\
           Set2Seq Transformer (ISAB + PMA) + PE   &             0.776 &             0.865 &             0.772 &             0.848 &             0.741 & \underline{0.851} &             0.755 &             0.832 &             0.736 &                0.816   \\
            Set2Seq Transformer (ISAB + PMA) + TE\textsuperscript{MY}  &             0.786 &    \textbf{0.871} &             0.777 &             0.845 &             0.734 &             0.811 &             0.753 &             0.835 &             0.748 &                 0.822  \\
      Set2Seq Transformer (ISAB + PMA) + PE + TE\textsuperscript{DM} & \underline{0.791} &             0.863 &             0.766 &             0.846 &             0.767 &             0.834 &             0.736 &             0.838 &             0.740 &                 0.821  \\
          Set2Seq Transformer (ISAB + PMA) + PE + TE\textsuperscript{MY}  &             0.784 & \underline{0.866} &    \textbf{0.783} & \underline{0.861} &             0.772 &    \textbf{0.856} & \underline{0.771} &    \textbf{0.850} & \underline{0.761} &                \textbf{0.835}   \\
          Set2Seq Transformer (ISAB + PMA) + PE + TE\textsuperscript{DMY} &    \textbf{0.796} &             0.864 &    \textbf{0.783} &             0.854 &             0.773 &             0.850 &             0.745 &             0.839 &             0.754 &                 0.827  \\ \bottomrule
\end{tabular}}